\documentclass[10pt, twocolumn, twoside]{IEEEtran}
\ifCLASSINFOpdf
\else
\fi

\usepackage{amsmath,epsfig}
\usepackage{graphicx}
\usepackage{cite}
\usepackage{amssymb}
\usepackage{subfigure}

\newtheorem{theorem}{Theorem}[section]

\newenvironment{definition}[1][Definition]{\begin{trivlist}
\item[\hskip \labelsep {\bfseries #1}]}{\end{trivlist}}

  \newcommand{\R}{\mathbb{R}}

  \renewcommand{\c}{\mathbf{c}}
  \newcommand{\e}{\mathbf{e}}
  \newcommand{\f}{\mathbf{f}}
  \newcommand{\g}{\mathbf{g}}

  \newcommand{\uu}{\mathbf{u}}
  
  \renewcommand{\v}{\mathbf{v}}
  \newcommand{\V}{\mathbf{V}}
  \newcommand{\w}{\mathbf{w}}
  
  \newcommand{\x}{\mathbf{x}}
  
  \newcommand{\y}{\mathbf{y}}
  \newcommand{\z}{\mathbf{z}}
  \newcommand{\0}{\mathbf{0}}

  \newcommand{\cS}{\mathcal{S}}

  \newcommand{\cX}{\mathcal{X}}
  \newcommand{\cY}{\mathcal{Y}}
  \newcommand{\cZ}{\mathcal{Z}}

  \newcommand{\trans}{^\top}

  \newcommand{\rank}{\mathrm{rank\;}}

\begin{document}
%
\title{Compressive Sensing of Sparse Tensors}

\author{Shmuel~Friedland,~
       Qun~Li*,~\IEEEmembership{Member,~IEEE,}
        and~Dan~Schonfeld,~\IEEEmembership{Fellow,~IEEE}
\thanks{Copyright (c) 2013 IEEE. Personal use of this material is permitted. However, permission to use this material for any other purposes must be obtained from the IEEE by sending a request to pubs-permissions@ieee.org.}
\thanks{S. Friedland is with the Department
of Mathematics, Statistics \& Computer Science, University of
Illinois at Chicago (UIC), Chicago,
IL, 60607-7045 USA. e-mail: friedlan@uic.edu. This work was supported by NSF grant DMS-1216393.}
\thanks{Q. Li is with PARC, Xerox Corporation, Webster, NY, 14580 USA. e-mail: Qun.Li@xerox.com. This work was done during her PhD study at UIC, advised by Prof. Schonfeld.}\thanks{D. Schonfeld is with the Department of Electrical and Computer Engineering, University of Illinois at Chicago, Chicago, IL, 60607 USA. e-mail: dans@uic.edu.}}

\markboth{Manuscript of IEEE Transactions on Image Processing, EDICS:~SMR-Rep}%
{}

\maketitle

\begin{abstract}
Compressive sensing (CS) has triggered enormous research activity
since its first appearance. CS exploits the signal's sparsity or
compressibility in a particular domain and integrates data
compression and acquisition, thus allowing exact reconstruction
through relatively few non-adaptive linear measurements. While
conventional CS theory relies on data representation in the form
of vectors, many data types in various applications such as color
imaging, video sequences, and multi-sensor networks, are
intrinsically represented by higher-order tensors. Application of
CS to higher-order data representation is typically performed by
conversion of the data to very long vectors that must be measured
using very large sampling matrices, thus imposing a huge
computational and memory burden. In this paper, we propose
Generalized Tensor Compressive Sensing (GTCS)--a unified framework
for compressive sensing of higher-order tensors which preserves
the intrinsic structure of tensor data with reduced computational
complexity at reconstruction. GTCS offers an efficient means for
representation of multidimensional data by providing simultaneous
acquisition and compression from all tensor modes. In addition, we
propound two reconstruction procedures, a serial method (GTCS-S)
and a parallelizable method (GTCS-P). We then compare the
performance of the proposed method with Kronecker compressive
sensing (KCS) and multi-way compressive sensing (MWCS). We
demonstrate experimentally that GTCS outperforms KCS and MWCS in
terms of both reconstruction accuracy (within a range of
compression ratios) and processing speed. The major disadvantage
of our methods (and of MWCS as well), is that the compression
ratios may be worse than that offered by KCS.
\end{abstract}


\section{Introduction}
\label{sec:intro} Recent literature has witnessed an explosion of
interest in sensing that exploits structured prior knowledge in
the general form of sparsity, meaning that signals can be
represented by only a few coefficients in some domain. Central to
much of this recent work is the paradigm of compressive sensing
(CS), also known under the terminology of compressed sensing or
compressive sampling \cite{Robustuncertainty,Nearoptimal,Donoho}.
CS theory permits relatively few linear measurements of the signal
while still allowing exact reconstruction via nonlinear recovery
process. The key idea is that the sparsity helps in isolating the
original vector. The first intuitive approach to a reconstruction
algorithm consists in searching for the sparsest vector that is
consistent with the linear measurements. However, this
$\ell_0$-minimization problem is NP-hard in general and thus
computationally infeasible. There are essentially two approaches
for tractable alternative algorithms. The first is convex
relaxation, leading to $\ell_1$-minimization
\cite{Donoho_Logan_1992}, also known as basis pursuit
\cite{Chen96atomicdecomposition}, whereas the second constructs
greedy algorithms. Besides, in image processing, the use of
total-variation minimization which is closely connected to
$\ell_1$-minimization first appears in
\cite{Rudin_Osher_Fatemi_1992} and is widely applied later on. By
now basic properties of the measurement matrix which ensure sparse
recovery by $\ell_1$-minimization are known: the null space
property (NSP) \cite{Cohen09compressedsensing} and the restricted
isometry property (RIP) \cite{Candes_2008}.

An intrinsic limitation in conventional CS theory is that it
relies on data representation in the form of vectors. In fact,
many data types do not lend themselves to vector data
representation. For example, images are intrinsically matrices. As
a result, great efforts have been made to extend traditional CS to
CS of data in matrix representation. A straightforward
implementation of CS on 2D images recasts the 2D problem as
traditional 1D CS problem by converting images to long vectors,
such as in \cite{Vectorization}. However, despite of considerably
huge memory and computational burden imposed by the use of long
vector data and large sampling matrix, the sparse solutions
produced by straightforward $\ell_1$-minimization often incur
visually unpleasant, high-frequency oscillations. This is due to
the neglect of attributes known to be widely possessed by images,
such as smoothness. In \cite{totalvariation}, instead of seeking
sparsity in the transformed domain, a total variation-based
minimization was proposed to promote smoothness of the
reconstructed image. Later, as an alternative for alleviating the
huge computational and memory burden associated with image
vectorization, block-based CS (BCS) was proposed in
\cite{blockCS}. In BCS, an image is divided into non-overlapping
blocks and acquired using an appropriately-sized measurement
matrix.

Another direction in the extension of CS to matrix CS generalizes CS concept
and outlines a dictionary relating concepts from cardinality minimization to
those of rank minimization
\cite{matrixrank,garanteedminimumrank,matrxicompletion}. The affine rank
minimization problem consists of finding a matrix of minimum rank that
satisfies a given set of linear equality constraints. It encompasses commonly
seen low-rank matrix completion problem \cite{matrxicompletion} and low-rank
matrix approximation problem as special cases. \cite{matrixrank} first
introduced recovery of the minimum-rank matrix via nuclear norm minimization.
\cite{garanteedminimumrank} generalized the RIP in \cite{Candes_2008} to matrix
case and established the theoretical condition under which the nuclear norm
heuristic can be guaranteed to produce the minimum-rank solution.

Real-world signals of practical interest such as color imaging, video sequences
and multi-sensor networks, are usually generated by the interaction of multiple
factors or multimedia and thus can be intrinsically represented by higher-order
tensors. Therefore, the higher-order extension of CS theory for
multidimensional data has become an emerging topic. One direction attempts to
find the best rank-R tensor approximation as a recovery of the original data
tensor as in \cite{tensordecompositionCS}, they also proved the existence and
uniqueness of the best rank-R tensor approximation in the case of 3rd order
tensors under appropriate assumptions. In \cite{MWCS}, multi-way compressed
sensing (MWCS) for sparse and low-rank tensors suggests a two-step recovery
process: fitting a low-rank model in compressed domain, followed by per-mode
decompression. However, the performance of MWCS relies highly on the estimation
of the tensor rank, which is an NP-hard problem. The other direction
\cite{kroneckerproductmatrices}\cite{kroneckerCS} uses Kronecker product
matrices in CS to act as sparsifying bases that jointly model the structure
present in all of the signal dimensions as well as to represent the measurement
protocols used in distributed settings. However, the recovery procedure, due to
the vectorization of multidimensional signals, is rather time consuming and not
applicable in practice. We proposed in \cite{ICME2013GTCS} Generalized Tensor
Compressive Sensing (GTCS)--a unified framework for compressive sensing of
higher-order tensors. In addition, we presented two reconstruction procedures,
a serial method (GTCS-S) and a parallelizable method (GTCS-P). Experimental
results demonstrated the outstanding performance of GTCS in terms of both
recovery accuracy and speed. In this paper, we not only illustrate the
technical details of GTCS more thoroughly, but also further examine its
performance on the recovery of various types of data including sparse image,
compressible image, sparse video and compressible video comprehensively.

The rest of the paper is organized as follows. Section
\ref{bkreview} briefly reviews concepts and operations from
multilinear algebra used later in the paper. It also introduces
conventional compressive sensing theory. Section \ref{tensorcomp}
proposes GTCS theorems along with their detailed proofs. Section
\ref{experiments} then compares experimentally the performance of
the proposed method with existing methods. Finally, Section
\ref{conclusion} concludes the paper.
\vspace{-0.3cm}
\section{Background}\label{bkreview}
Throughout the discussion, lower-case characters represent scalar
values $(a,b,\ldots)$, bold-face characters represent vectors
$(\mathbf{a},\mathbf{b},\ldots)$, capitals represent matrices
$(A,B,\ldots)$ and calligraphic capitals represent tensors
$(\mathcal{A},\mathcal{B},\ldots)$. Let $[N]$ denote the set $\{1,
2, \ldots,N\}$, where $N$ is a positive integer.
\vspace{-0.3cm}
\subsection{Multilinear algebra}\label{mareview}
A tensor is a multidimensional array. The order of a tensor is the
number of modes. For instance, tensor $\mathcal{X}\in
\mathbb{R}^{N_1\times \ldots\times N_d}$ has order $d$ and the
dimension of its $i^{th}$ mode (also called mode $i$ directly) is
$N_i$.

\begin{definition}
[Kronecker product] {The Kronecker product of matrices $A\in
\mathbb{R}^{I\times J}$ and $B\in \mathbb{R}^{K\times L}$ is
denoted by $A\otimes B$. The result is a matrix of size$(I\cdot
K)\times (J\cdot L)$ defined by

$A\otimes B=\left(%
\begin{array}{cccc}
  a_{11}B & a_{12}B & \cdots & a_{1J}B \\
  a_{21}B & a_{22}B & \cdots & a_{2J}B \\
  \vdots & \vdots & \ddots & \vdots \\
  a_{I1}B & a_{I2}B & \cdots & a_{IJ}B \\
\end{array}%
\right)$}.
\end{definition}

\begin{definition}
[Outer product and tensor product] {The operator $\circ$ denotes
the tensor product between two vectors. In linear algebra, the
outer product typically refers to the tensor product between two
vectors, that is, $u\circ v=uv\trans$. In this paper, the terms
outer product and tensor product are equivalent. The Kronecker
product and the tensor product between two vectors are related by
$u\circ v=u\otimes v\trans.$}
\end{definition}
\begin{definition}
[Mode-i product] {The mode-i product of a tensor
$\mathcal{X}=[x_{\alpha_1,\ldots,\alpha_i, \ldots, \alpha_d}]\in
\mathbb{R}^{N_1\times \ldots\times N_i\times\ldots\times N_d}$ and
a matrix $U=[u_{j,\alpha_i}]\in \mathbb{R}^{J\times N_i}$ is
denoted by $\mathcal{X}\times _i U$ and is of size
$N_1\times\ldots\times N_{i-1}\times J \times N_{i+1}\times
\ldots\times N_d$. Element-wise, the mode-$i$ product can be
written as $(\mathcal{X}\times _i
U)_{\alpha_1,\ldots,\alpha_{i-1},j,\alpha_{i+1},\ldots,\alpha_d} =
\sum_{\alpha_i=1}^{N_i}x_{\alpha_1,\ldots,\alpha_i,
\ldots,\alpha_d}u_{j,\alpha_i}$.}
\end{definition}
\begin{definition}
[Mode-i fiber and mode-i unfolding] {A mode-i fiber of a tensor
$\mathcal{X}=[x_{\alpha_1,\ldots,\alpha_i, \ldots,\alpha_d}]\in
\mathbb{R}^{N_1\times\ldots\times N_i\times\ldots\times N_d}$ is
obtained by fixing every index but $\alpha_i$. The mode-i
unfolding $X_{(i)}$ of $\mathcal{X}$ arranges the mode-i fibers to
be the columns of the resulting $N_i\times (N_1\cdot\ldots\cdot
N_{i-1}\cdot N_{i+1}\cdot\ldots\cdot N_d)$ matrix. }

$\mathcal{Y} = \mathcal{X}\times_1 U_1\times \ldots\times_d U_d$
is equivalent to $Y_{(i)} = U_iX_{(i)}(U_d\otimes \ldots\otimes
U_{i+1}\otimes U_{i-1}\otimes \ldots\otimes U_1)\trans$.
\end{definition}

\begin{definition}
[Core Tucker decomposition]\cite{Tuck1964}{ Let $\cX\in
\R^{N_1\times \ldots \times N_d}$ with mode-i unfolding
$X_{(i)}\in \R^{N_i\times (N_1\cdot\ldots \cdot N_{i-1}\cdot
N_{i+1}\cdot\ldots \cdot N_d)}$. Denote by $R_i(\cX)\subset
\R^{N_i}$ the column space of $X_{(i)}$ whose rank is $r_i$. Let
$\c_{1,i},\ldots, \c_{r_i,i}$ be a basis in $R_i(\cX)$. Then the
subspace $\V(\cX):=R_1(\cX)\circ\ldots\circ R_d(\cX) \subset
\R^{N_1\times \ldots \times N_d}$ contains $\cX$. Clearly a basis
in $\V$ consists of the vectors $\c_{i_1,1}\circ\ldots\circ
\c_{i_d,d}$ where $i_j\in [r_j]$ and $j\in [d]$. Hence the core
Tucker decomposition of $\cX$ is
 \begin{equation}\label{cortcukdec}
 \cX=\sum_{i_j\in[r_j],j\in[d]} \xi_{i_1,\ldots,i_d} \c_{i_1,1}\circ\ldots\circ \c_{i_d,d}.
 \end{equation}}\end{definition}

A special case of core Tucker decomposition is the higher-order
singular value decomposition (HOSVD). Any tensor $\cX\in
\R^{N_1\times \ldots \times N_d}$ can be written as

\begin{equation}\label{HOSVD}
\cX=\cS\times_1 U_1\times\ldots\times_d U_d,
\end{equation}
where $U_i = [\mathbf{u}_1\cdots\mathbf{u}_{N_i}]$ is orthogonal
for $i\in [d]$ and $\mathcal{S}$ is called the core tensor which
can be obtained easily by $\mathcal{S} = \mathcal{X}\times_1
U_1\trans\times\ldots\times_d U_d\trans$.

There are many ways to get a weaker decomposition as
\begin{equation}\label{ranklikedec}
\cX=\sum_{i=1}^K \mathbf{a}_i^{(1)}\circ\ldots\circ
\mathbf{a}_i^{(d)}, \quad \mathbf{a}_i^{(j)}\in R_j(\cX), j\in
[d].
\end{equation}
A simple constructive way is as follows. First decompose $X_{(1)}$
as $X_{(1)}=\sum_{j=1}^{r_1} \c_{j,1} \g_{j,1}\trans$ (e.g.\ by
singular value decomposition (SVD)). Now each $\g_{j,1}$ can be
represented as a tensor of order $d-1$ $\in R_2(\cX)\circ
\ldots\circ R_d(\cX)\subset\mathbb{R}^{N_2\times \ldots\times
N_d}$. Unfold each $\g_{j,1}$ in mode $2$ to obtain
${\g_{j,1}}_{(2)}$ and decompose
${\g_{j,1}}_{(2)}=\sum_{l=1}^{r_2}\mathbf{d}_{l,2,j}\f_{l,2,j}\trans,
\quad \mathbf{d}_{l,2,j}\in R_2(\cX), \mathbf{f}_{l,2,j}\in
R_3(\cX)\circ\ldots\circ R_d(\cX).$ By successively unfolding and
decomposing each remaining tensor mode, a decomposition of the
form in Eq.~\eqref{ranklikedec} is obtained. Note that if $\cX$ is
$s$-sparse then each vector in $R_i(\cX)$ is $s$-sparse and
$r_i\leq s$ for $i\in [d]$. Hence $K\le s^{d-1}$.
\begin{definition}
[CANDECOMP/PARAFAC
decomposition]\cite{Kolda09tensordecompositions}{ For a tensor
$\mathcal{X}\in \mathbb{R}^{N_1\times \ldots\times N_d}$, the
CANDECOMP/PARAFAC (CP) decomposition is $\mathcal{X}\approx
[\lambda; A^{(1)},\ldots,A^{(d)}]\equiv\sum_{r=1}^R\lambda_r
\mathbf{a}_r^{(1)}\circ\ldots\circ \mathbf{a}_r^{(d)},$ where
$\mathbf{\lambda}=[\lambda_1\ldots\lambda_R]\trans\in
\mathbb{R}^R$ and
$A^{(i)}=[\mathbf{a}_1^{(i)}\cdots\mathbf{a}_R^{(i)}]\in
\mathbb{R}^{N_i\times R}$ for $i\in [d].$}
\end{definition}
\vspace{-0.3cm}
\subsection{Compressive sensing}
Compressive sensing is a framework for reconstruction of signals
that have sparse representations. A vector $\mathbf{x}\in
\mathbb{R}^N$ is called $s$-sparse if it has at most $s$ nonzero
entries. The CS measurement protocol measures the signal
$\mathbf{x}$ with the measurement matrix $A\in \mathbb{R}^{m\times
N}$ where $m<N$ and encodes the information as $\mathbf{y}\in
\mathbb{R}^m$ where $\mathbf{y} = A\mathbf{x}$. The decoder knows
$A$ and attempts to recover $\mathbf{x}$ from $\mathbf{y}$. Since
$m<N$, typically there are infinitely many solutions for such an
under-constrained problem. However, if $\mathbf{x}$ is known to be
sufficiently sparse, then exact recovery of $\mathbf{x}$ is
possible, which establishes the fundamental tenet of CS theory.
The recovery is achieved by finding a solution $\z^\star\in\R^N$
satisfying
 \begin{equation}\label{l1minrec}
 \z^\star=\arg \min\{\|\z\|_1,\; A\z=\y\}.
 \end{equation}
Such $\z^\star$ coincides with $\x$ under certain condition. The
following well known result states that each $s$-sparse signal can
be recovered uniquely if $A$ satisfies the null space property of
order $s$, denoted as NSP$_s$. That is, if $A\w=\0,
\w\in\R^N\setminus\{\0\}$, then for any subset $S\subset \{1, 2,
\ldots,N\}$ with cardinality $|S|=s$ it holds that
$\|\mathbf{v}_S\|_1<\|\mathbf{v}_{S^c}\|_1$, where $\mathbf{v}_S$
denotes the vector that coincides with $\mathbf{v}$ on the index
set $S$ and is set to zero on $S^c$.

One way to generate such $A$ is by sampling its entries using
numbers generated from a Gaussian or a Bernoulli distribution.
This matrix generation process guarantees that there exists a
universal constant $c$ such that if
\begin{equation}\label{nspcond}
m\ge 2cs\ln \frac{N}{s},
\end{equation}
then the recovery of $\x$ using \eqref{l1minrec} is successful
with probability greater than $1- \exp(-\frac{m}{2c})$
\cite{Rauhut_compressivesensing}.

In fact, most signals of practical interest are not really sparse
in any domain. Instead, they are only compressible, meaning that
in some particular domain, the coefficients, when sorted by
magnitude, decay according to the power law
\cite{Donoho98datacompression}. Given a signal $\x\in
\mathbb{R}^N$ which can be represented by $\mathbf{\theta}\in
\mathbb{R}^N$ in some transformed domain, i.e.\ $\x = \Phi\theta$,
with sorted coefficients such that $\|\theta\|_{(1)}\geq \ldots
\geq \|\theta\|_{(N)}$, it obeys that $\|\theta\|_{(n)}\leq R
n^{-\frac{1}{p}}$ for each $n\in [N]$, where $0<p<1$ and $R$ is
some constant. According to \cite{Nearoptimal,
Candes06sparsityand}, when $A$ is drawn randomly from a Gaussian
or Bernoulli distribution and $\Phi$ is an orthobasis, with
overwhelming probability, the solution $\g^\star\in \R^N$ to

\begin{equation}\label{l1minrecoptimal}
 \g^\star=\arg \min\{\|\g\|_1,\; A\Phi\g=\y\}
 \end{equation}
is unique. Furthermore, denote by $\x_\sharp$ the recovered signal
via $\x_\sharp = \Phi \g$, with a very large probability we have
the approximation
\[\|\x-\x_\sharp\|_2\leq CR(\frac{m}{\ln
N})^{\frac{1}{2}-\frac{1}{p}},\] where $A\in \R^{m\times N}$ is
generated randomly as stated above, $m$ is the number of
measurements, and $C$ is some constant. This provides theoretical
foundation for CS of compressible signals.

Consider the case where the observation $\y$ is noisy. For a given
integer $s$, a matrix $A\in \R^{m\times N}$ satisfies the
restricted isometry property (RIP$_s$) if
 $$(1-\delta_s)\|\x\|_2\le \|A\x\|_2\le (1+\delta_s)\|\x\|_2$$
 for all s-sparse $\x\in\mathbb{R}^N$ and for some $\delta_s\in(0,1)$. Given a noisy observation $\y$ with bounded error
$\epsilon$, an approximation of the signal $\x$, $\f^\star\in
\R^N$, can be obtained by solving the following relaxed recovery
problem \cite{Noisyrecover},

\begin{equation}\label{l1minrecnoisy}
 \f^\star=\arg \min\{\|\f\|_1,\; \|A\f-\y\|_2\leq \epsilon\}.
 \end{equation}

It is known that if $A$ satisfies the RIP$_{2s}$ property with
$\delta_{2s}\in (0, \sqrt{2}-1)$, then
\begin{equation}\label{sigrecer1}
 \|\f^\star -\x\|_2\le C_2\epsilon,  \textrm{ where } C_2=\frac{4\sqrt{1+\delta_{2s}}}{1-(1+\sqrt{2})\delta_{2s}}.
\end{equation}

Recently, the extension of CS theory for multidimensional signals
has become an emerging topic. The objective of our paper is to
consider the case where the $s$-sparse vector $\x$ is represented
as an $s$-sparse tensor $\cX=[x_{i_1, i_2, \ldots,i_d}]\in
\R^{N_1\times N_2 \times \ldots\times N_d}$. Specifically, in the
sampling phase, we construct a set of measurement matrices
$\{U_1,U_2,\ldots,U_d\}$ for all tensor modes, where $U_i\in
\R^{m_i\times N_i}$ for $i\in [d]$, and sample $\cX$ to obtain
$\cY=\cX\times_1 U_1\times_2 U_2\times \ldots \times_d
U_d\in\R^{m_1\times m_2\times \ldots\times m_d}$. Note that our
sampling method is mathematically equivalent to that proposed in
\cite{kroneckerCS}, where $A$ is expressed as a Kronecker product
$A:=U_1\otimes U_2\otimes\ldots\otimes U_d$, which requires $m$ to
satisfy
\begin{equation}\label{nspcondmd}
m\ge 2cs(-\ln s+\sum_{i=1}^d\ln N_i).
\end{equation}

We show that if each $U_i$ satisfies the NSP$_s$ property, then we
can recover $\cX$ uniquely from $\cY$ by solving a sequence of
$\ell_1$ minimization problems, each similar to the expression in
\eqref{l1minrec}. This approach is advantageous relative to
vectorization-based compressive sensing methods because the
corresponding recovery problems are in terms of $U_i$'s instead of
$A$, which results in greatly reduced complexity. If the entries
of $U_i$ are sampled from Gaussian or Bernoulli distributions, the
following set of condition needs to be satisfied:
\begin{equation}\label{nspcondmdi}
m_i\ge 2cs\ln \frac{N_i}{s}, \quad i\in [d].
\end{equation}
Observe that the dimensionality of the original signal $\cX$,
namely $N=N_1\cdot\ldots\cdot N_d$, is compressed to
$m=m_1\cdot\ldots\cdot m_d$. Hence, the number of measurements
required by our method must satisfy
\begin{equation}\label{totcompineq}
m\ge (2cs)^d \prod_{i=1}^d  \ln \frac{N_i}{s},
\end{equation}
which indicates a worse compression ratio than that from
\eqref{nspcondmd}. Note that \eqref{totcompineq} is derived under
the assumption that each fiber has the same sparsity as the
tensor, and hence is very loose.

We propose two reconstruction procedures, a serial method (GTCS-S)
and a parallelizable method (GTCS-P) in terms of recovery of each
tensor mode. A similar idea to GTCS-P, namely multi-way
compressive sensing (MWCS) \cite{MWCS} for sparse and low-rank
tensors, also suggests a two-step recovery process: fitting a
low-rank model in the compressed domain, followed by per-mode
decompression. However, the performance of MWCS relies highly on
the estimation of the tensor rank, which is an NP-hard problem.
The proposed GTCS manages to get rid of tensor rank estimation and
thus considerably reduces the computational complexity in
comparison to MWCS.

\section{Generalized tensor compressive sensing}\label{tensorcomp}
In each of the following subsection, we first discuss our method
for matrices, i.e., $d=2$ and then for tensors, i.e., $d\ge 3$.
\subsection{Generalized tensor compressive sensing with serial recovery (GTCS-S)}
\begin{theorem}\label{compsensmatrev}
Let $X=[x_{ij}]\in\R^{N_1\times N_2}$ be $s$-sparse.  Let $U_i\in
\R^{m_i\times N_i}$ and assume that $U_i$ satisfies the NSP$_s$
property for $i\in [2]$. Define
\begin{equation}\label{defmatY}
 Y=[y_{pq}]=U_1 X U_2\trans \in \R^{m_1\times m_2}.
\end{equation}
Then $X$ can be recovered uniquely as follows. Let
$\y_1,\ldots,\y_{m_2}\in\R^{m_1}$ be the columns of $Y$. Let
$\z_i^\star\in\R^{N_1}$ be a solution of
\begin{equation}\label{defzstarMatrix}
 \z_i^\star=\min\{\|\z_i\|_1, \; U_1\z_i=\y_i\}, \quad i\in [m_2].
\end{equation}
Then each $\z_i^\star$ is unique and $s$-sparse.  Let
$Z\in\R^{N_1\times m_2}$ be the matrix with columns
$\z_1^\star,\ldots,\z_{m_2}^\star$. Let $\w_1\trans,\ldots,
\w_{N_1}\trans$ be the rows of $Z$. Then $\uu_j^\star\in\R^{N_2}$,
the transpose of the $j^{th}$ row of $X$, is the solution of
\begin{equation}\label{defustarMatrix}
 \uu_j^\star=\min\{\|\uu_j\|_1, \; U_2\uu_j=\w_j\}, \quad j\in [N_1].
 \end{equation}
\end{theorem}
\begin{IEEEproof}
Let $Z=XU_2\trans\in\R^{N_1\times m_2}$.  Assume that
$\z_1^\star,\ldots,\z_{m_2}^\star$ are the columns of $Z$. Note
that $\z_i^\star$ is a linear combination of the columns of $X$.
$\z_i^\star$ has at most $s$ nonzero coordinates, because the
total number of nonzero elements in $X$ is $s$. Since $Y=U_1 Z$,
it follows that $\y_i=U_1\z_i^\star$. Also, since $U_1$ satisfies
the NSP$_s$ property, we arrive at \eqref{defzstarMatrix}. Observe
that $Z\trans=U_2 X\trans$; hence, $\w_j=U_2 \uu_j^\star$. Since
$X$ is $s$-sparse, then each $\uu_j^\star$ is $s$-sparse.  The
assumption that $U_2$ satisfies the NSP$_s$ property implies
\eqref{defustarMatrix}. This completes the proof.
\end{IEEEproof}

If the entries of $U_1$ and $U_2$ are drawn from random
distributions as described above, then the set of conditions from
\eqref{nspcondmdi} needs to be met as well. Note that although
Theorem \ref{compsensmatrev} requires both $U_1$ and $U_2$ to
satisfy the NSP$_s$ property, such constraints can be relaxed if
each row of $X$ is $s_2$-sparse, where $s_2<s$, and each column of
$XU_2^T$ is $s_1$-sparse, where $s_1<s$.  In this case, it follows
from the proof of Theorem \ref{compsensmatrev} that $X$ can be
recovered as long as $U_1$ and $U_2$ satisfy the NSP$_{s_1}$ and
the NSP$_{s_2}$ properties respectively.

\begin{theorem}[GTCS-S]\label{compsenstenrev}
Let $\cX=[x_{i_1,\ldots,i_d}]\in\R^{N_1\times \ldots \times N_d}$
be $s$-sparse. Let $U_i\in \R^{m_i\times N_i}$ and assume that
$U_i$ satisfies the NSP$_s$ property for $i\in [d]$. Define
\begin{equation}\label{deftenY}
\cY=[y_{j_1,\ldots,j_d}]=\cX\times_1 U_1\times \ldots \times_d U_d  \in
\R^{m_1\times\ldots\times m_d}.
\end{equation}
Then $\cX$ can be recovered uniquely as follows. Unfold $\cY$ in
mode $1$,
\begin{equation*}
Y_{(1)} = U_1X_{(1)}[\otimes_{k=d}^2 U_k]\trans\in\R^{m_1\times
(m_2\cdot\ldots\cdot m_d)}.
\end{equation*}
Let $\y_1,\ldots,\y_{m_2\cdot\ldots\cdot m_d}$ be the columns of
$Y_{(1)}$. Then $\y_i=U_1\z_i$, where each $\z_i\in \R^{N_1}$ is
$s$-sparse. Recover each $\z_i$ using \eqref{l1minrec}. Let $\cZ =
\cX \times_2 U_2\times \ldots \times_d U_d \in \R^{N_1\times
m_2\times\ldots\times m_d}$ with its mode-1 fibers being
$\z_1,\ldots,\z_{m_2\cdot\ldots\cdot m_d}$. Unfold $\cZ$ in mode
2,
\begin{equation*}
Z_{(2)} = U_2X_{(2)}[\otimes_{k=d}^3 U_k\otimes
I]\trans\in\R^{m_2\times (N_1\cdot m_3\cdot\ldots\cdot m_d)}.
\end{equation*}
Let $\w_1,\ldots,\w_{N_1\cdot m_3\cdot\ldots\cdot m_d}$ be the
columns of $Z_{(2)}$. Then $\mathbf{w}_j=U_2\mathbf{v}_j$, where
each $\mathbf{v}_j\in \R^{N_2}$ is $s$-sparse. Recover each
$\mathbf{v}_j$ using \eqref{l1minrec}. $\cX$ can be reconstructed
by successively applying the above procedure to tensor modes
$3,\ldots, d$.
\end{theorem}

The proof follows directly that of Theorem \ref{compsensmatrev}
and hence is skipped here.

Note that although Theorem \ref{compsenstenrev} requires $U_i$ to
satisfy the NSP$_s$ property for $i\in [d]$, such constraints can
be relaxed if each mode-$i$ fiber of $\mathcal{X}\times_{i+1}
U_{i+1}\times \ldots \times_d U_d$ is $s_i$-sparse for $i\in
[d-1]$, and each mode-$d$ fiber of $\mathcal{X}$ is $s_d$-sparse,
where $s_i\leq s$, for $i\in [d]$. In this case, it follows from
the proof of Theorem \ref{compsenstenrev} that $X$ can be
recovered as long as $U_i$ satisfies the NSP$_{s_i}$ property, for
$i\in [d]$.
\vspace{-0.3cm}
\subsection{Generalized tensor compressive sensing with parallelizable recovery (GTCS-P)}
Employing the same definitions of $X$ and $Y$ as in Theorem
\ref{compsensmatrev}, consider a rank decomposition of $X$ with
$\rank (X)=r$, $X=\sum_{i=1}^r \mathbf{z}_i \mathbf{u}_i\trans$,
which could be obtained using either Gauss elimination or SVD.
After sampling we have,
\begin{equation}\label{DecompoYa}
Y=\sum_{i=1}^r (U_1\z_i)(U_2\uu_i)\trans.
\end{equation}

We first show that the above decomposition of $Y$ is also a
rank-$r$ decomposition, i.e., $U_1\z_1,\ldots,U_1\z_r$ and
$U_2\uu_1,\ldots,U_2\uu_r$ are two sets of linearly independent
vectors.

Since $X$ is $s$-sparse, $\rank (Y)\le \rank (X)\le s$.
Furthermore, denote by $R(X)$ the column space of $X$, both $R(X)$
and $R (X\trans)$ are vector subspaces whose elements are
$s$-sparse. Note that $\z_i\in R (X), \uu_i\in R(X\trans)$. Since
$U_1$ and $U_2$ satisfy the NSP$_s$ property, then $\dim (U_1
R(X))= \dim (U_2R (X\trans))=\rank (X)$. Hence the above
decomposition of $Y$ in \eqref{DecompoYa} is a rank-$r$
decomposition of $Y$.
\begin{theorem}\label{GTCSPmatrec}
Let $X=[x_{ij}]\in\R^{N_1\times N_2}$ be $s$-sparse.  Let $U_i\in
\R^{m_i\times N_i}$ and assume that $U_i$ satisfies the NSP$_s$
property for $i \in [2]$.  If $Y$ is given by \eqref{defmatY},
then $X$ can be recovered uniquely as follows. Consider a rank
decomposition (e.g., SVD) of $Y$ such that
\begin{equation}\label{ranklikedec1Y}
Y=\sum_{i=1}^K \mathbf{b}_i^{(1)} (\mathbf{b}_i^{(2)})\trans,
\end{equation}
where $K = \rank(Y)$. Let $\w_i^{(j)\star}\in \mathbb{R}^{N_j}$ be
a solution of
\begin{equation}\label{defzstarP}
\w_i^{(j)\star}=\min\{\|\w_i^{(j)}\|_1,
U_j\w_i^{(j)}=\mathbf{b}_i^{(j)}\}, \quad i\in [K], j\in
[2].\nonumber
\end{equation}
Thus each $\w_i^{(j)\star}$ is unique and $s$-sparse. Then,
\begin{equation}\label{ranklikedecYP}
X=\sum_{i=1}^K \w_i^{(1)\star} (\w_i^{(2)\star})\trans.
\end{equation}
\end{theorem}
\begin{IEEEproof}
First observe that $R( Y)\subset U_1R( X)$ and $R( Y\trans)\subset
U_2R( X\trans)$. Since \eqref{ranklikedec1Y} is a rank
decomposition of $Y$ and $U_i$ satisfies the NSP$_s$ property for
$i \in [2]$, it follows that $\mathbf{b}_i^{(1)}\in U_1R( X)$ and
$\mathbf{b}_i^{(2)}\in U_2R( X\trans)$. Hence $\w_i^{(1)\star}\in
R( X), \w_i^{(2)\star}\in R( X\trans)$ are unique and $s$-sparse.
Let $\hat X:=\sum_{i=1}^K \w_i^{(1)\star}
(\w_i^{(2)\star})\trans$. Assume to the contrary that $X-\hat X\ne
0$. Clearly $R(X-\hat X)\subset R (X), R (X\trans - \hat
X\trans)\subset R( X\trans)$. Let $X-\hat X=\sum_{i=1}^{J}
\uu_i^{(1)} (\uu_i^{(2)})\trans$ be a rank decomposition of
$X-\hat X$.  Hence $\uu_1^{(1)},\ldots,\uu_J^{(1)}\in R( X)$ and
$\uu_1^{(2)},\ldots,\uu_J^{(2)}\in R( X\trans)$ are two sets of
$J$ linearly independent vectors.   Since each vector either in
$R( X)$ or in $R( X\trans)$ is $s$-sparse, and $U_1, U_2$ satisfy
the NSP$_s$ property, it follows that $U_1\uu_1^{(j)},\ldots,
U_1\uu_J^{(j)}$ are linearly independent for $j\in [2]$.  Hence
the matrix $Z:=\sum_{i=1}^J(U_1\uu_i^{(1)})
(U_2\uu_i^{(2)})\trans$ has rank $J$. In particular, $Z\ne 0$.  On
the other hand, $Z=U_1(X-\hat X)U_2\trans=Y-Y=0$, which
contradicts the previous statement.  So $X=\hat X$. This completes
the proof.
\end{IEEEproof}

The above recovery procedure consists of two stages, namely, the
decomposition stage and the reconstruction stage, where the latter
can be implemented in parallel for each matrix mode. Note that the
above theorem is equivalent to multi-way compressive sensing for
matrices (MWCS) introduced in \cite{MWCS}.
\begin{theorem}[GTCS-P]\label{TheoremGTCSP}
Let $\cX=[x_{i_1,\ldots,i_d}]\in\R^{N_1\times \ldots \times N_d}$
be $s$-sparse. Let $U_i\in \R^{m_i\times N_i}$ and assume that
$U_i$ satisfies the NSP$_s$ property for $i\in [d]$. Define $\cY$
as in \eqref{deftenY}, then $\cX$ can be recovered as follows.
Consider a decomposition of $\cY$ such that,
 \begin{align}\label{ranklikedecY}
 &\cY=\sum_{i=1}^K \mathbf{b}_i^{(1)}\circ\ldots\circ \mathbf{b}_i^{(d)}, \quad \mathbf{b}_i^{(j)}\in R_j(\cY)\subseteq U_jR_j(\cX),\nonumber\\
  &  j\in [d].
 \end{align}
 Let
$\w_i^{(j)\star}\in R_j(\cX)\subset \mathbb{R}^{N_j}$ be a solution of
\begin{align}\label{defzstarP}
 &\w_i^{(j)\star}=\min\{\|\w_i^{(j)}\|_1, \; U_j\w_i^{(j)}=\mathbf{b}_i^{(j)}\}, \quad
 i\in [K],\nonumber\\
 &j\in [d].
\end{align}
Thus each $\w_i^{(j)\star}$ is unique and $s$-sparse. Then,
 \begin{equation}\label{ranklikedecYP}
 \cX=\sum_{i=1}^K \w_i^{(1)\star}\circ\ldots\circ \w_i^{(d)\star}, \quad \w_i^{(j)\star}\in
 R_j(\cX), j\in [d].
 \end{equation}
\end{theorem}
\begin{IEEEproof}
Since $\cX$ is $s$-sparse, each vector in $R_j(\cX)$ is
$s$-sparse. If each $U_j$ satisfies the NSP$_s$, then
$\mathbf{w}_i^{(j)\star}\in R_j(\cX)$ is unique and $s$-sparse.
Define
 \begin{equation}\label{ranklikedecZ}
 \cZ=\sum_{i=1}^K \w_i^{(1)\star}\circ\ldots\circ \w_i^{(d)\star}, \quad \w_i^{(j)\star}\in R_j(\cX), j\in [d].
 \end{equation}
Then
\begin{equation}\label{inductioninitial}
(\cX-\cZ)\times_1 U_1\times\ldots \times_d U_d=0.
\end{equation}

To show $\cZ=\cX$, assume a slightly more general scenario, where
each $R_j(\cX)\subseteq \V_j \subset\R^{N_j}$, such that each
nonzero vector in $\V_j$ is $s$-sparse. Then $R_j(\cY)\subseteq
U_jR_j(\cX)\subseteq U_j\V_j$ for $j\in[d]$. Assume to the
contrary that $\cX\neq\cZ$. This hypothesis can be disproven via
induction on mode $m$ as follows.

Suppose
\begin{equation}\label{inductionm}
(\cX-\cZ)\times_m U_m\times\ldots\times_d U_d=0.
\end{equation}

Unfold $\cX$ and $\cZ$ in mode $m$, then the column (row) spaces
of $X_{(m)}$ and $Z_{(m)}$ are contained in $\V_m$
($\hat\V_m:=\V_1\circ\ldots\circ
\V_{m-1}\circ\V_{m+1}\circ\ldots\circ \V_d$). Since $\cX\neq\cZ$,
$X_{(m)}-Z_{(m)}\ne 0$. Then $X_{(m)}-Z_{(m)}=\sum_{i=1}^p \uu_
i\v_i\trans$, where $\rank (X_{(m)}-Z_{(m)})=p$, and
$\uu_1,\ldots,\uu_p\in\V_m, \v_1,\ldots,\v_p\in \hat \V_m$ are two
sets of linearly independent vectors.

Since $(\cX-\cZ)\times_m U_m\times\ldots\times_d U_d=0$,
\begin{align*}
0&=U_m(X_{(m)}-Z_{(m)})(U_d\otimes\ldots\otimes U_{m+1}\otimes I)\trans\\
&=U_m(X_{(m)}-Z_{(m)})\hat U_m\trans\\
&=\sum_{i=1}^p (U_m\uu_i)(\hat U_m\v_i)\trans.
\end{align*}

Since $U_m\uu_1,\ldots,U_m\uu_p$ are linearly independent, it
follows that $\hat U_m\v_i=0$ for $i\in [p]$. Therefore,
\[(X_{(m)}-Z_{(m)})\hat U_m\trans=(\sum_{i=1}^p \uu_i\v_i\trans)\hat U_m\trans = \sum_{i=1}^p \uu_i(\hat U_m\v_i)\trans =
0,\] which is equivalent to (in tensor form, after folding)
\begin{align}\label{inductionm1}
&(\cX-\cZ)\times_m I_m\times_{m+1} U_{m+1}\times\ldots
\times_d U_d\nonumber\\
&=(\cX-\cZ)\times_{m+1} U_{m+1}\times\ldots \times_d U_d=0,
\end{align}
where $I_m$ is the $N_m\times N_m$ identity matrix. Note that
\eqref{inductionm} leads to \eqref{inductionm1} upon replacing
$U_m$ with $I_m$. Similarly, when $m=1$, $U_1$ can be replaced
with $I_1$ in \eqref{inductioninitial}. By successively replacing
$U_m$ with $I_m$ for $2\leq m\leq d$,
 \begin{align*}
&(\cX-\cZ)\times_1 U_1\times\ldots \times_d U_d\\
=&(\cX-\cZ)\times_1 I_1\times\ldots \times_d I_d\\
=&\cX-\cZ=0, \label{eqstepdm1}
\end{align*}
which contradicts the assumption that $\cX\neq\cZ$. Thus,
$\cX=\cZ$. This completes the proof.
\end{IEEEproof}

Note that although Theorem \ref{TheoremGTCSP} requires $U_i$ to
satisfy the NSP$_s$ property for $i\in [d]$, such constraints can
be relaxed if all vectors $\in R_i(\cX)$ are $s_i$-sparse. In this
case, it follows from the proof of Theorem \ref{TheoremGTCSP} that
$X$ can be recovered as long as $U_i$ satisfies the NSP$_{s_i}$,
for $i\in [d]$. As in the matrix case, the reconstruction stage of
the recovery process can be implemented in parallel for each
tensor mode.

The above described procedure allows exact recovery. In some
cases, recovery of a rank-$R$ approximation of $\cX$,
$\mathcal{\hat{X}}=\sum_{r=1}^R\w_r^{(1)}\circ \ldots\circ
\w_r^{(d)}$, suffices. In such scenarios, $\cY$ in
\eqref{ranklikedecY} can be replaced by its rank-$R$
approximation, namely, $\cY=\sum_{r=1}^R\mathbf{b}_r^{(1)}\circ
\ldots\circ \mathbf{b}_r^{(d)}$ (obtained e.g., \ by CP
decomposition).
\vspace{-0.3cm}
\subsection{GTCS reconstruction with the presence of noise}
We next briefly discuss the case where the observation is noisy.
We state informally here two theorems for matrix case. For a
detailed proof of the theorems as well as the generalization to
tensor case, please refer to \cite{BookChapter}. Assume that the
notations and the assumptions of Theorem \ref{compsensmatrev}
hold. Let $Y'=Y+E = U_1 X U_2\trans +E, \quad Y \in \R^{m_1\times
m_2}.$ Here $E$ denotes the noise matrix, and $\|E\|_F\le
\varepsilon$ for some real nonnegative number $\varepsilon$.

Our first recovery result is as follows: assume that that each
$U_i$ satisfies the RIP$_{2s}$ property for some $\delta_{2s}\in
(0, \sqrt{2}-1)$. Then $X$ can be recovered uniquely as follows.
Let $\c_1(Y'),\ldots,\c_{m_2}(Y')\in\R^{m_1}$ denote the columns
of $Y'$. Let $\z_i^\star\in\R^{N_1}$ be a solution of
\begin{equation}\label{defzstar}
\z_i^\star=\min\{\|\z_i\|_1, \|\c_i(Y') - U_1\z_i\|_2\le
\varepsilon\}, \quad i\in [m_2].
\end{equation}
Define $Z\in\R^{N_1\times m_2}$ to be the matrix whose columns are
$ \z_1^\star,\ldots,\z_{m_2}^\star$. According to
\eqref{sigrecer1},
$\|\c_i(Z)-\c_i(XU_2\trans)\|_2=\|\z_i^\star-\c_i(XU_2\trans)\|_2\le
C_2\varepsilon$. Hence $\|Z-XU_2\trans\|_F\le
\sqrt{m_2}C_2\varepsilon$. Let $\c_1(Z\trans),\ldots,
\c_{N_1}(Z\trans)$ be the columns of $Z\trans$. Then
$\uu_j^\star\in\R^{N_2}$, the solution of
\begin{equation}\label{defustar}
\uu_j^\star=\min\{\|\uu_j\|_1 , \|\c_j(Z\trans)-U_2 \uu_j\|_2 \le
\sqrt{m_2}C_2\varepsilon\}, \quad j\in [N_1],
\end{equation}
is the $j^{th}$ column of $X\trans$. Denote by $\hat X$ the
recovered matrix, then according to \eqref{sigrecer1},
\begin{equation}\label{MatrixBound}
\|\hat X- X\|_F\le \sqrt{m_2N_1}C_2^2\varepsilon.
\end{equation}

The upper bound in \eqref{MatrixBound} can be tightened by
assuming that the entries of $E$ adhere to a specific type of
distribution. Let $E=[\e_1,\ldots,\e_{m_2}]$. Suppose that each
entry of $E$ is an independent random variable with a given
distribution having zero mean.  Then we can assume that
$\|\e_j\|_2\le\frac{\varepsilon}{\sqrt{m_2}}$, which implies that
$\|E\|_F\le \varepsilon$. In such case, $\|\hat X- X\|_F\le
C_2^2\varepsilon.$

Our second recovery result is as follows: assume that $U_i$
satisfies the RIP$_{2s}$ property for some $\delta_{2s}\in (0,
\sqrt{2}-1)$, $i\in [2]$. Then $X$ can be recovered uniquely as
follows. Assume that $s'$ is the smallest between $s$ and the
number of singular values of $Y'$ greater than
$\frac{\varepsilon}{\sqrt{s}}$. Let $Y'_{s'}$ be a best rank-$s'$
approximation of $Y'$:
\begin{equation}\label{SVDkaproxY}
Y'_{s'}=\sum_{i=1}^{s'} (\sqrt{\tilde\sigma_i}\tilde
\uu_i)(\sqrt{\tilde \sigma_i}\tilde \v_i)\trans.
\end{equation} Then $\hat X = \sum_{i=1}^{s'}
\frac{1}{\tilde\sigma_i} \x_i^\star {\y_i^\star}\trans$ and
\begin{equation}\label{esterrMS}
\|X-\hat X\|_F\le C^2\varepsilon,
\end{equation}

where
\begin{eqnarray*}\label{hatxrecov}
&&\x_i^\star=\min\{ \|\x_i\|_1, \|\tilde\sigma_i\tilde \uu_i - U_1\x_i\|_2 \le \frac{\varepsilon}{\sqrt{2s}}\},\\
\label{hatyrecov} &&\y_i^\star=\min\{ \|\y_i\|_1,
\|\tilde\sigma_i\tilde \v_i - U_2\y_i\|_2 \le
\frac{\varepsilon}{\sqrt{2s}}\},i\in [s].
\end{eqnarray*}

\section{Experimental Results}\label{experiments}
We experimentally demonstrate the performance of GTCS methods on
the reconstruction of sparse and compressible images and video
sequences. As demonstrated in \cite{kroneckerCS}, KCS outperforms
several other methods including independent measurements and
partitioned measurements in terms of reconstruction accuracy in
tasks related to compression of multidimensional signals. A more
recently proposed method is MWCS, which stands out for its
reconstruction efficiency. For the above reasons, we compare our
methods with both KCS and MWCS. Our experiments use the
$\ell_1$-minimization solvers from \cite{l1}. We set the same
threshold to determine the termination of $\ell_1$-minimization in
all subsequent experiments. All simulations are executed on a
desktop with 2.4 GHz Intel Core i5 CPU and 8GB RAM.

\subsection{Sparse image representation}
\begin{figure}[htb]
\begin{center}
\subfigure[The original sparse
image]{\label{Originallogo}\includegraphics[width=0.45\linewidth]{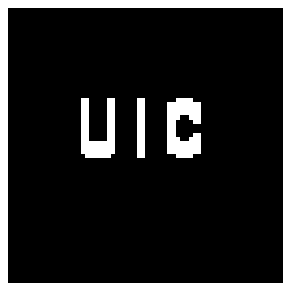}}
\subfigure[GTCS-S recovered
image]{\label{GTCSSlogo}\includegraphics[width=0.45\linewidth]{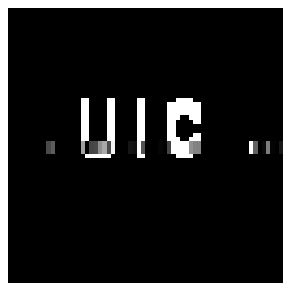}}\\\vspace{-.3cm}
\subfigure[GTCS-P recovered
image]{\label{GTCSPlogo}\includegraphics[width=0.45\linewidth]{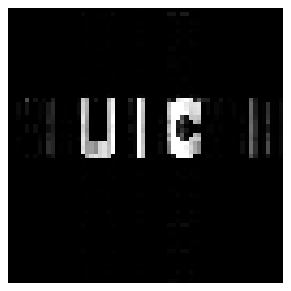}}
\subfigure[KCS recovered
image]{\label{KCSlogo}\includegraphics[width=0.45\linewidth]{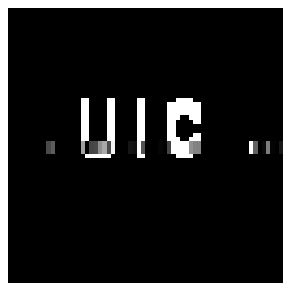}}\vspace{-.3cm}
\caption{The original image and the recovered images by GTCS-S
(PSNR = 22.28 dB), GTCS-P (PSNR = 23.26 dB) and KCS (PSNR = 22.28
dB) when $m = 38$, using 0.35 normalized number of samples.}
\end{center}
\end{figure}

As shown in Figure \ref{Originallogo}, the original black and
white image is of size $64\times 64$ ($N=4096$ pixels). Its
columns are $14$-sparse and rows are $18$-sparse. The image itself
is $178$-sparse. We let the number of measurements evenly split
among the two modes, that is, for each mode, the randomly
constructed Gaussian matrix $U$ is of size $m\times 64$. Therefore
the KCS measurement matrix $U\otimes U$ is of size $m^2\times
4096$. Thus the total number of samples is $m^2$. We define the
normalized number of samples to be $\frac{m^2}{N}$. For GTCS, both
the serial recovery method GTCS-S and the parallelizable recovery
method GTCS-P are implemented. In the matrix case, GTCS-P
coincides with MWCS and we simply conduct SVD on the compressed
image in the decomposition stage of GTCS-P. Although the
reconstruction stage of GTCS-P is parallelizable, we here recover
each vector in series. We comprehensively examine the performance
of all the above methods by varying $m$ from 1 to 45.

\begin{figure}[htb]
\begin{center}
\subfigure[PSNR
comparison]{\label{UICPSNR}\includegraphics[width=0.45\linewidth]{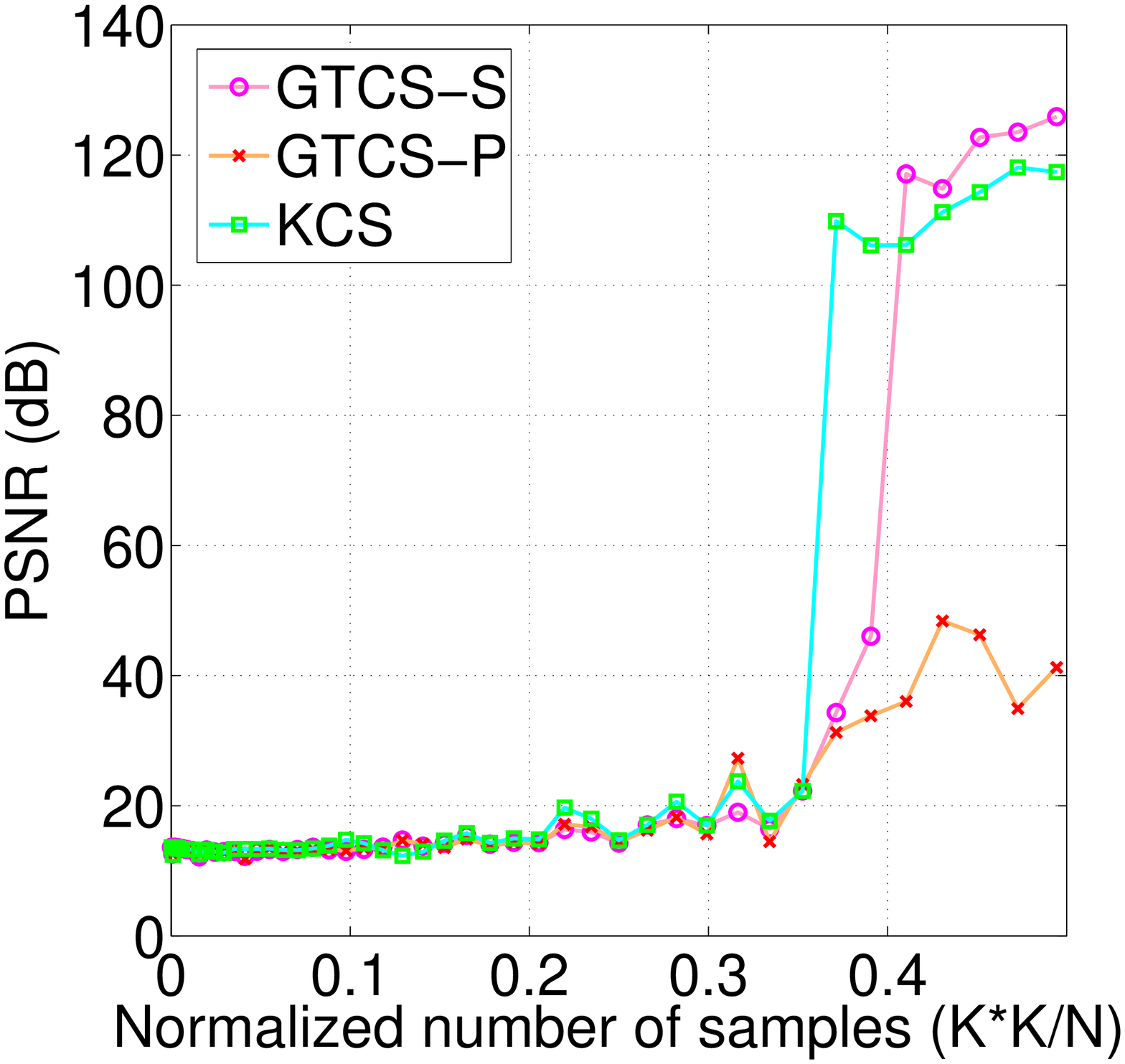}}
\subfigure[Recovery time
comparison]{\label{UICTime}\includegraphics[width=0.45\linewidth]{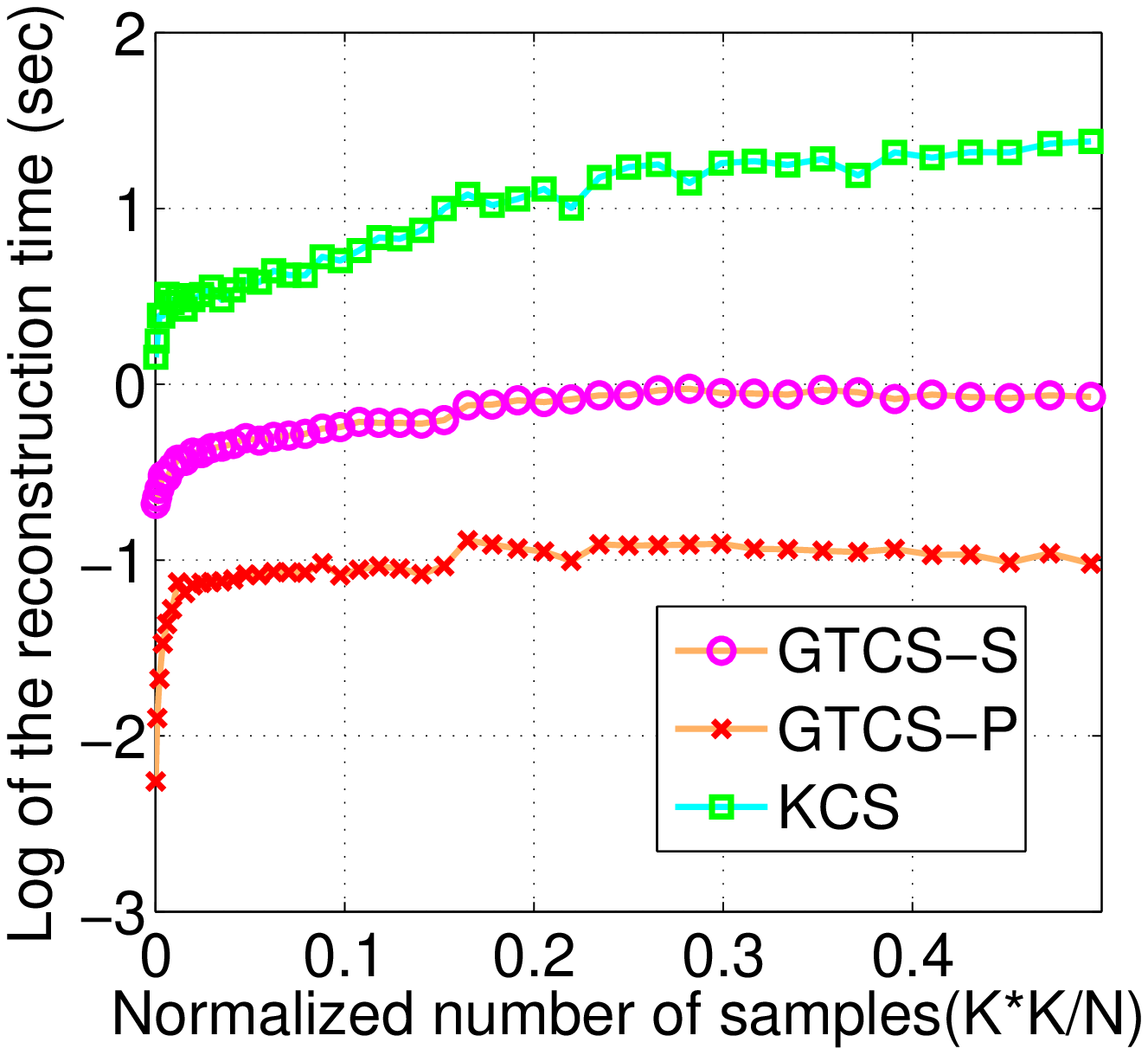}}
\caption{PSNR and reconstruction time comparison on sparse image.}
\label{3figs}
\end{center}
\end{figure}

Figure \ref{UICPSNR} and \ref{UICTime} compare the peak signal to
noise ratio (PSNR) and the recovery time respectively. Both KCS
and GTCS methods achieve PSNR over 30dB when $m=39$. Note that at
this turning point, PSNR of KCS is higher than that of GTCS, which
is consistent with the observation that GTCS usually requires
slightly more number of measurements to achieve the same
reconstruction accuracy in comparison with KCS. As $m$ increases,
GTCS-S tends to outperform KCS in terms of both accuracy and
efficiency. Although PSNR of GTCS-P is the lowest among the three
methods, it is most time efficient. Moreover, with parallelization
of GTCS-P, the recovery procedure can be further accelerated
considerably. The reconstructed images when $m=38$, that is, using
0.35 normalized number of samples, are shown in Figure
\ref{GTCSSlogo}, \ref{GTCSPlogo}, and \ref{KCSlogo}. Though GTCS-P
usually recovers much noisier image, it is good at recovering the
non-zero structure of the original image.

\subsection{Compressible image representation}
\begin{figure}[htb]
\begin{center}
\subfigure[Cameraman in space
domain]{\label{CMspace}\includegraphics[width=0.45\linewidth]{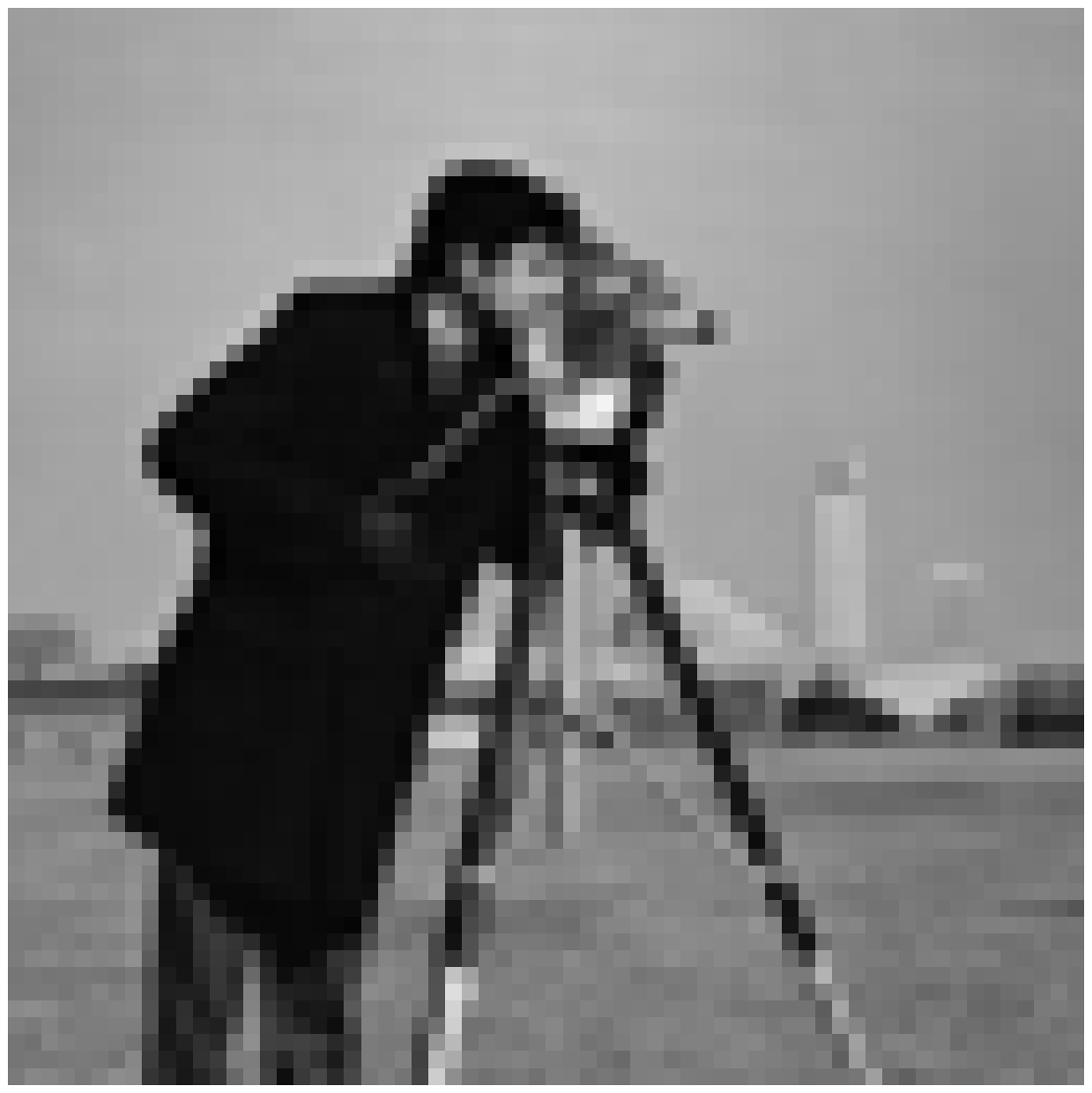}}
\subfigure[Cameraman in DCT
domain]{\label{CMDCT}\includegraphics[width=0.45\linewidth]{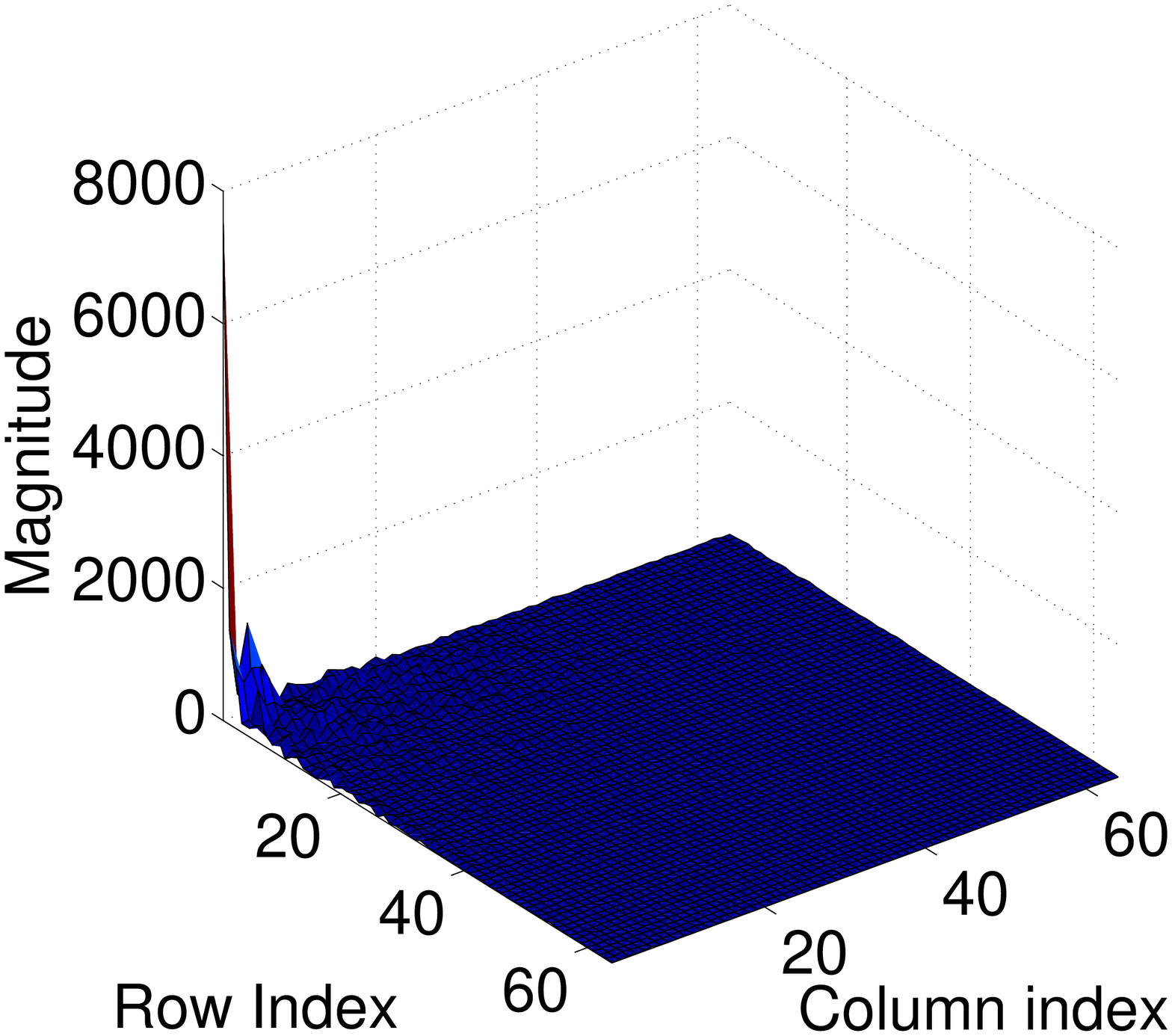}}
\caption{The original cameraman image (resized to 64 $\times$ 64
pixels) in the space domain (a) and the DCT domain (b).}
\end{center}
\end{figure}

As shown in Figure \ref{CMspace}, the cameraman image is resized
to $64\times 64$ ($N=4096$ pixels). The image itself is
non-sparse. However, in some transformed domain, such as discrete
cosine transformation (DCT) domain in this case, the magnitudes of
the coefficients decay by power law in both directions (see Figure
\ref{CMDCT}), thus are compressible. We let the number of
measurements evenly split among the two modes. Again, in matrix
data case, MWCS concurs with GTCS-P. We exhaustively vary $m$ from
1 to 64.

\begin{figure}[htb]
\begin{center}
\subfigure[PSNR
comparison]{\label{CMPSNR}\includegraphics[width=0.45\linewidth]{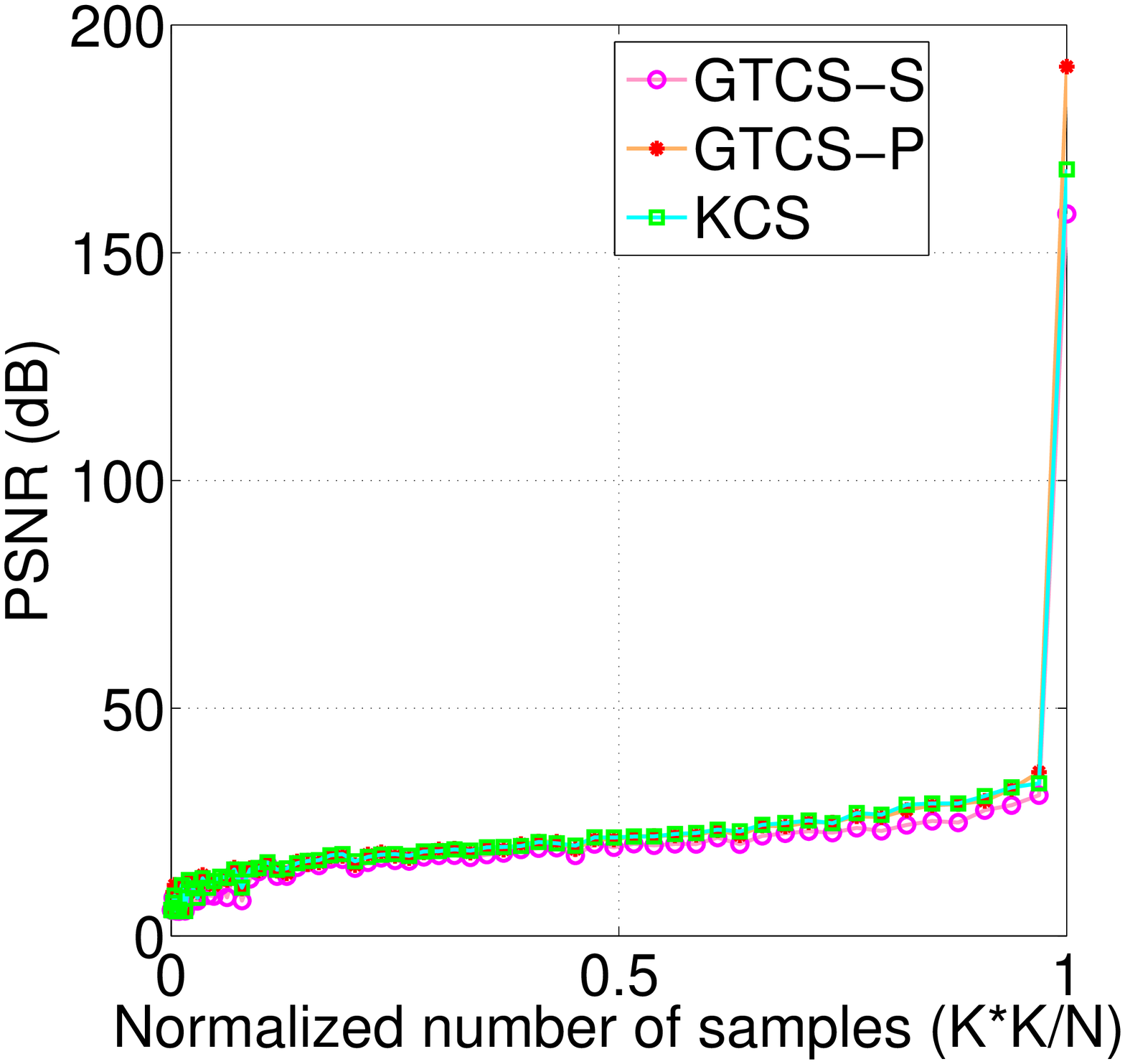}}
\subfigure[Recovery time
comparison]{\label{CmTime}\includegraphics[width=0.45\linewidth]{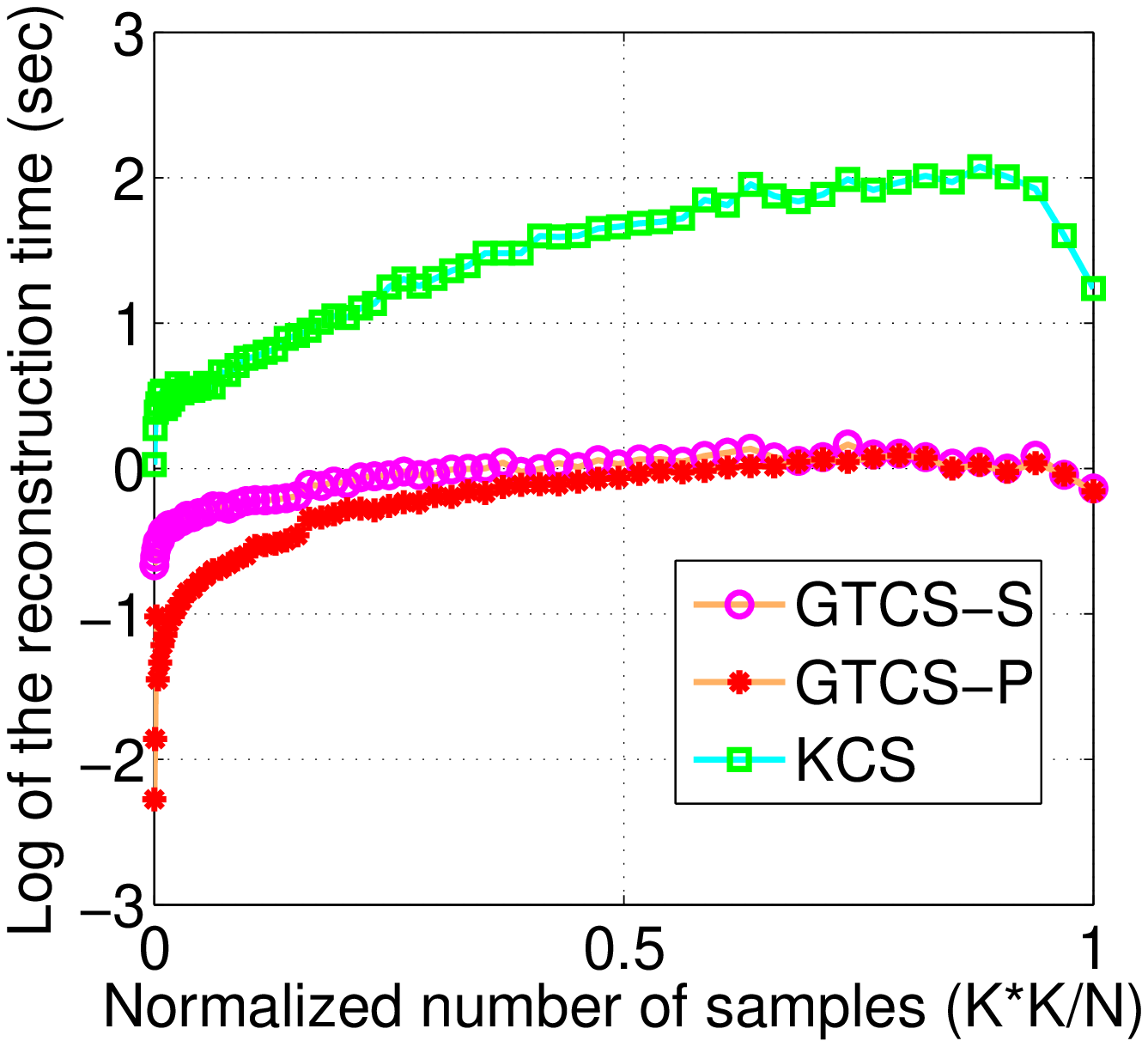}}
\caption{PSNR and reconstruction time comparison on compressible
image.} \label{3figs}
\end{center}
\end{figure}

\begin{figure}[htb]
\begin{center}
\subfigure[GTCS-S, m= 46, PSNR = 20.21
dB]{\includegraphics[width=0.45\linewidth]{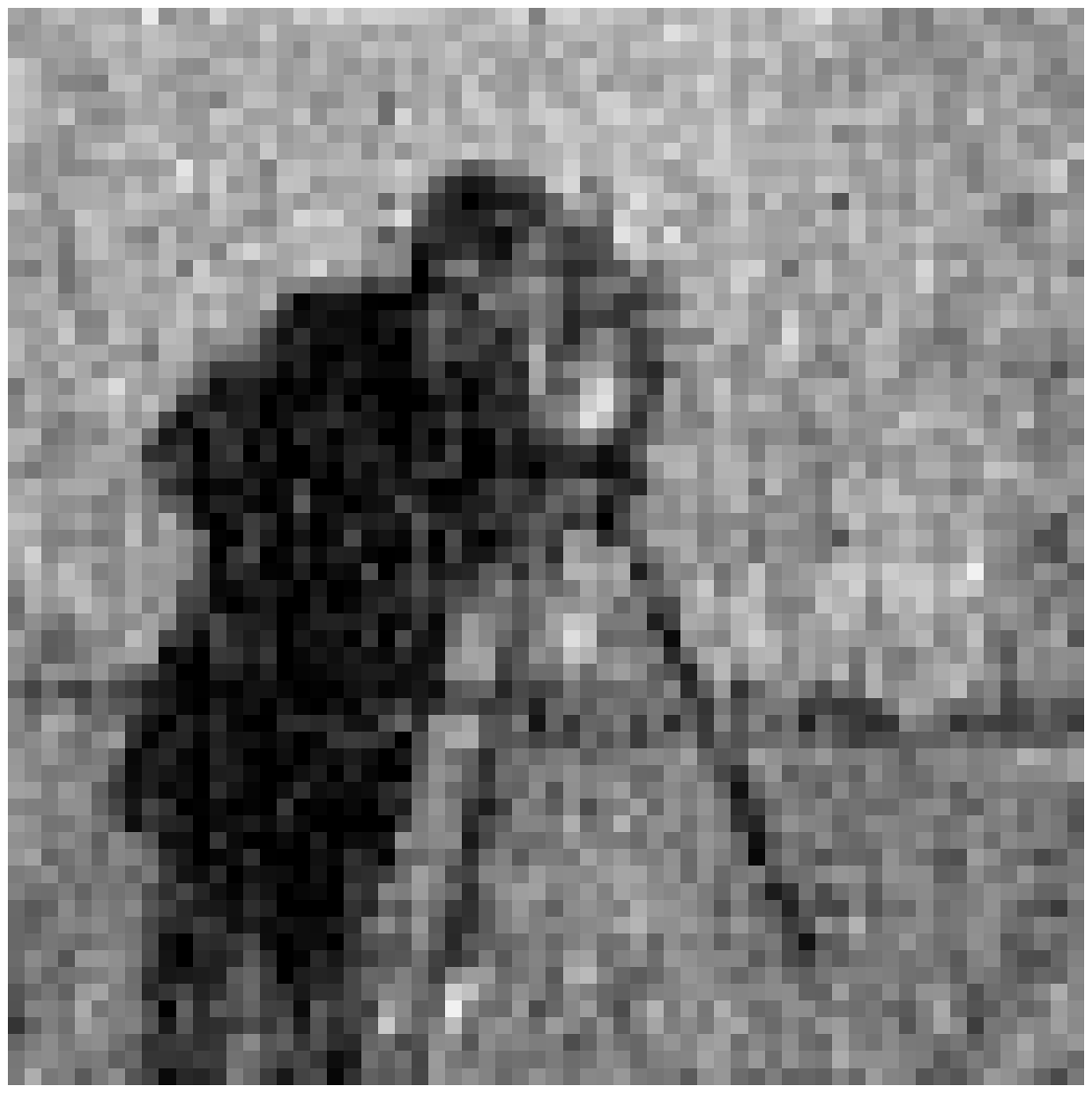}}
\subfigure[GTCS-S, m = 63, PSNR = 30.88
dB]{\includegraphics[width=0.45\linewidth]{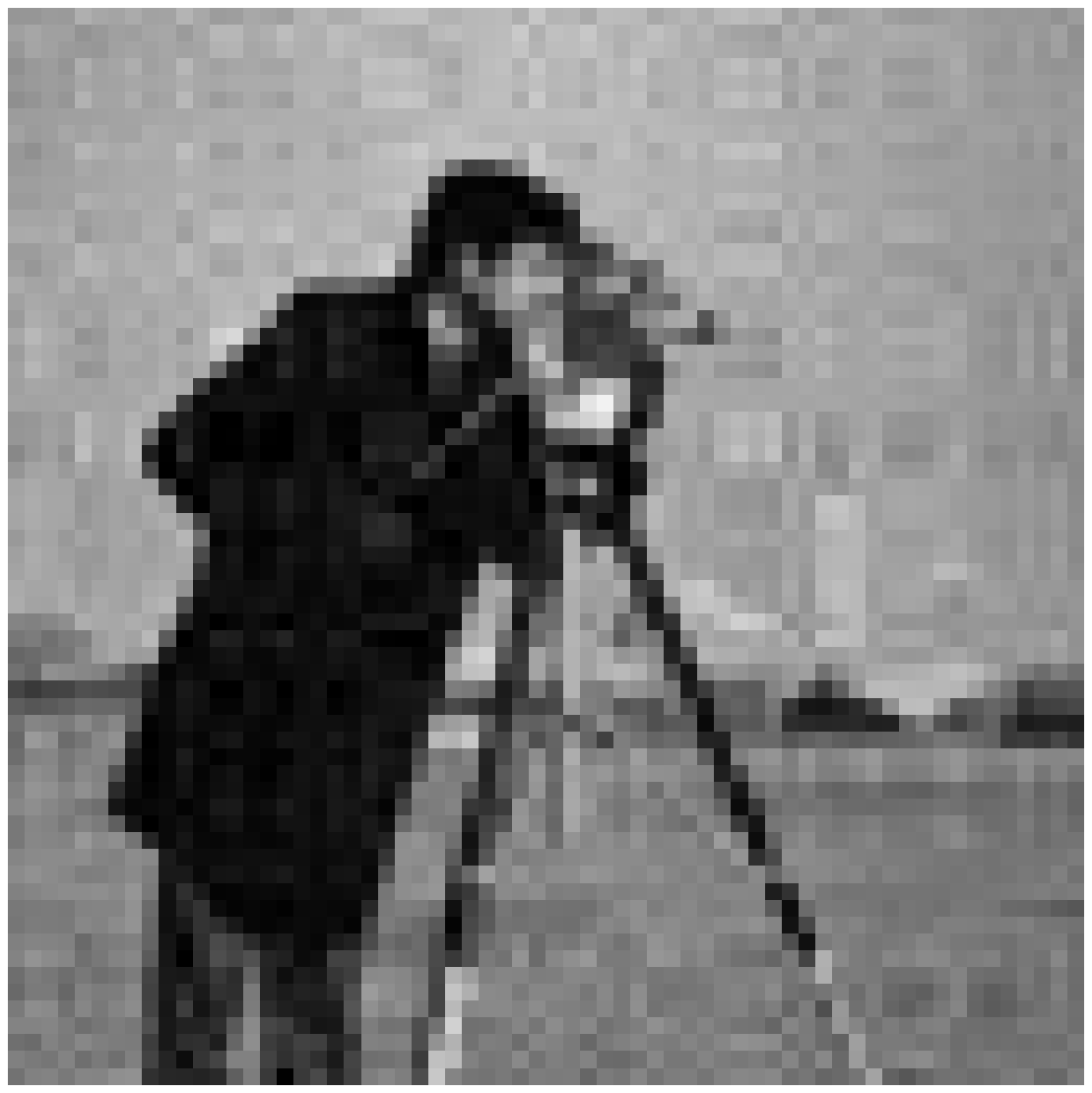}}\\\vspace{-.3cm}
\subfigure[GTCS-P/MWCS, m = 46, PSNR = 21.84
dB]{\includegraphics[width=0.45\linewidth]{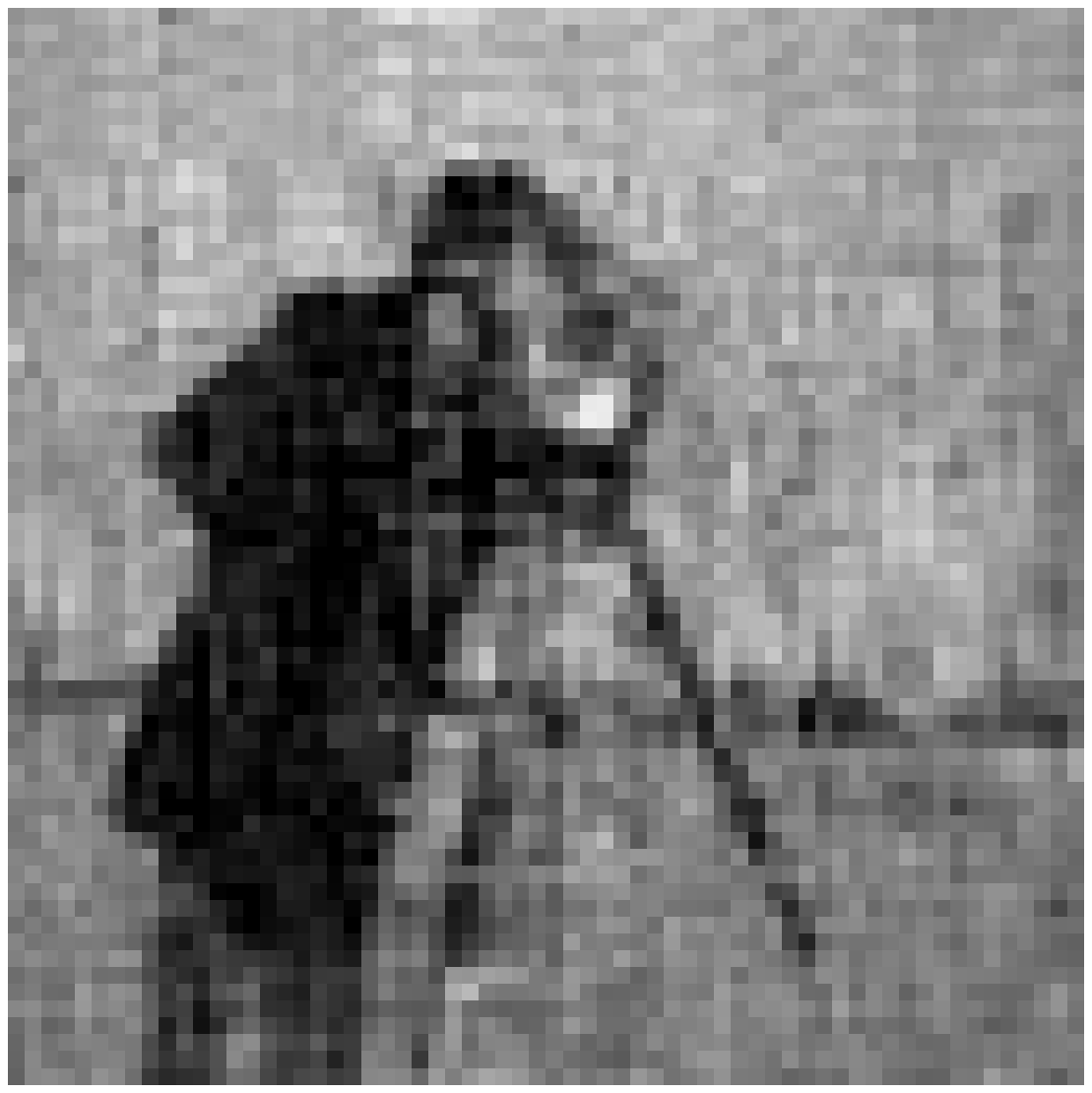}}
\subfigure[GTCS-P/MWCS, m = 63, PSNR = 35.95
dB]{\includegraphics[width=0.45\linewidth]{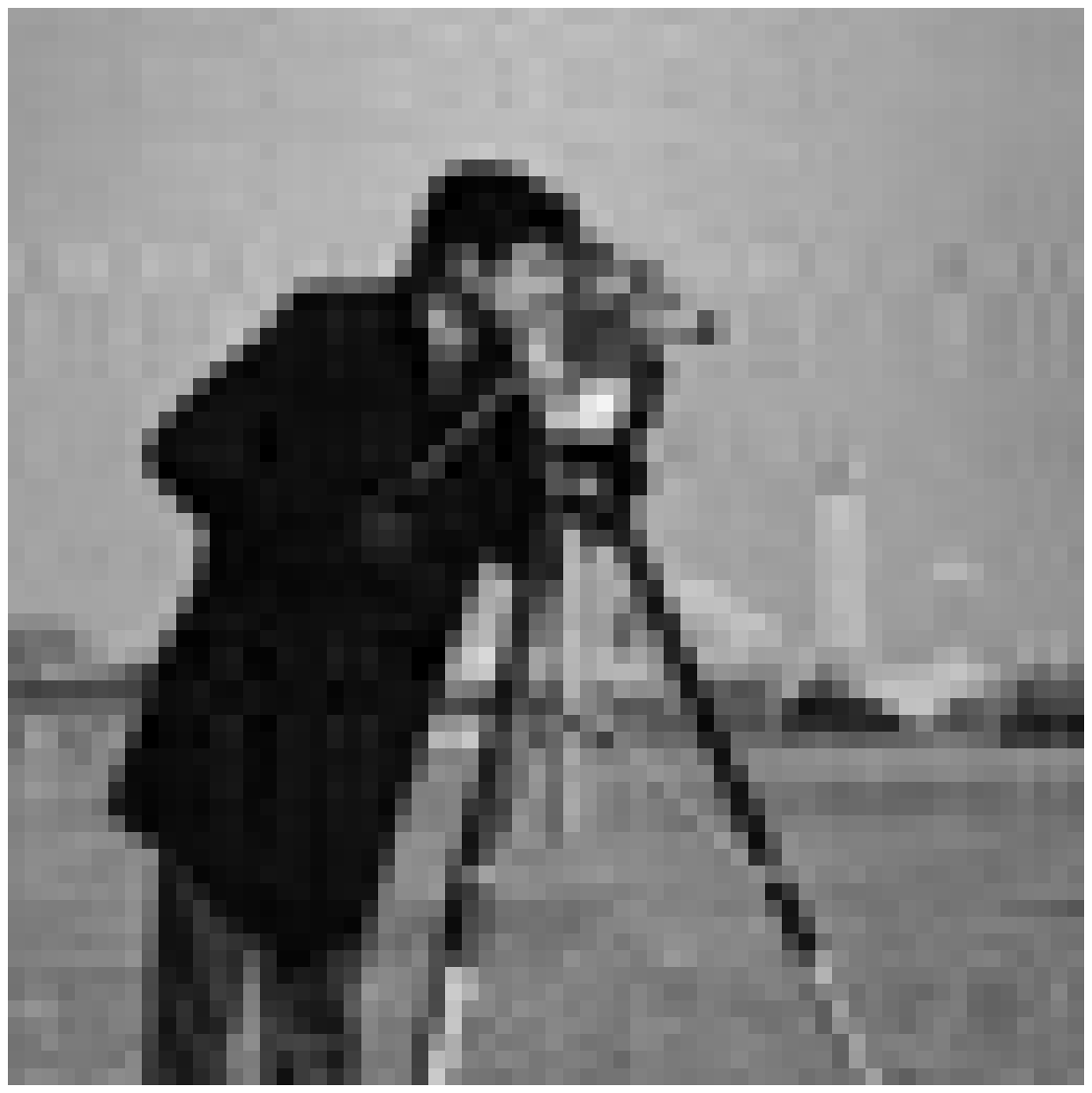}}\\\vspace{-.3cm}
\subfigure[KCS, m = 46, PSNR = 21.79
dB]{\includegraphics[width=0.45\linewidth]{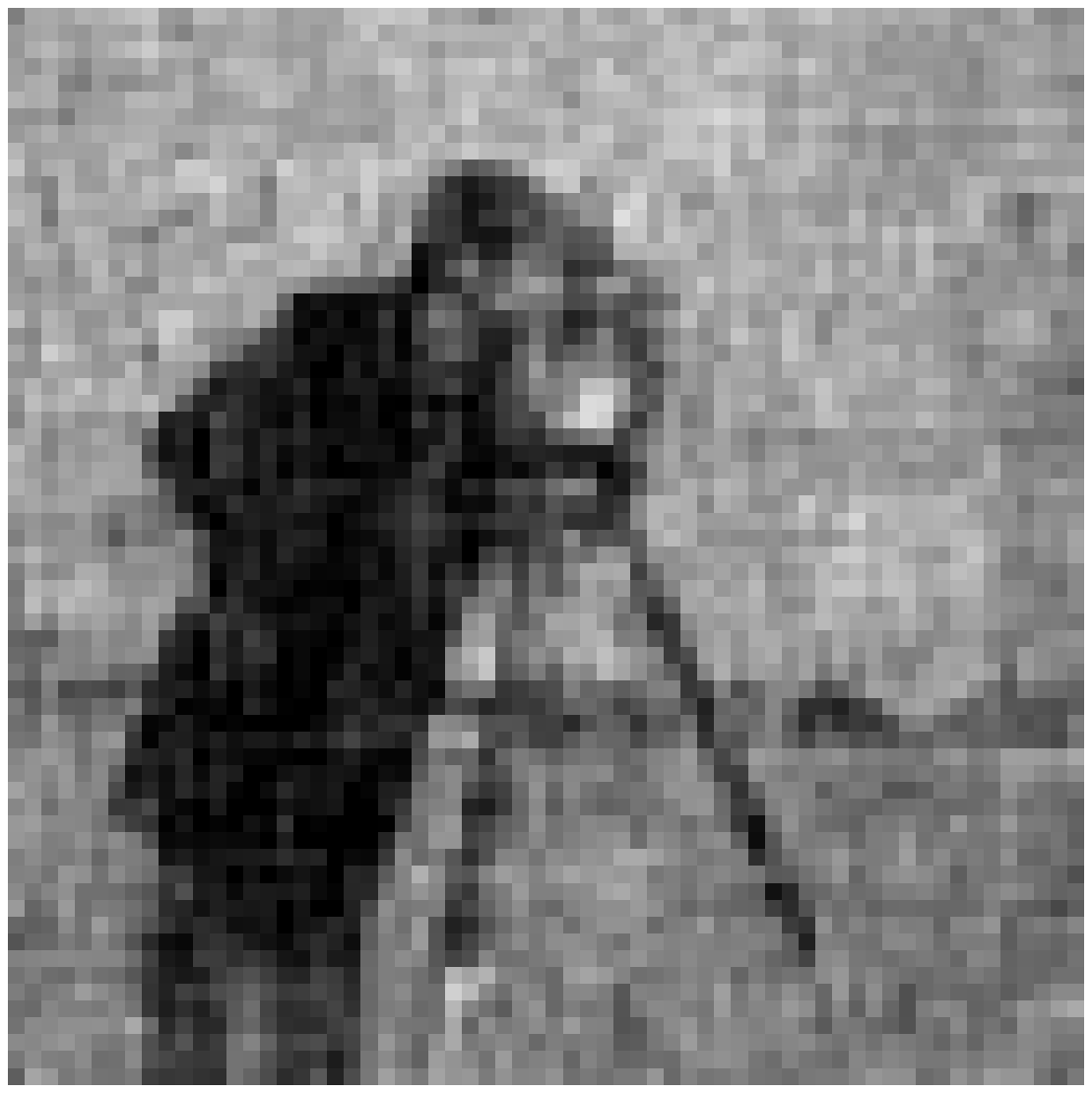}}
\subfigure[KCS, m = 63, PSNR = 33.46
dB]{\includegraphics[width=0.45\linewidth]{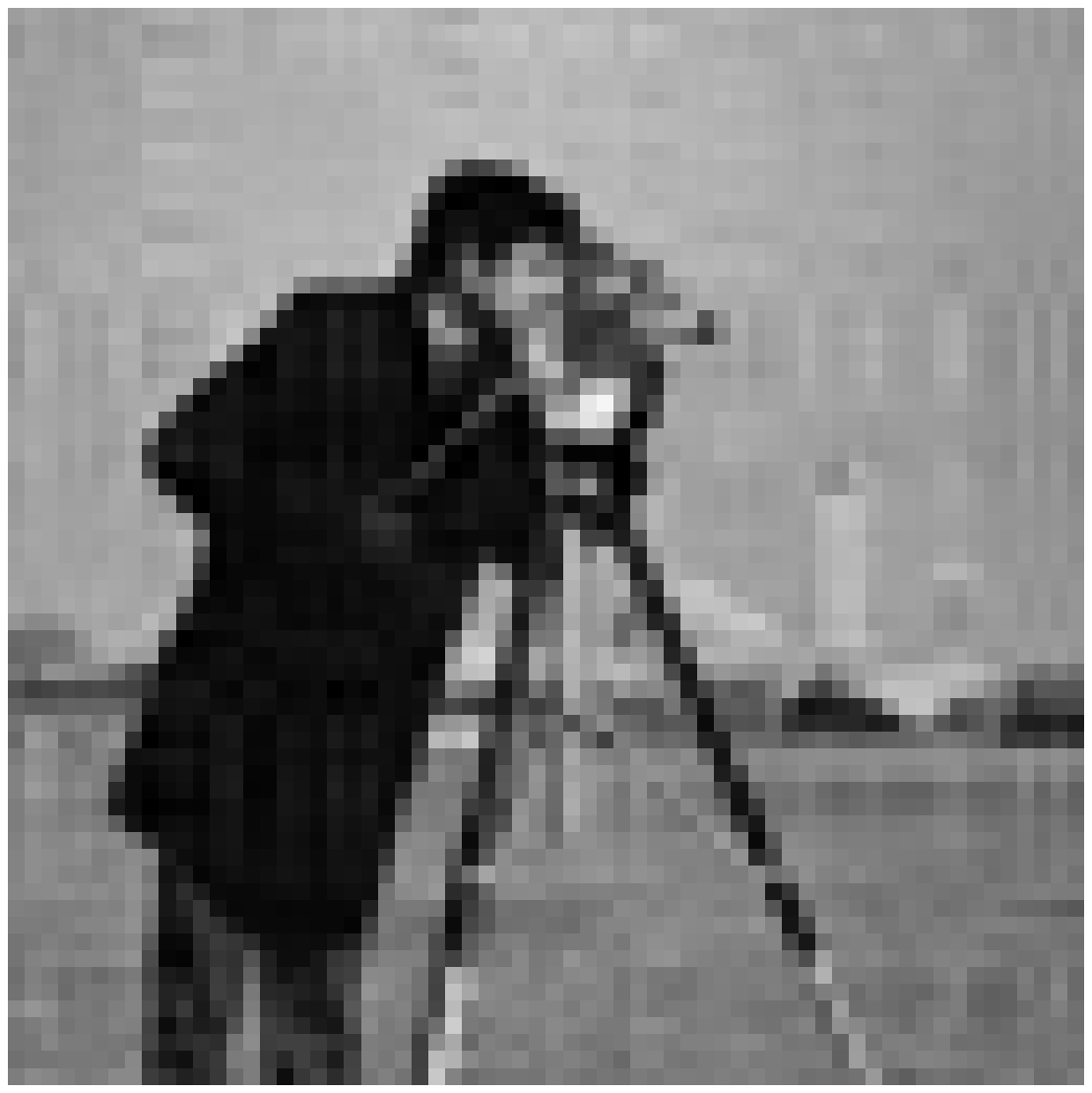}}\\\vspace{-.3cm}
\caption{Reconstructed cameraman images. In this two-dimensional
case, GTCS-P is equivalent to MWCS.}\label{CMresultsshow}
\end{center}
\end{figure}

Figure \ref{CMPSNR} and \ref{CmTime} compare the PSNR and the
recovery time respectively. Unlike the sparse image case, GTCS-P
shows outstanding performance in comparison with all other
methods, in terms of both accuracy and speed, followed by KCS and
then GTCS-S. The reconstructed images when $m=46$, using 0.51
normalized number of samples and when $m=63$, using 0.96
normalized number of samples are shown in Figure
\ref{CMresultsshow}.
\subsection{Sparse video representation}
We next compare the performance of GTCS and KCS on video data.
Each frame of the video sequence is preprocessed to have size
$24\times 24$ and we choose the first 24 frames. The video data
together is represented by a $24\times 24\times 24$ tensor and has
$N=13824$ voxels in total. To obtain a sparse tensor, we manually
keep only $6\times 6\times 6$ nonzero entries in the center of the
video tensor data and the rest are set to zero. Therefore, the
video tensor itself is $216$-sparse and its mode-$i$ fibers are
all $6$-sparse for $i\in [3].$ The randomly constructed Gaussian
measurement matrix for each mode is now of size $m\times 24$ and
the total number of samples is $m^3$. Therefore, the normalized
number of samples is $\frac{m^3}{N}$. In the decomposition stage
of GTCS-P, we employ a decomposition described in Section
\ref{mareview} to obtain a weaker form of the core Tucker
decomposition. We vary $m$ from 1 to 13.

\begin{figure}[htb]
\begin{center}
\subfigure[PSNR
comparison]{\label{SusiePSNR}\includegraphics[width=0.45\linewidth]{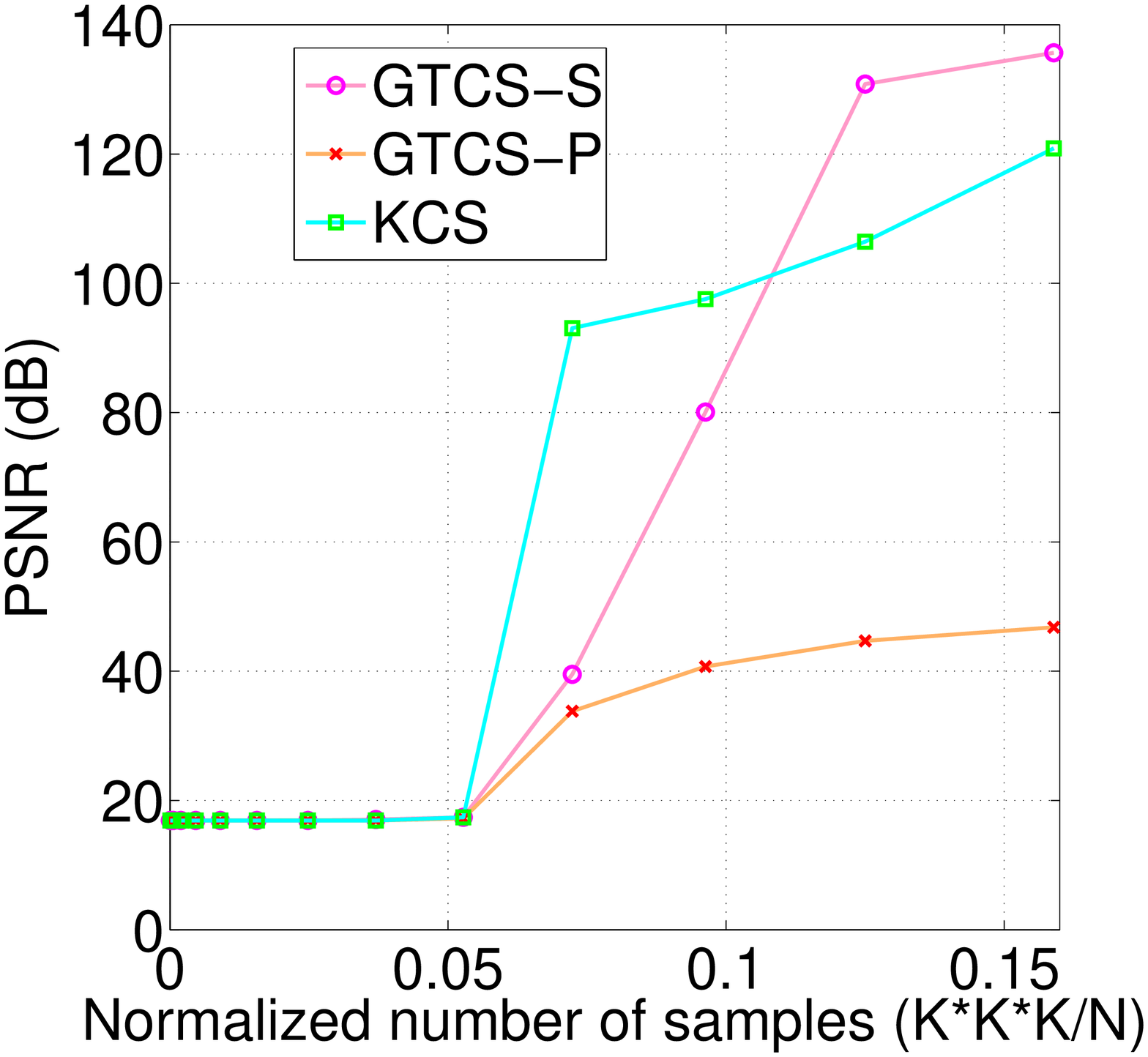}}
\subfigure[Recovery time
comparison]{\label{SusieTime}\includegraphics[width=0.45\linewidth]{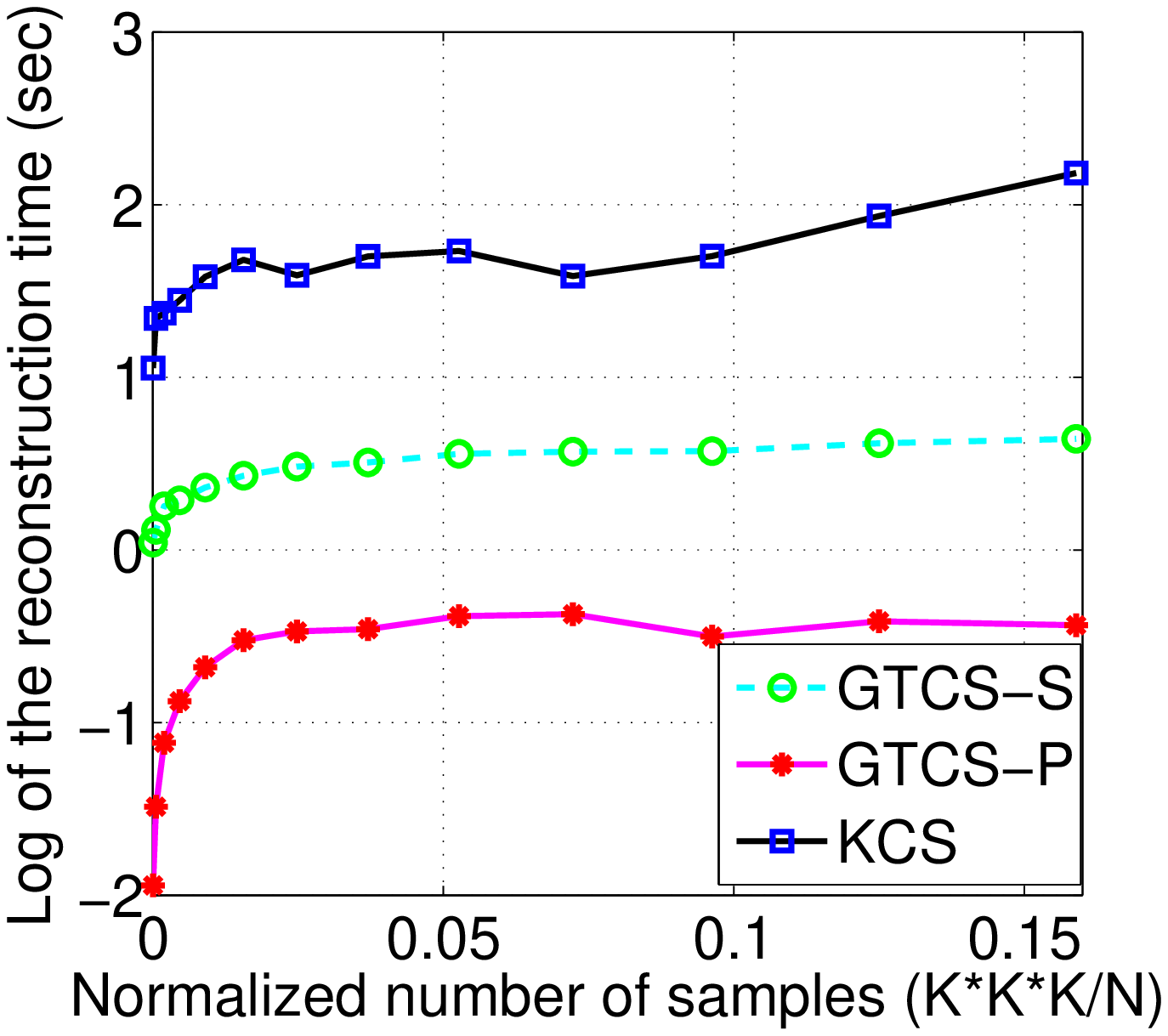}}
\caption{PSNR and reconstruction time comparison on sparse video.}
\label{3figs}
\end{center}
\end{figure}

\begin{figure*}[htb]
\begin{center}
\subfigure[Reconstruction error of
GTCS-S]{\label{GTCSSerr}\includegraphics[width=0.25\linewidth]{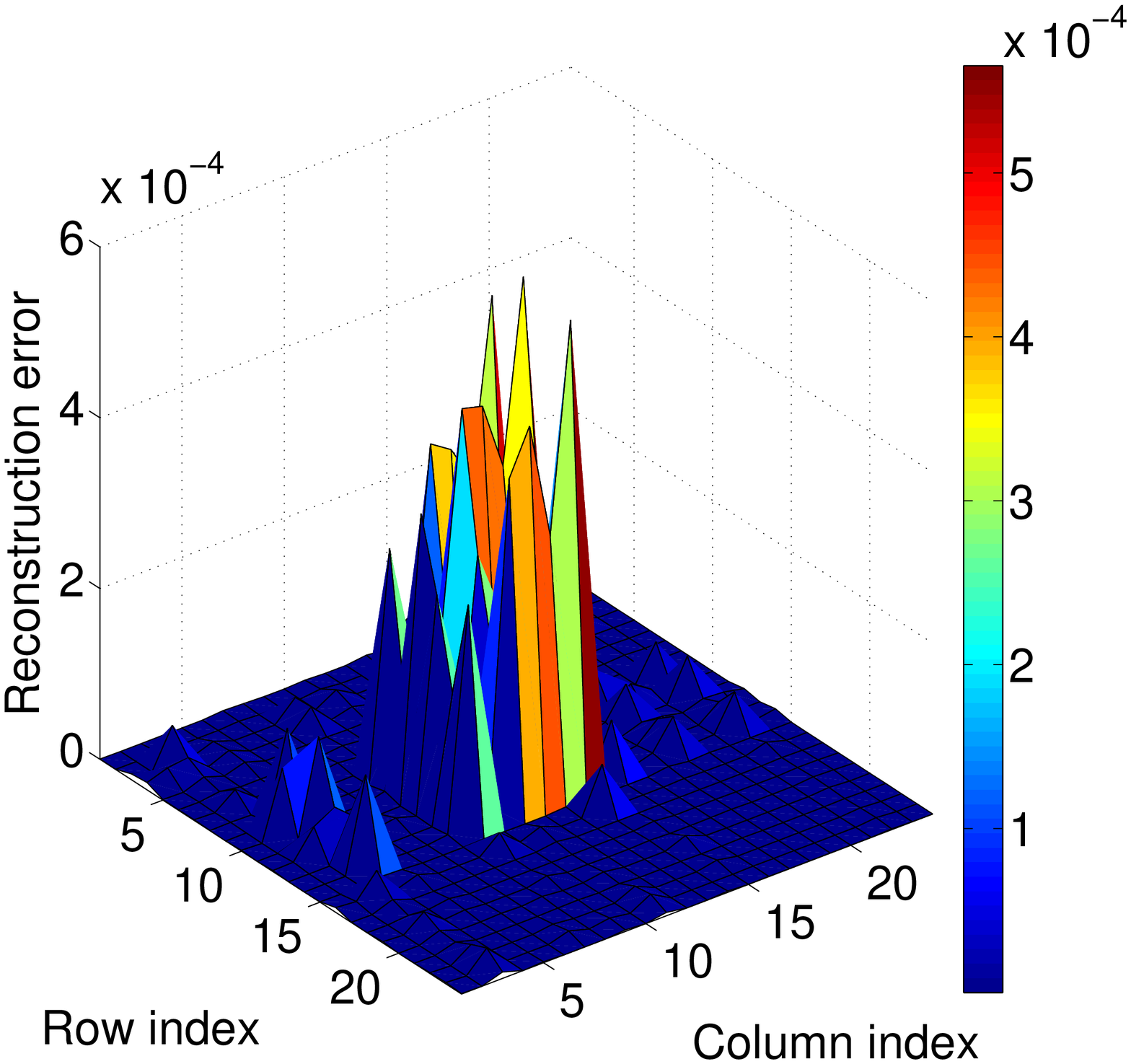}}
\subfigure[Reconstruction error of
GTCS-P]{\label{GTCSPerr}\includegraphics[width=0.25\linewidth]{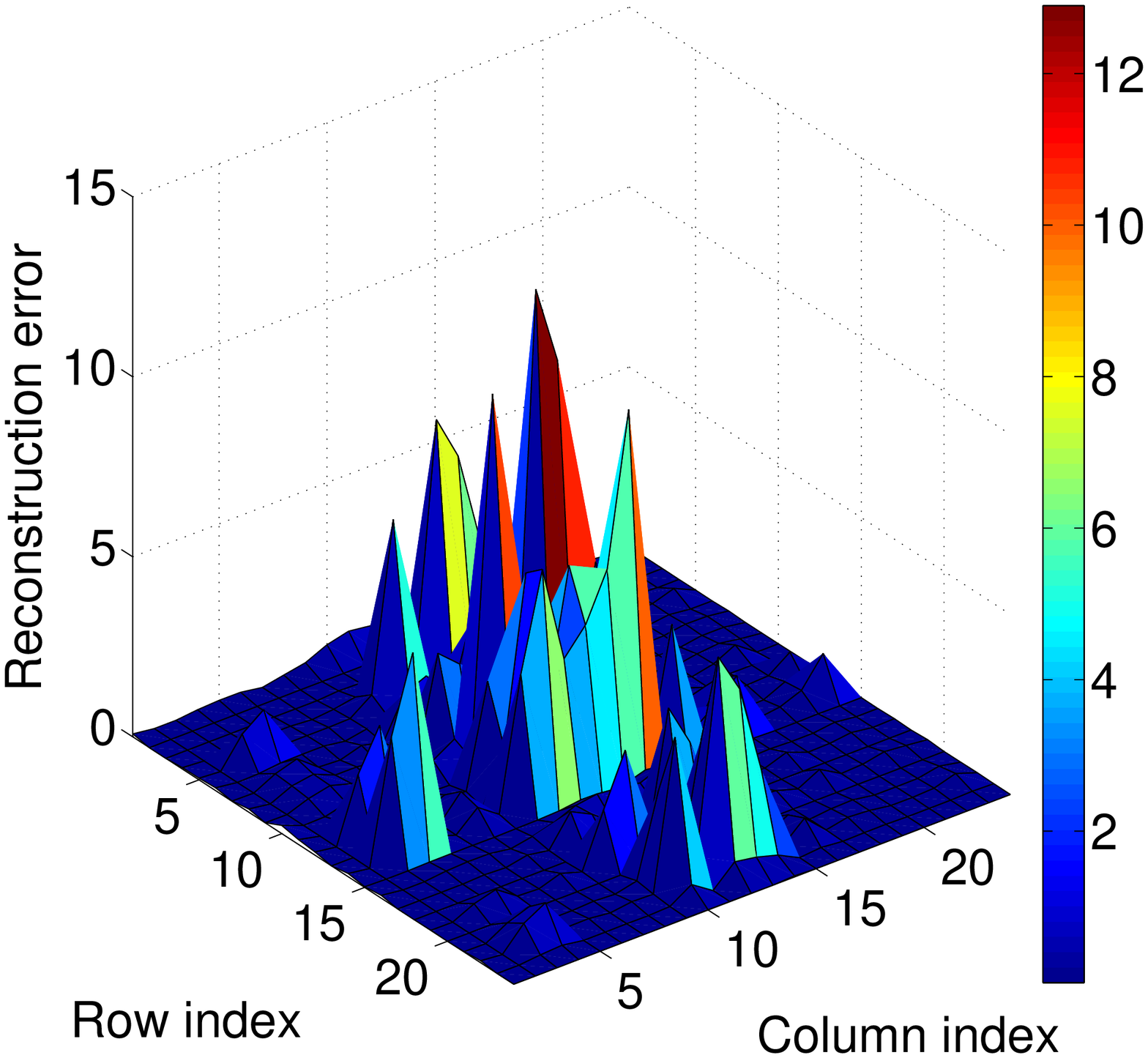}}
\subfigure[Reconstruction error of
KCS]{\label{KCSerr}\includegraphics[width=0.25\linewidth]{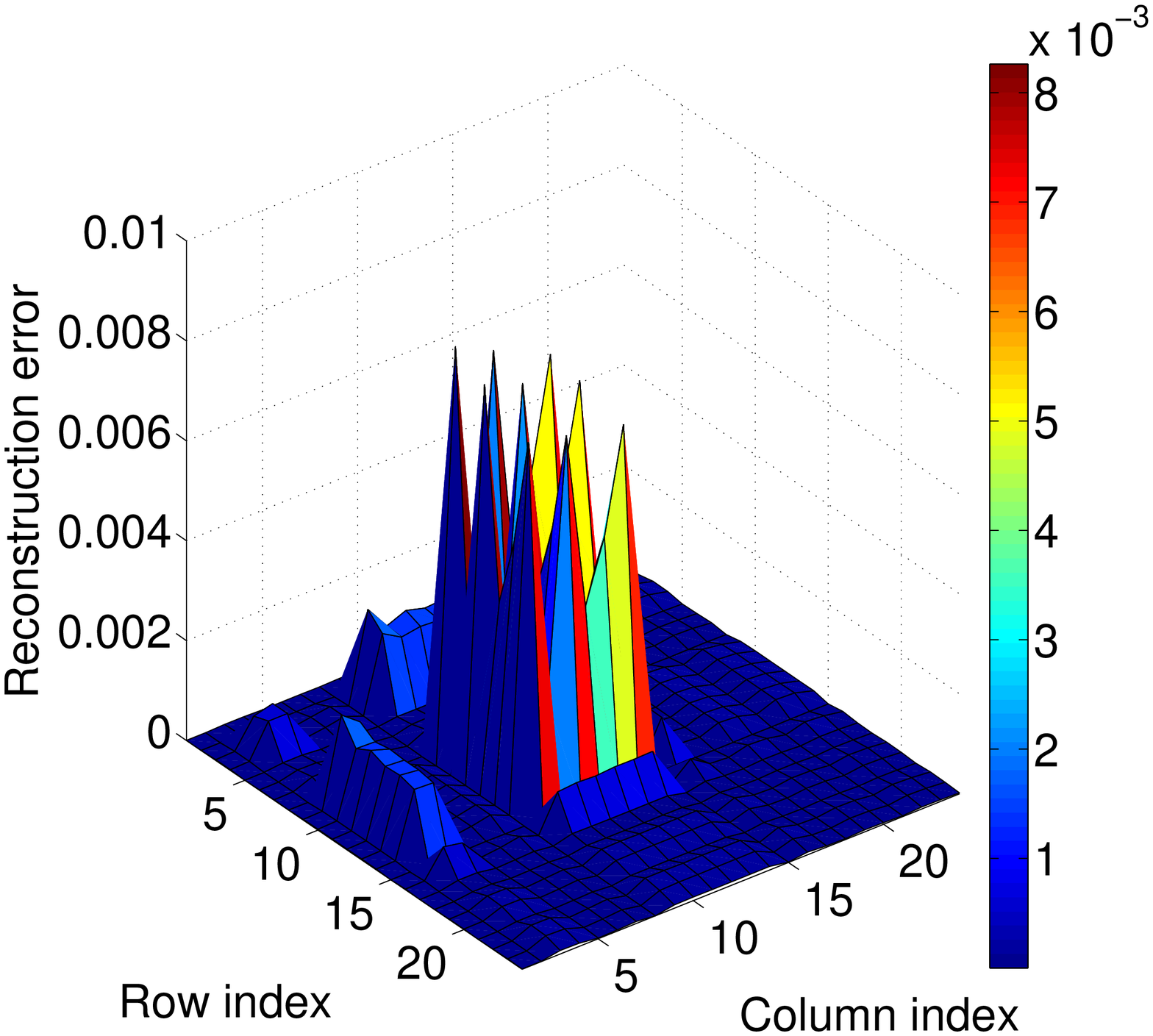}}
\caption{Visualization of the reconstruction error in the
recovered video frame 9 by GTCS-S (PSNR = 130.83 dB), GTCS-P (PSNR
= 44.69 dB) and KCS (PSNR = 106.43 dB) when $m=12$, using 0.125
normalized number of samples.}
\end{center}
\end{figure*}

Figure \ref{SusiePSNR} depicts PSNR of the first non-zero frame
recovered by all three methods. Please note that the PSNR values
of different video frames recovered by the same method are the
same. All methods exhibit an abrupt increase in PSNR at $m=10$
(using 0.07 normalized number of samples). Also, Figure
\ref{SusieTime} summarizes the recovery time. In comparison to the
image case, the time advantage of GTCS becomes more important in
the reconstruction of higher-order tensor data.
We specifically look into the recovered frames of all three
methods when $m=12$. Since all the recovered frames achieve a PSNR
higher than 40 dB, it is hard to visually observe any difference
compared to the original video frame. Therefore, we display the
reconstruction error image defined as the absolute difference
between the reconstructed image and the original image. Figures
\ref{GTCSSerr}, \ref{GTCSPerr}, and \ref{KCSerr} visualize the
reconstruction errors of all three methods. Compared to KCS,
GTCS-S achieves much lower reconstruction error using much less
time.

\begin{figure}[htb]
\begin{center}
\subfigure[PSNR
comparison]{\label{MWCSsusiesparsePSNR}\includegraphics[width=0.45\linewidth]{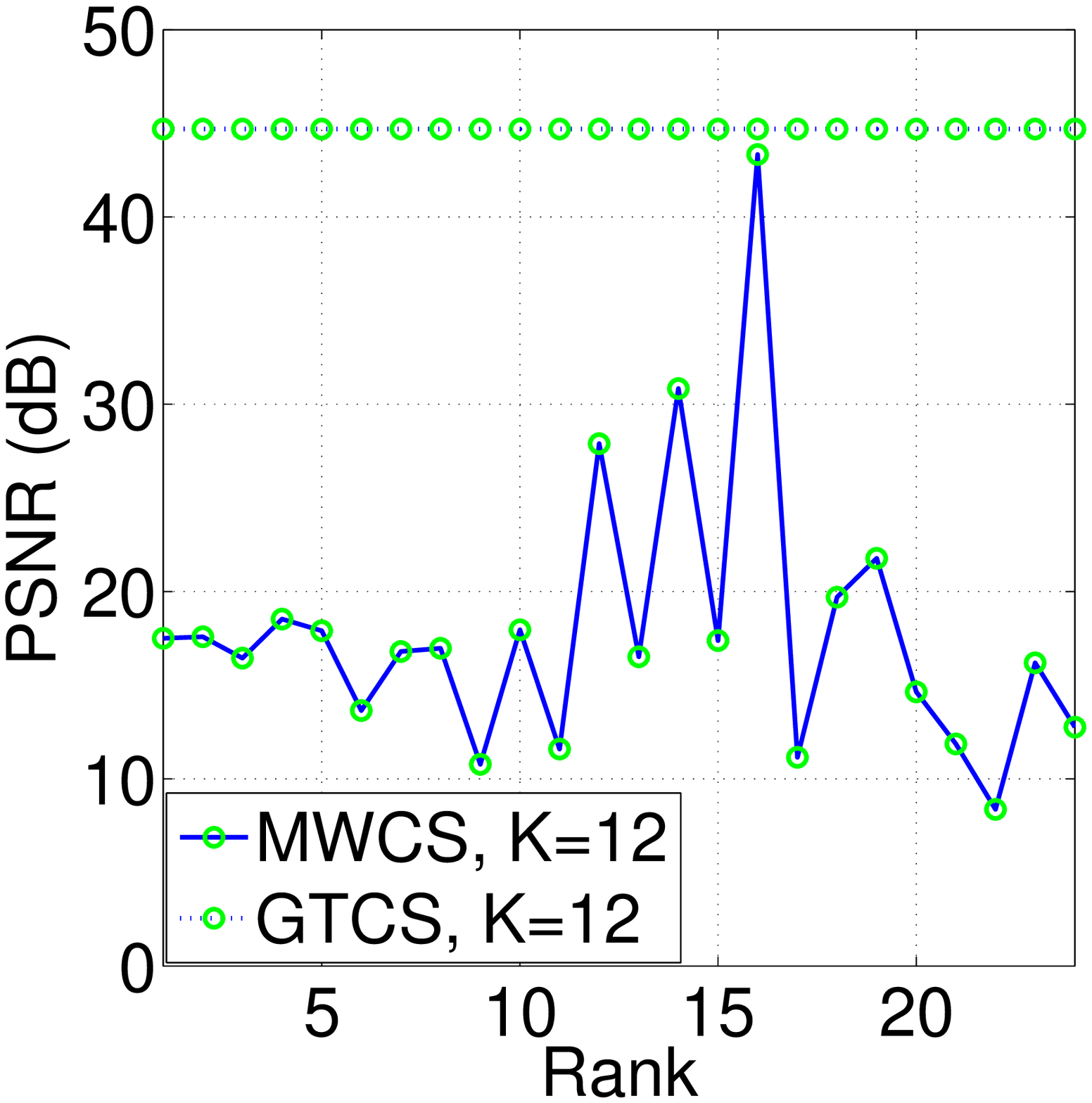}}
\subfigure[Recovery time
comparison]{\label{MWCSsusiesparsetime}\includegraphics[width=0.45\linewidth]{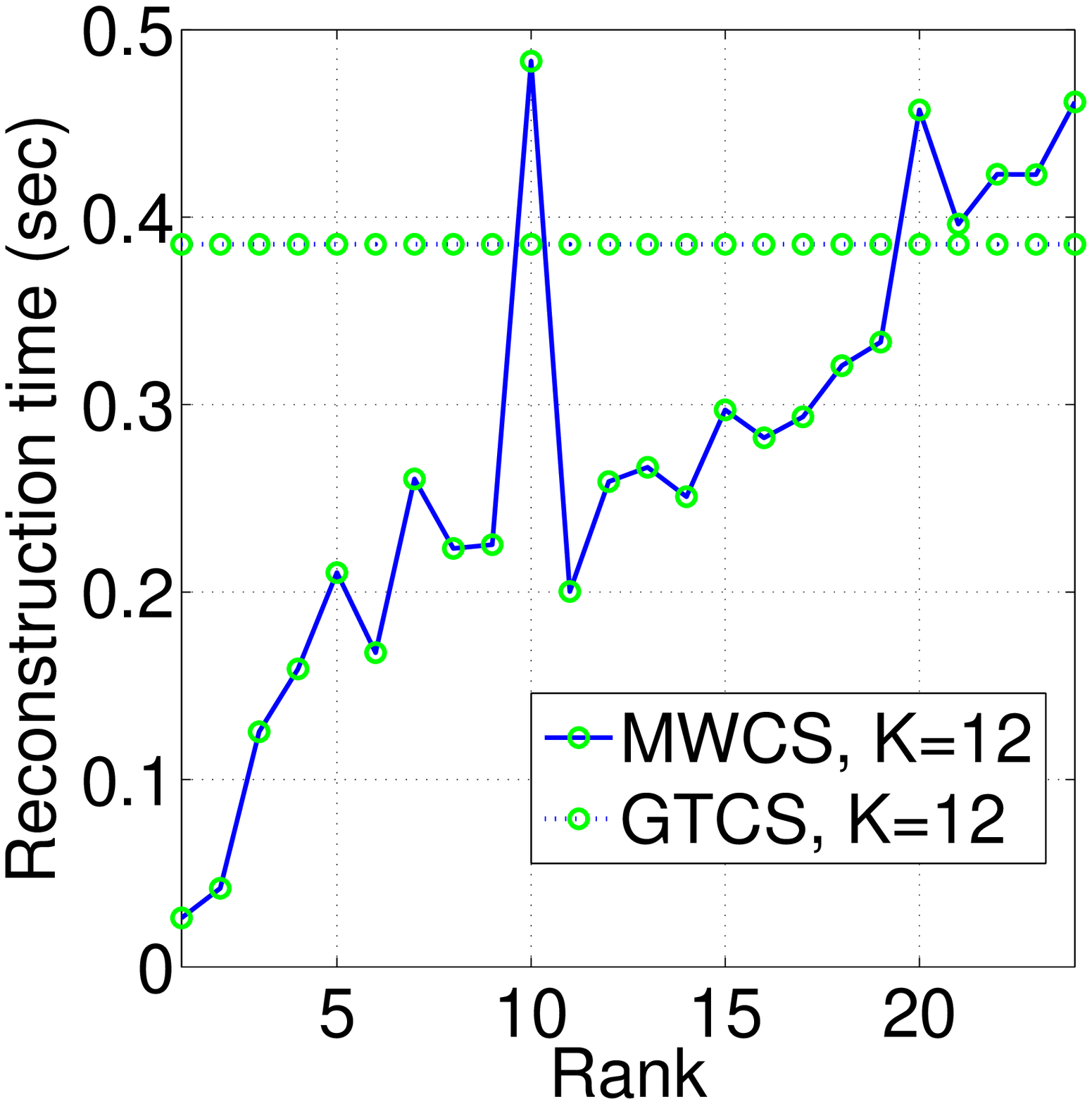}}
\caption{PSNR comparison of GTCS-P with MWCS on sparse video when
$m=12$, using 0.125 normalized number of samples. The highest PSNR
of MWCS with estimated tensor rank varying from 1 to 24 appears
when Rank = 16.}
\end{center}
\end{figure}

To compare the performance of GTCS-P with MWCS, we examine MWCS
with various tensor rank estimations and Figure
\ref{MWCSsusiesparsePSNR} and Figure \ref{MWCSsusiesparsetime}
depict its PSNR and reconstruction time respectively. The straight
line marked GTCS is merely used to indicate the corresponding
performance of GTCS-P with the same amount of measurements and has
nothing to do with various ranks.

\subsection{Compressible video representation}
\begin{figure}[htb]
\begin{center}
\subfigure[PSNR
comparison]{\label{SusiePSNR8}\includegraphics[width=0.45\linewidth]{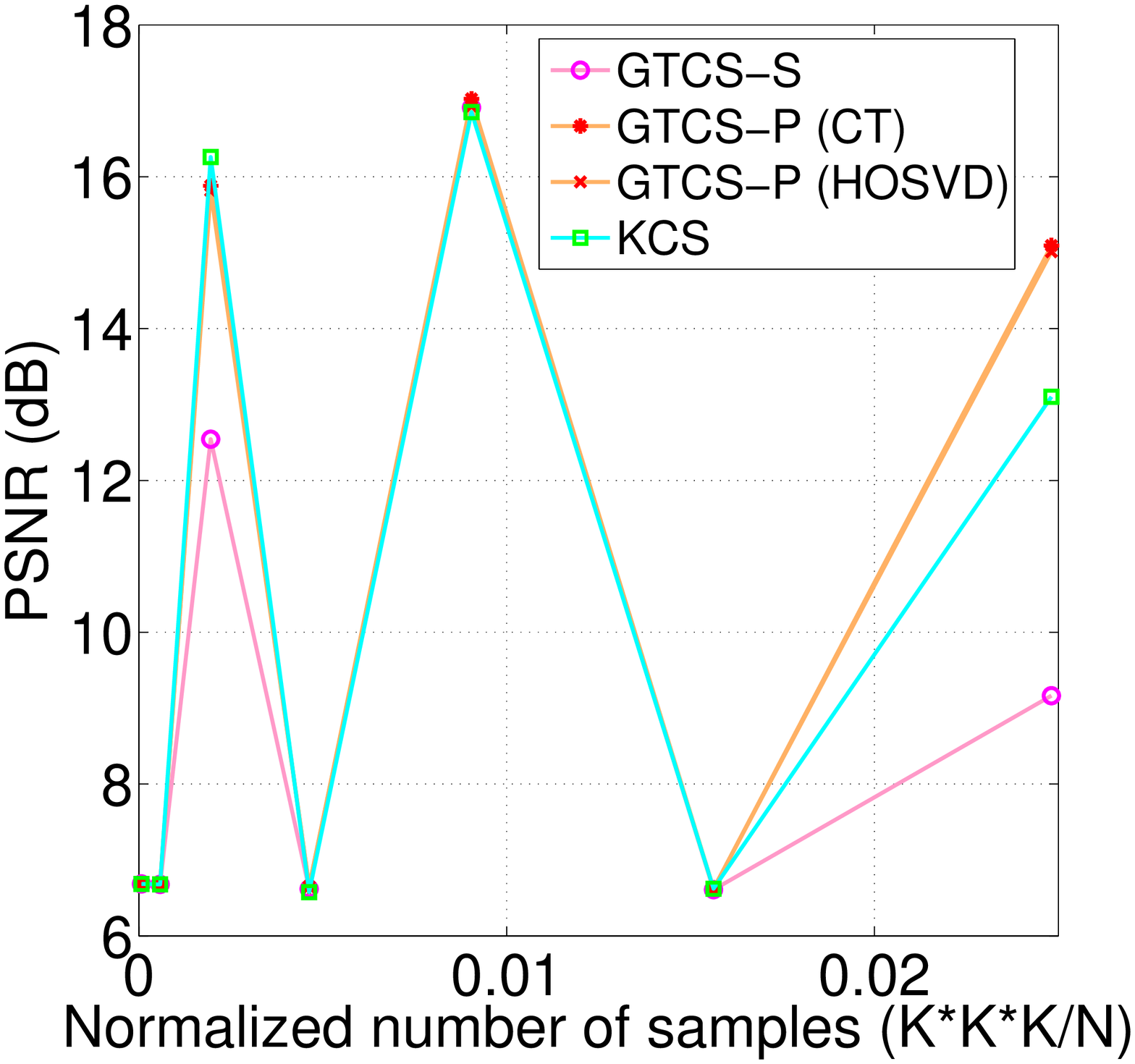}}
\subfigure[Recovery time
comparison]{\label{SusieTime8}\includegraphics[width=0.45\linewidth]{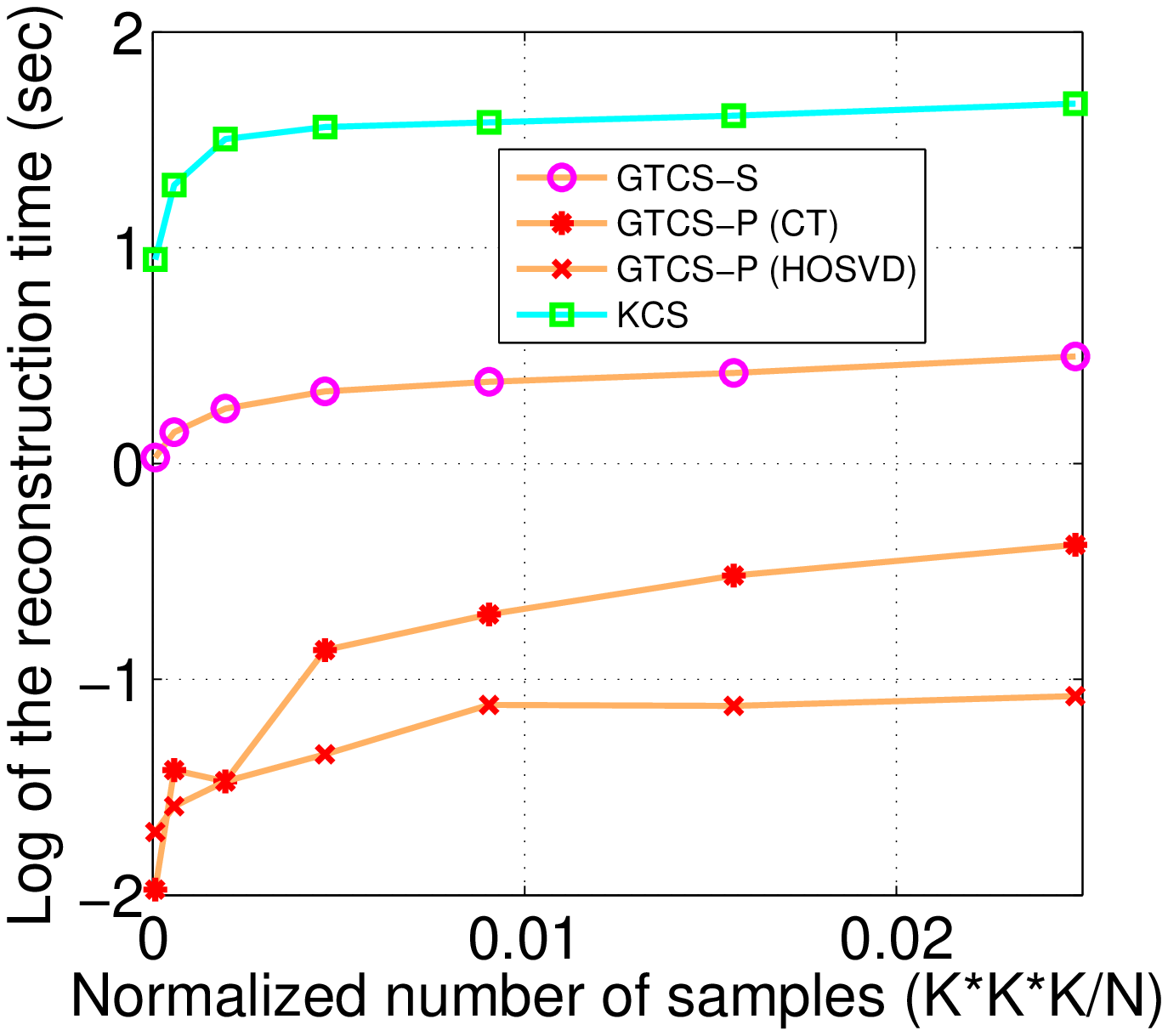}}\\\vspace{-.5cm}
\subfigure[PSNR
comparison]{\label{SusiePSNR21}\includegraphics[width=0.45\linewidth]{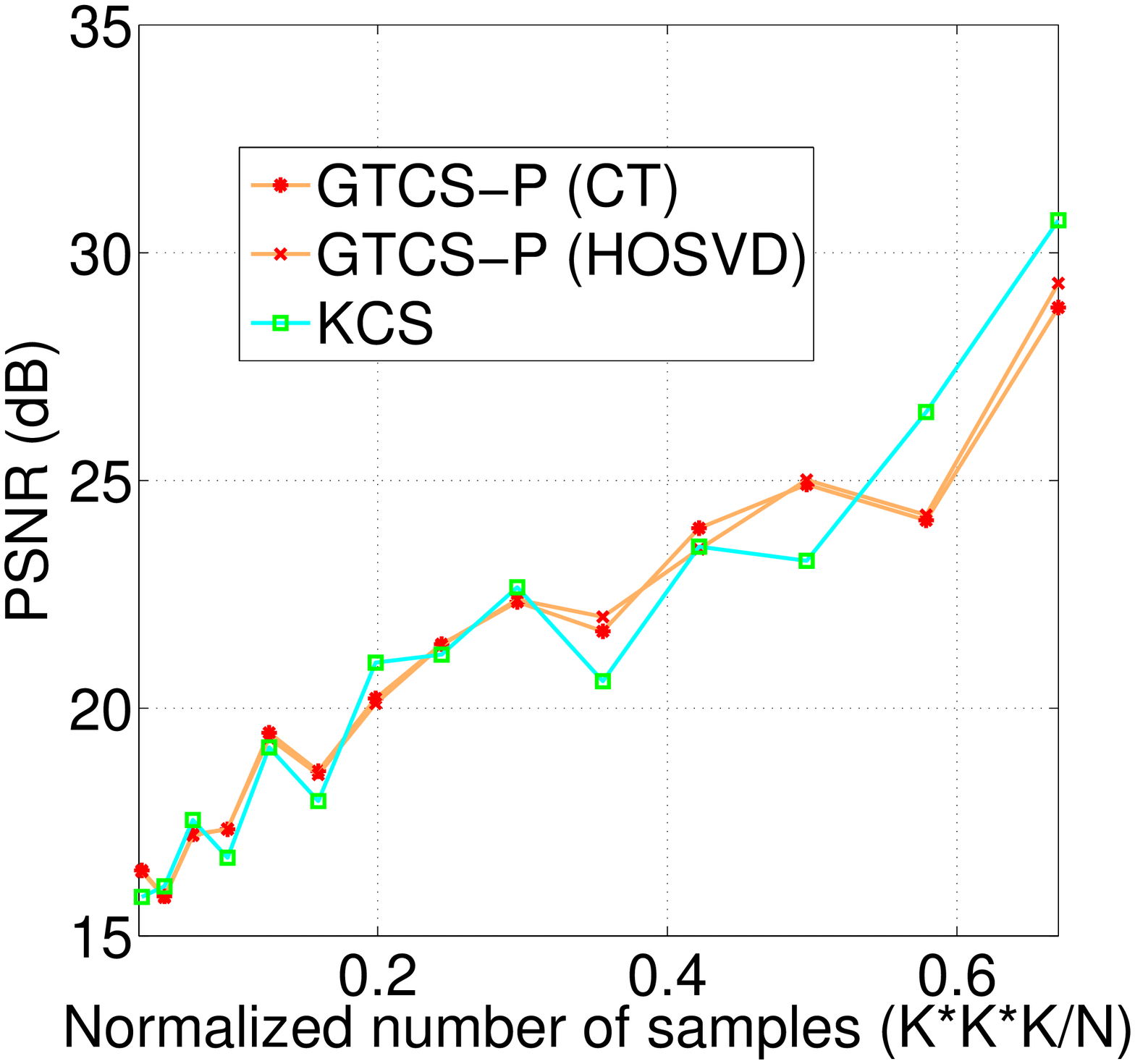}}
\subfigure[Recovery time
comparison]{\label{SusieTime21}\includegraphics[width=0.45\linewidth]{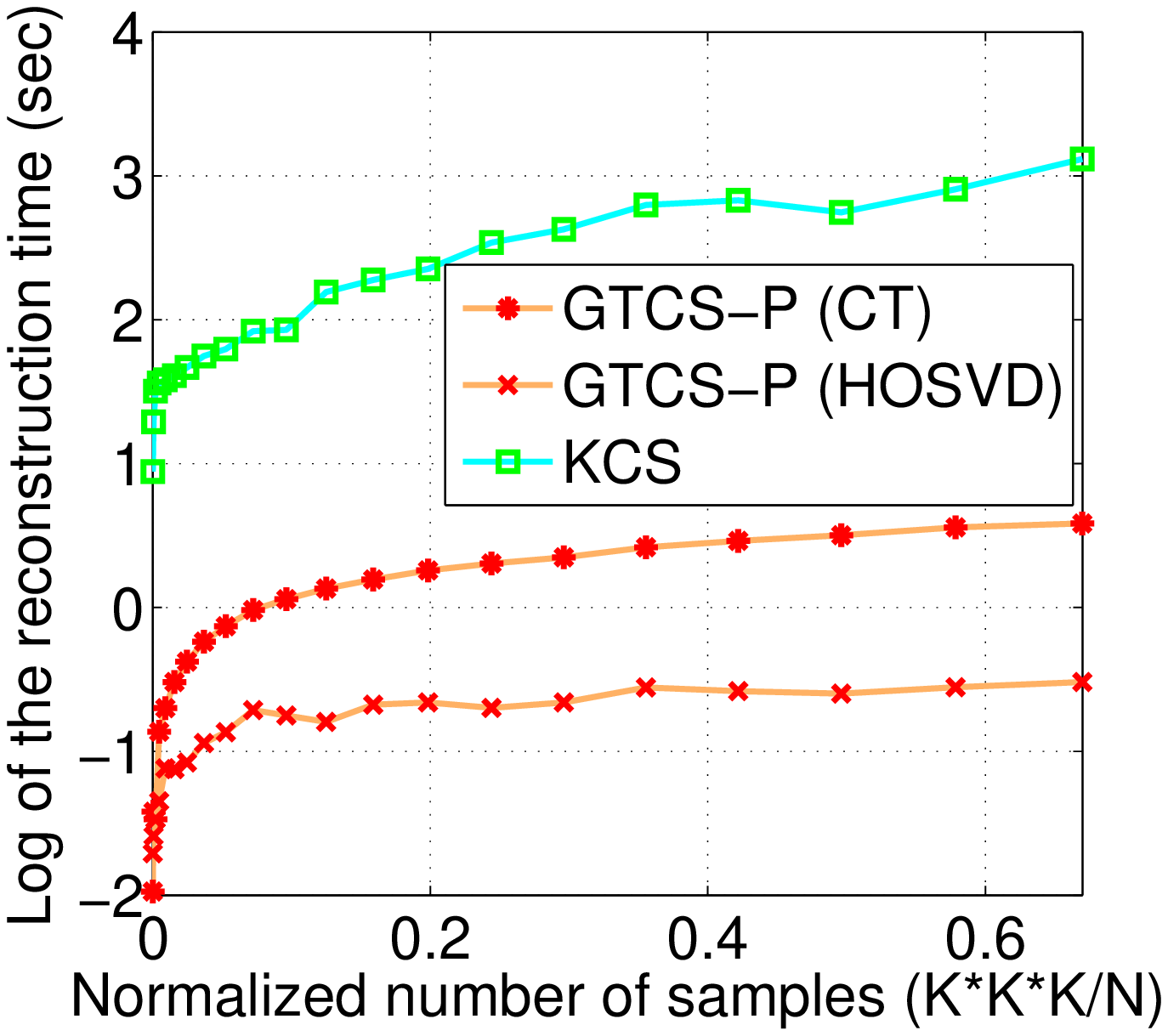}}\vspace{-.3cm}
\caption{PSNR and reconstruction time comparison on susie video.}
\end{center}
\end{figure}

We finally examine the performance of GTCS, KCS and MWCS on
compressible video data. Each frame of the video sequence is
preprocessed to have size $24\times 24$ and we choose the first 24
frames. The video data together is represented by a $24\times
24\times 24$ tensor. The video itself is non-sparse, yet
compressible in three-dimensional DCT domain. In the decomposition
stage of GTCS-P, we employ a decomposition described in Section
\ref{mareview} to obtain a weaker form of the core Tucker
decomposition and denote this method by GTCS-P (CT). We also test
the performance of GTCS-P by using HOSVD in the decomposition
stage and denote this method by GTCS-P (HOSVD) hereby. $m$ varies
from 1 to 21. Note that in GTCS-S, the reconstruction is not
perfect at each mode, and becomes more and more noisy as the
recovery by mode continues. Therefore the recovery method by
$\ell_1$-minimization using \eqref{l1minrec} would be
inappropriate or even has no solution at certain stage. In our
experiment, GTCS-S by \eqref{l1minrec} works for $m$ from 1 to 7.
To use GTCS-S for $m = 8$ and higher, relaxed recovery
\eqref{l1minrecnoisy} could be employed for reconstruction. Figure
\ref{SusiePSNR8} and Figure \ref{SusieTime8} depict PSNR and
reconstruction time of all methods up to $m=7$. For $m = 8$ to 21,
the results are shown in Figure \ref{SusiePSNR21} and Figure
\ref{SusieTime21}.

\begin{figure}[htb]
\begin{center}
\subfigure[Original frame
1]{\includegraphics[width=0.325\linewidth]{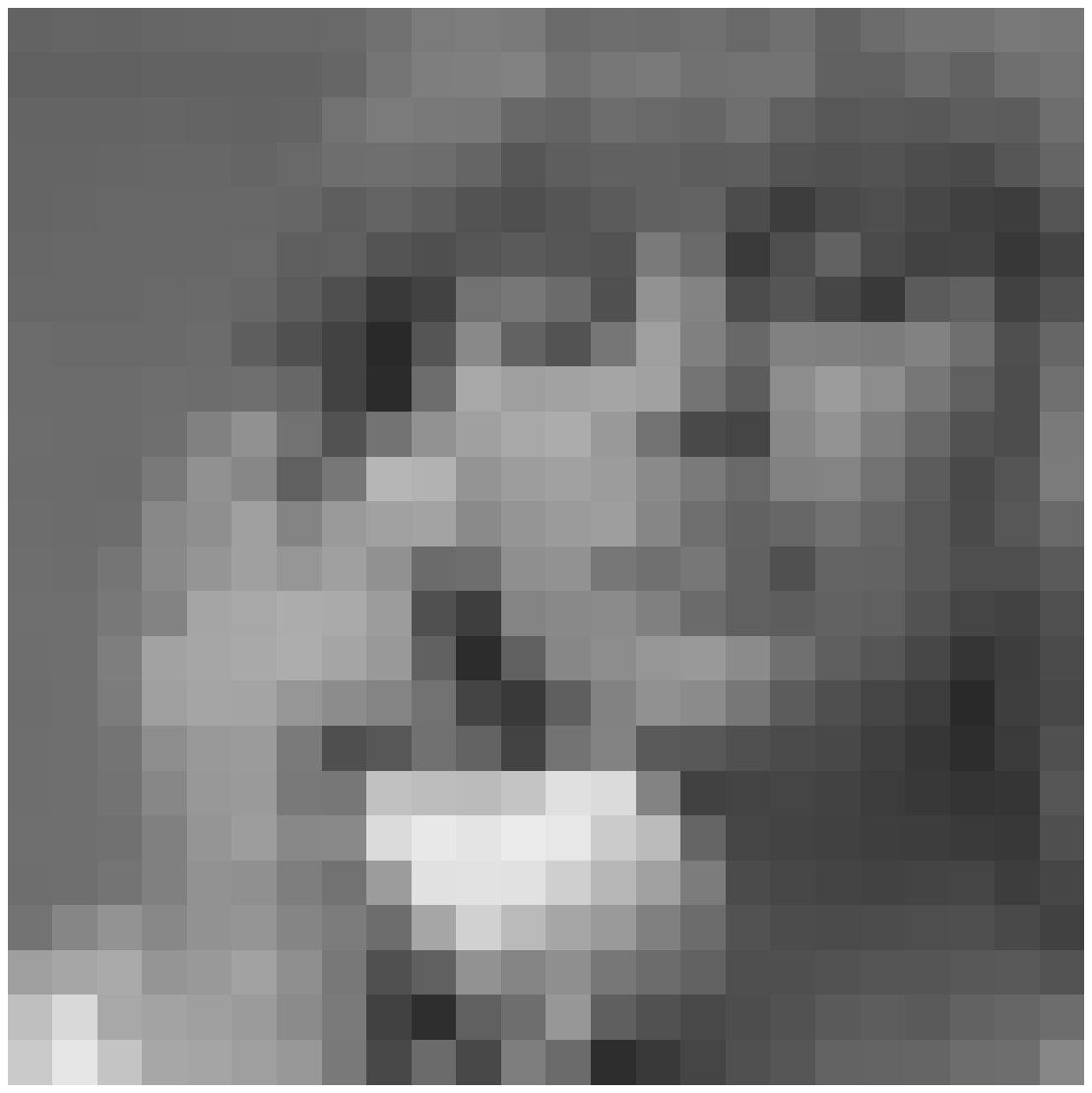}}
\subfigure[Original frame
9]{\includegraphics[width=0.325\linewidth]{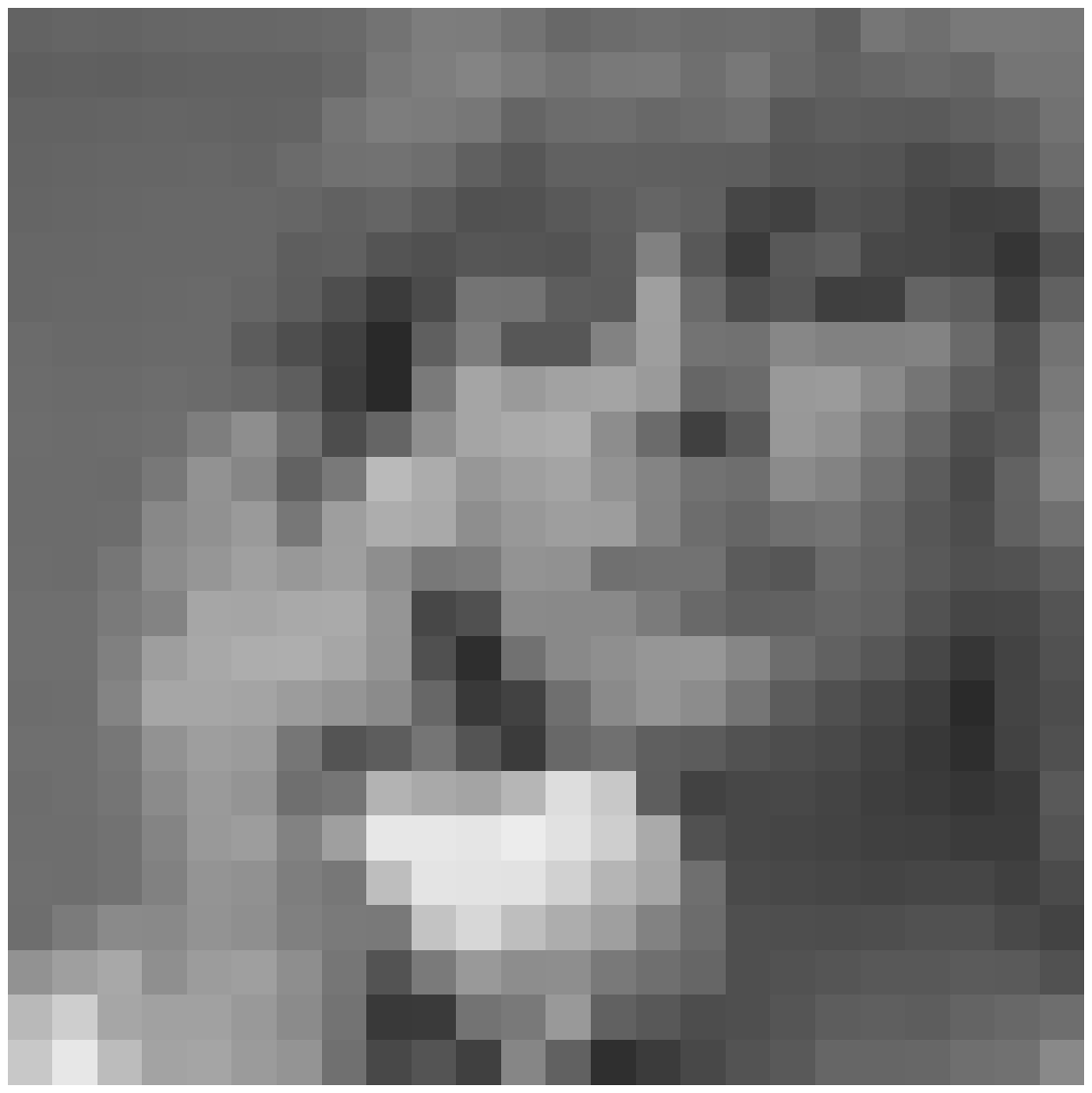}}
\subfigure[Original frame
17]{\includegraphics[width=0.325\linewidth]{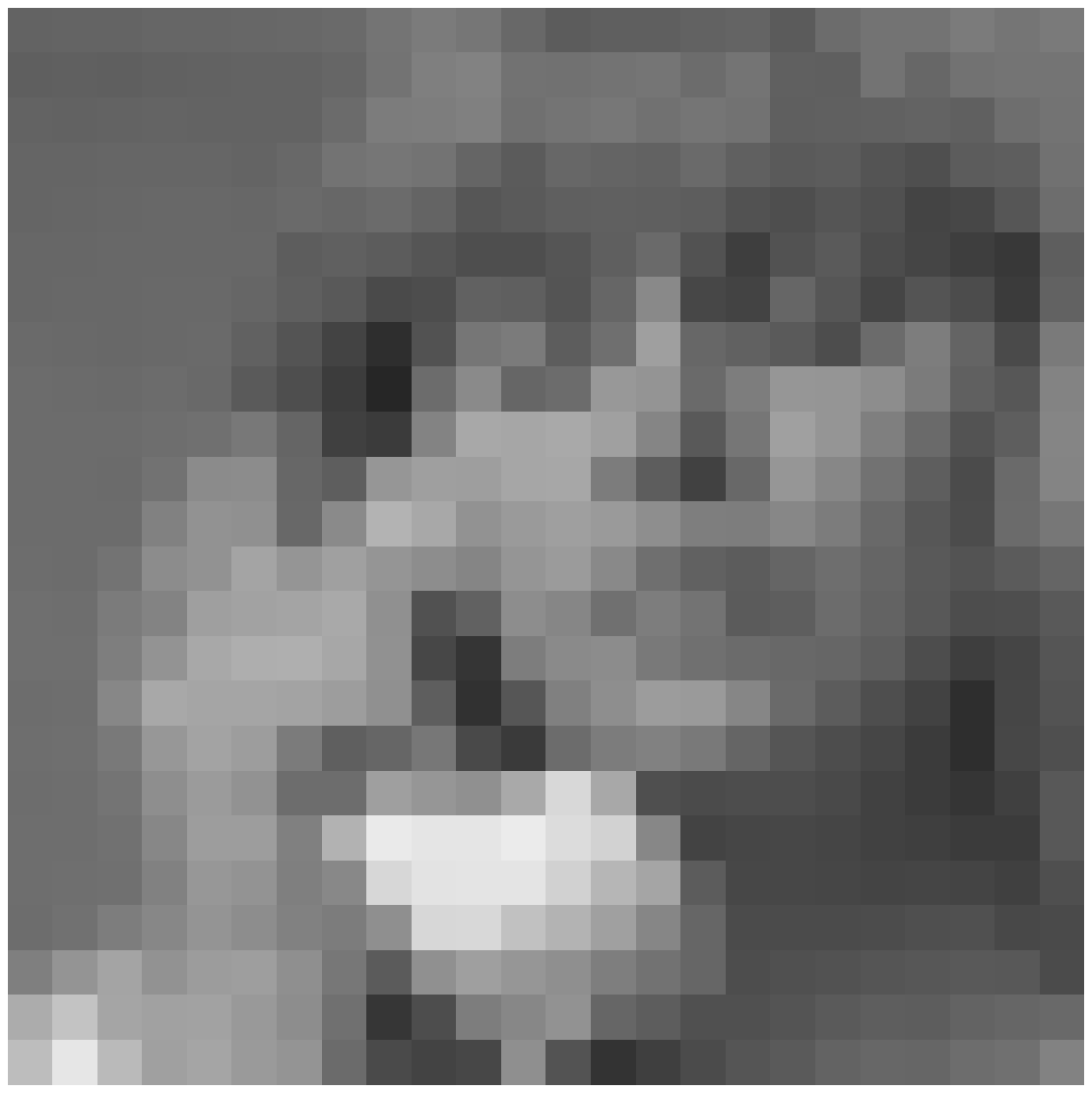}}
\caption{Original video frames.}\label{originalframes}
\end{center}
\end{figure}

\begin{figure}[htb]
\begin{center}
\subfigure[GTCS-P(HOSVD) reconstructed frame
1]{\includegraphics[width=0.325\linewidth]{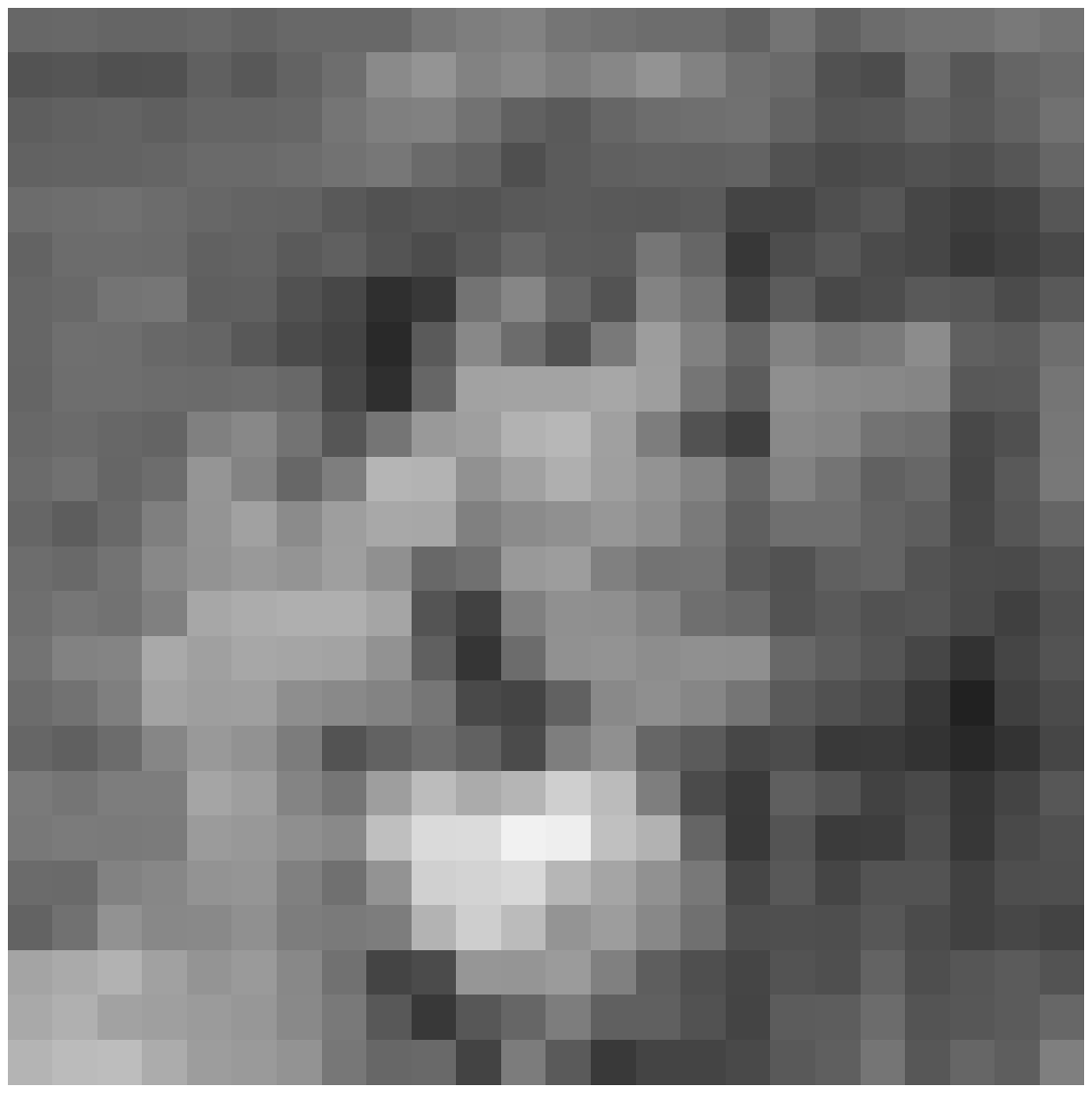}}
\subfigure[GTCS-P(HOSVD) reconstructed frame
9]{\includegraphics[width=0.325\linewidth]{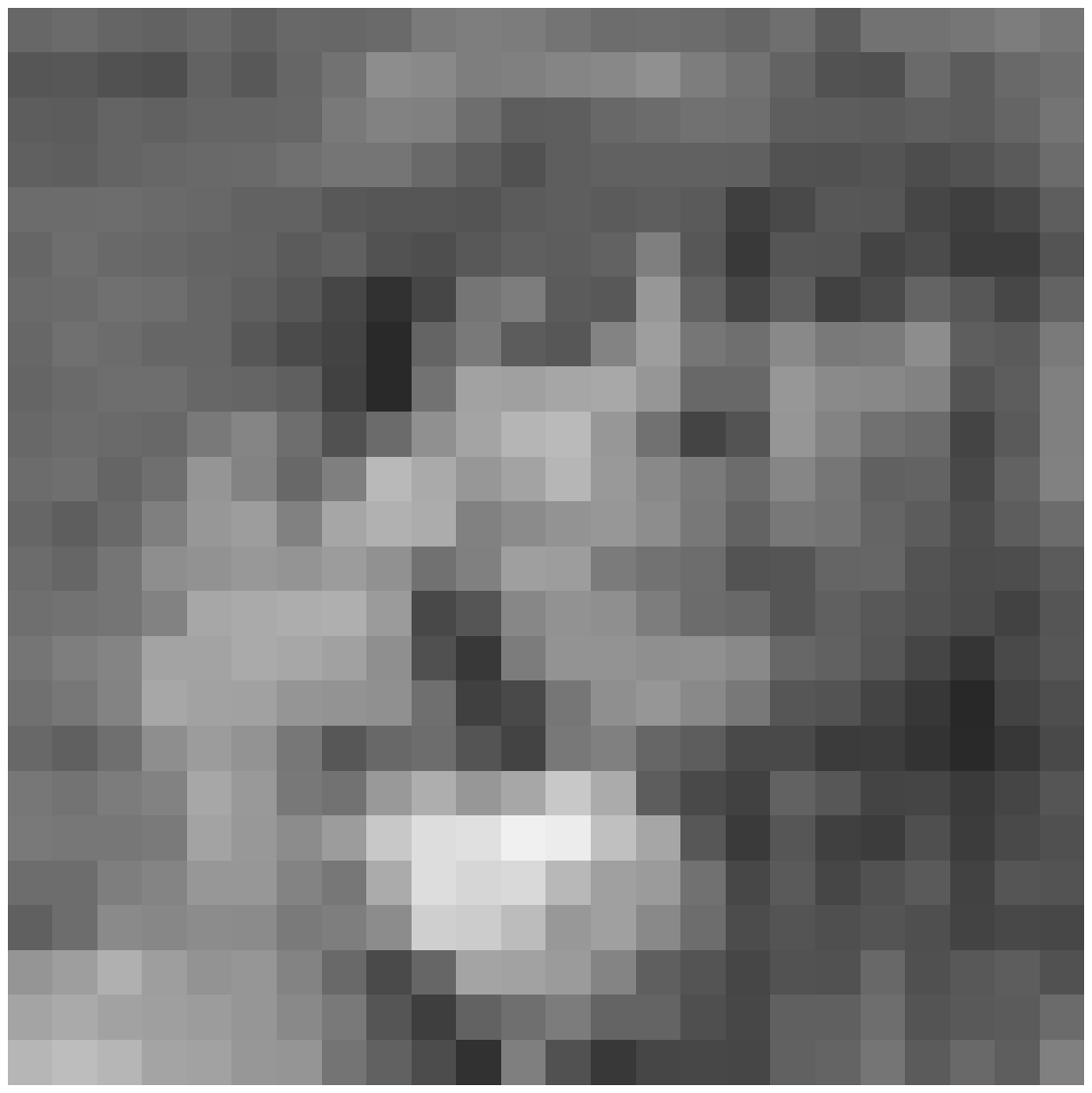}}
\subfigure[GTCS-P(HOSVD) reconstructed frame
17]{\includegraphics[width=0.325\linewidth]{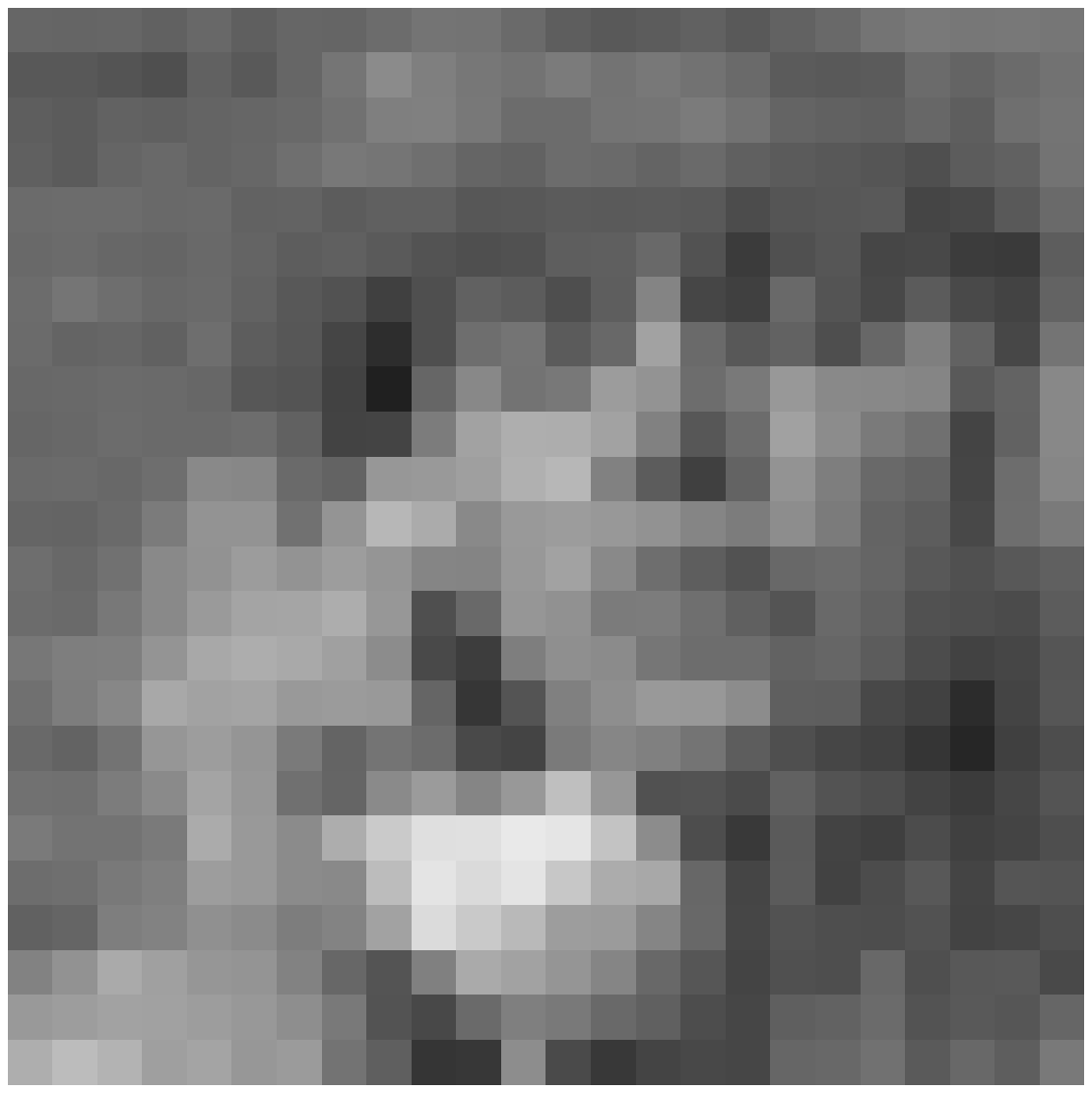}}\\\vspace{-.3cm}
\subfigure[GTCS-P(CT) reconstructed frame
1]{\includegraphics[width=0.325\linewidth]{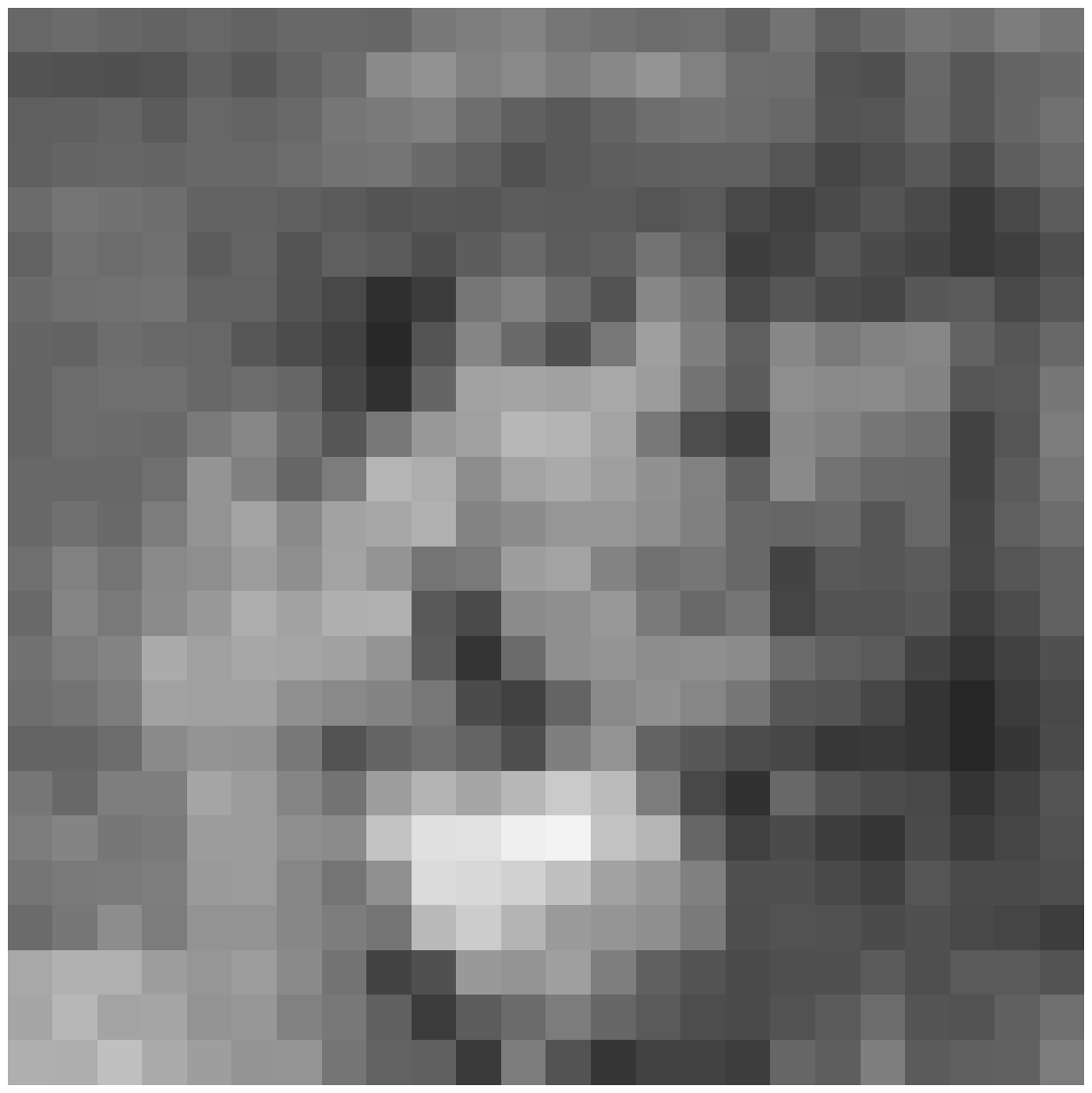}}
\subfigure[GTCS-P(CT) reconstructed frame
9]{\includegraphics[width=0.325\linewidth]{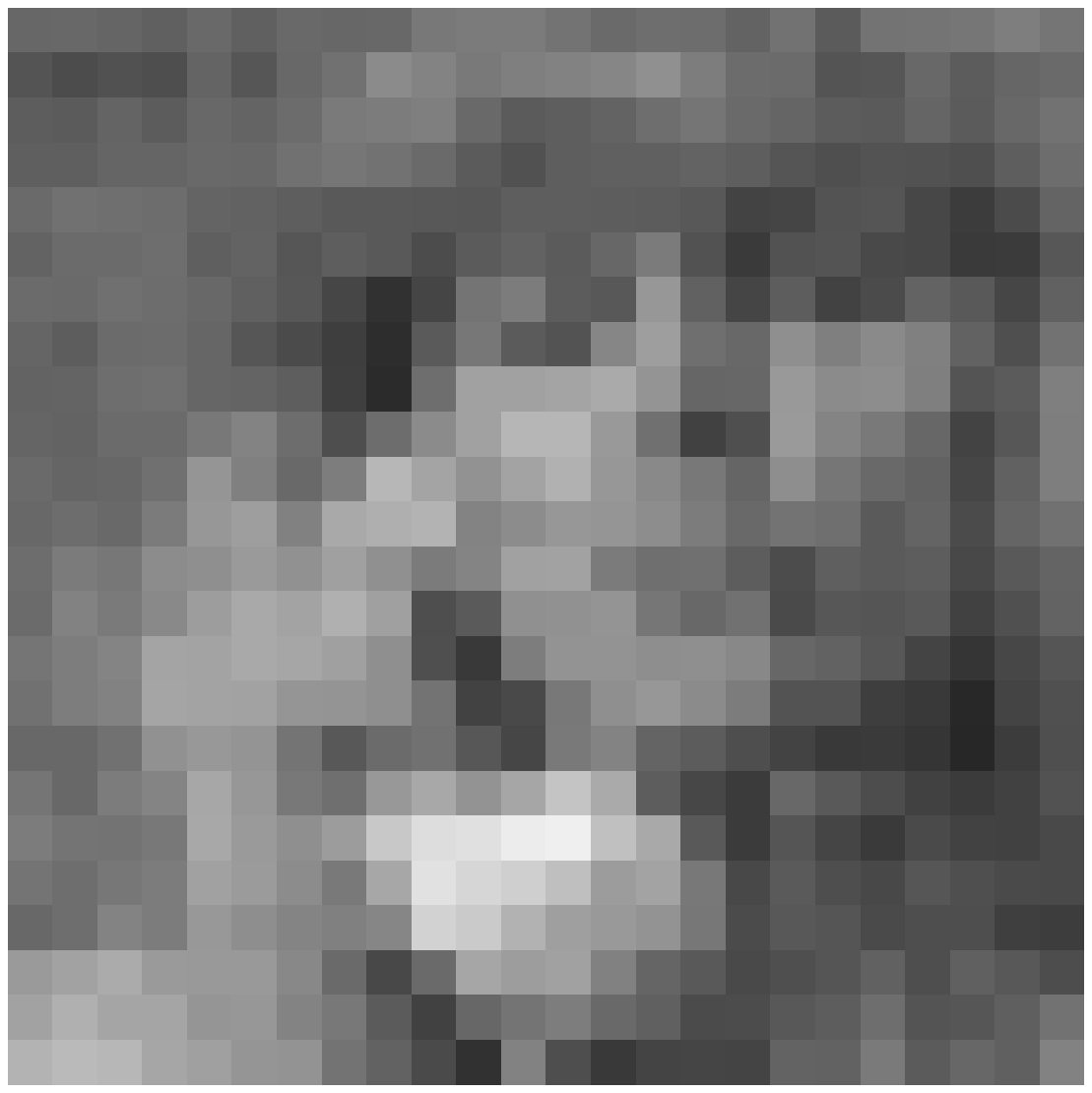}}
\subfigure[GTCS-P(CT) reconstructed frame
17]{\includegraphics[width=0.325\linewidth]{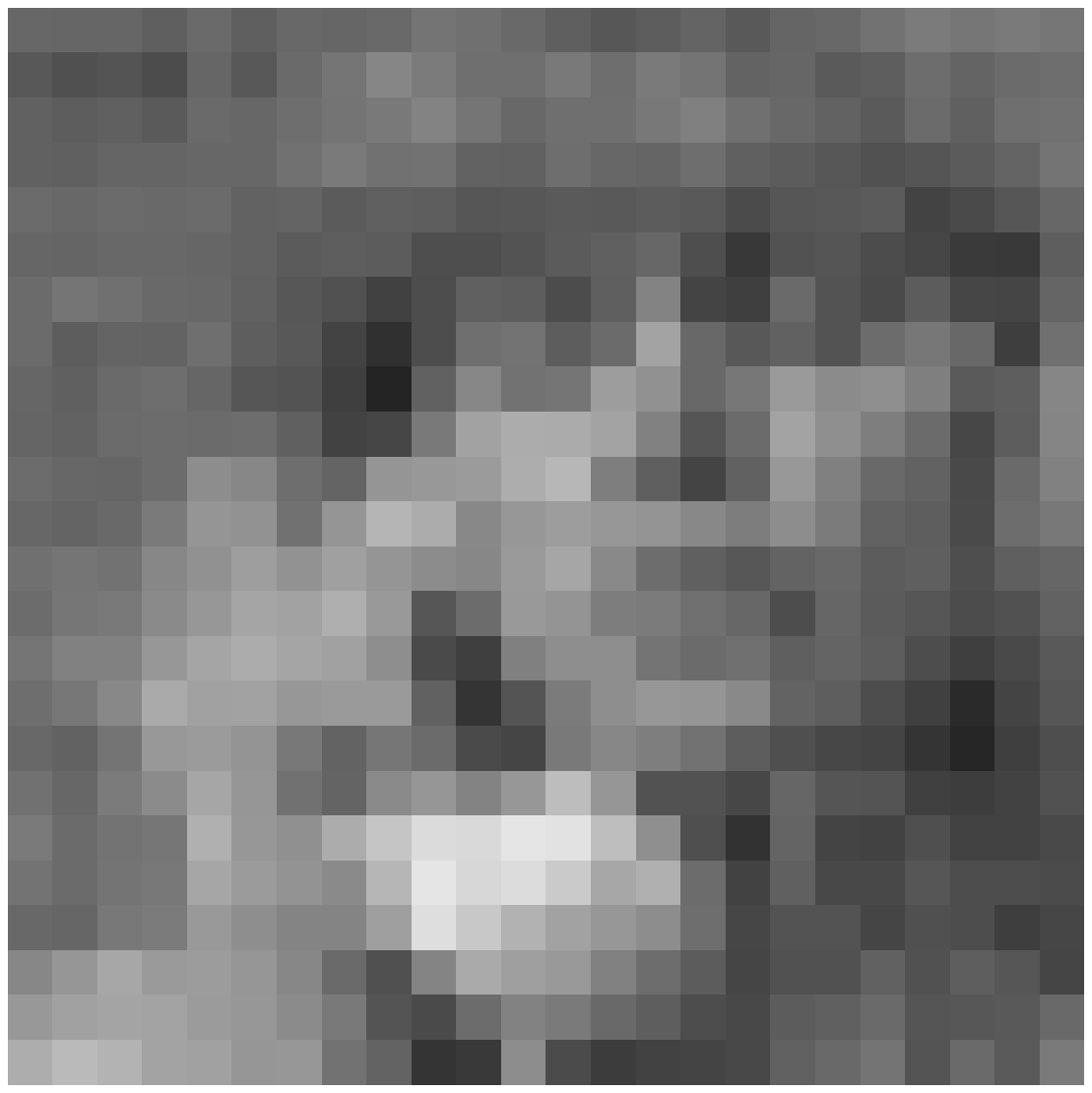}}\\\vspace{-.3cm}
\subfigure[KCS reconstructed frame
1]{\includegraphics[width=0.325\linewidth]{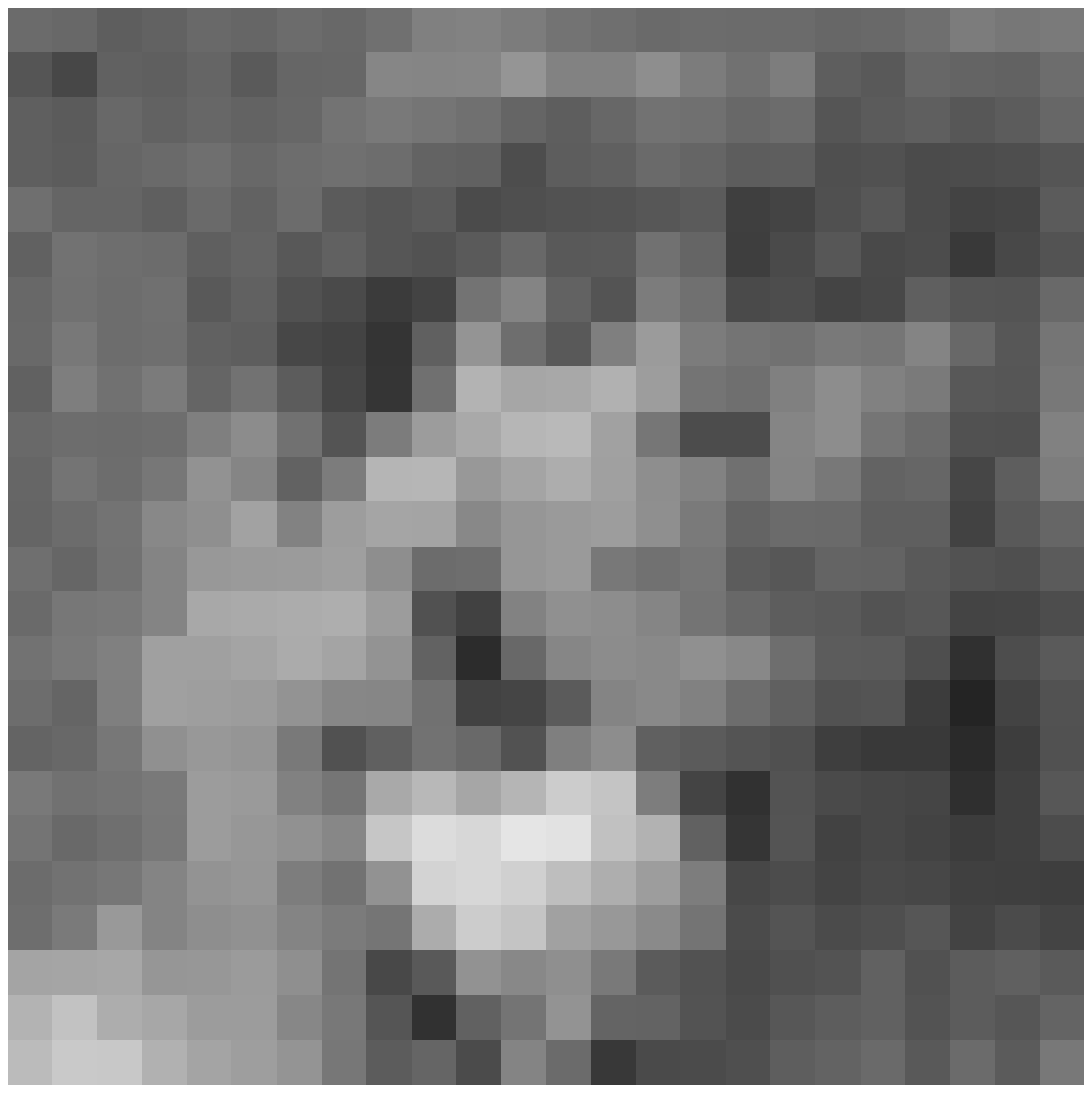}}
\subfigure[KCS reconstructed frame
9]{\includegraphics[width=0.325\linewidth]{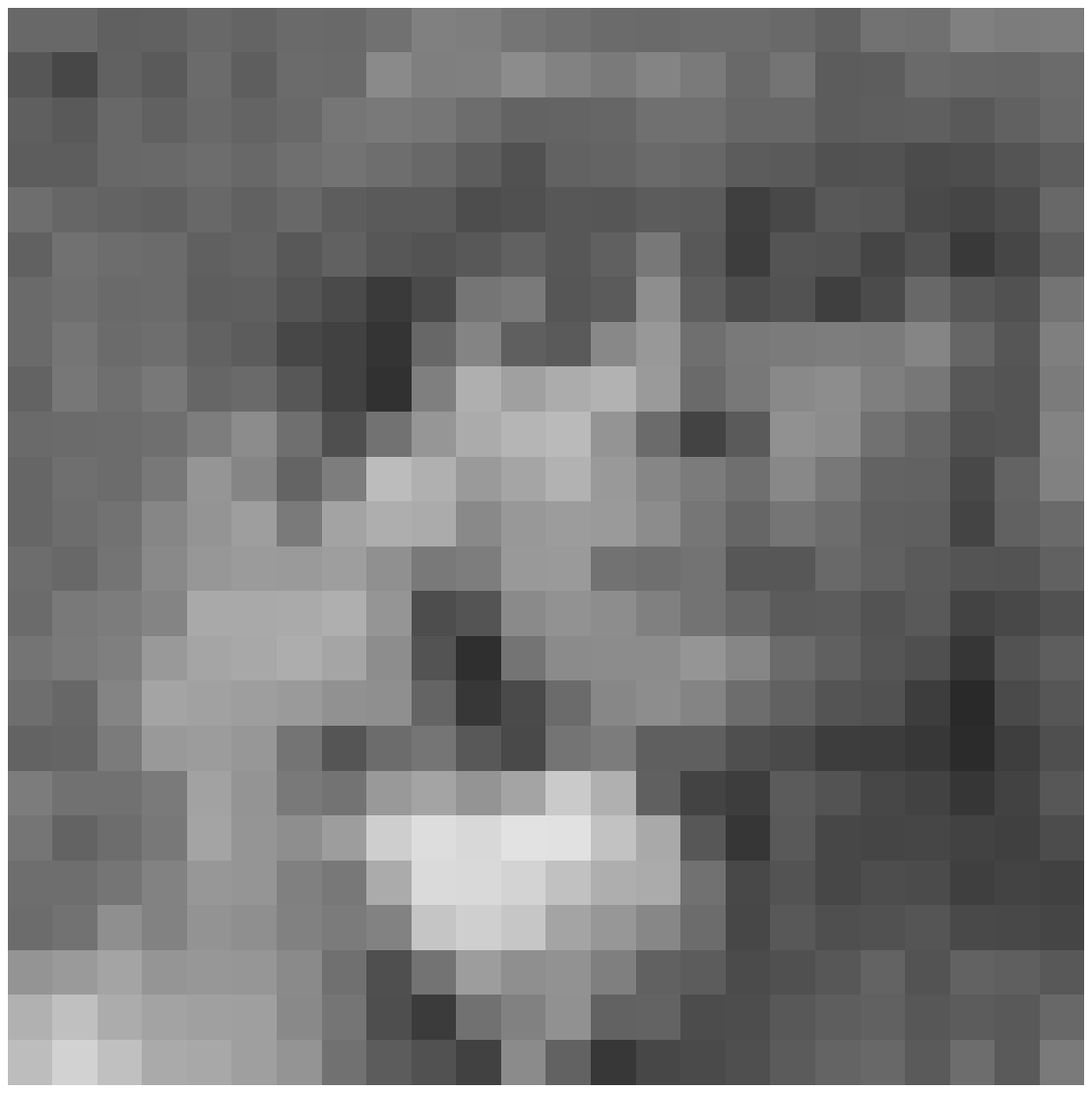}}
\subfigure[KCS reconstructed frame
17]{\includegraphics[width=0.325\linewidth]{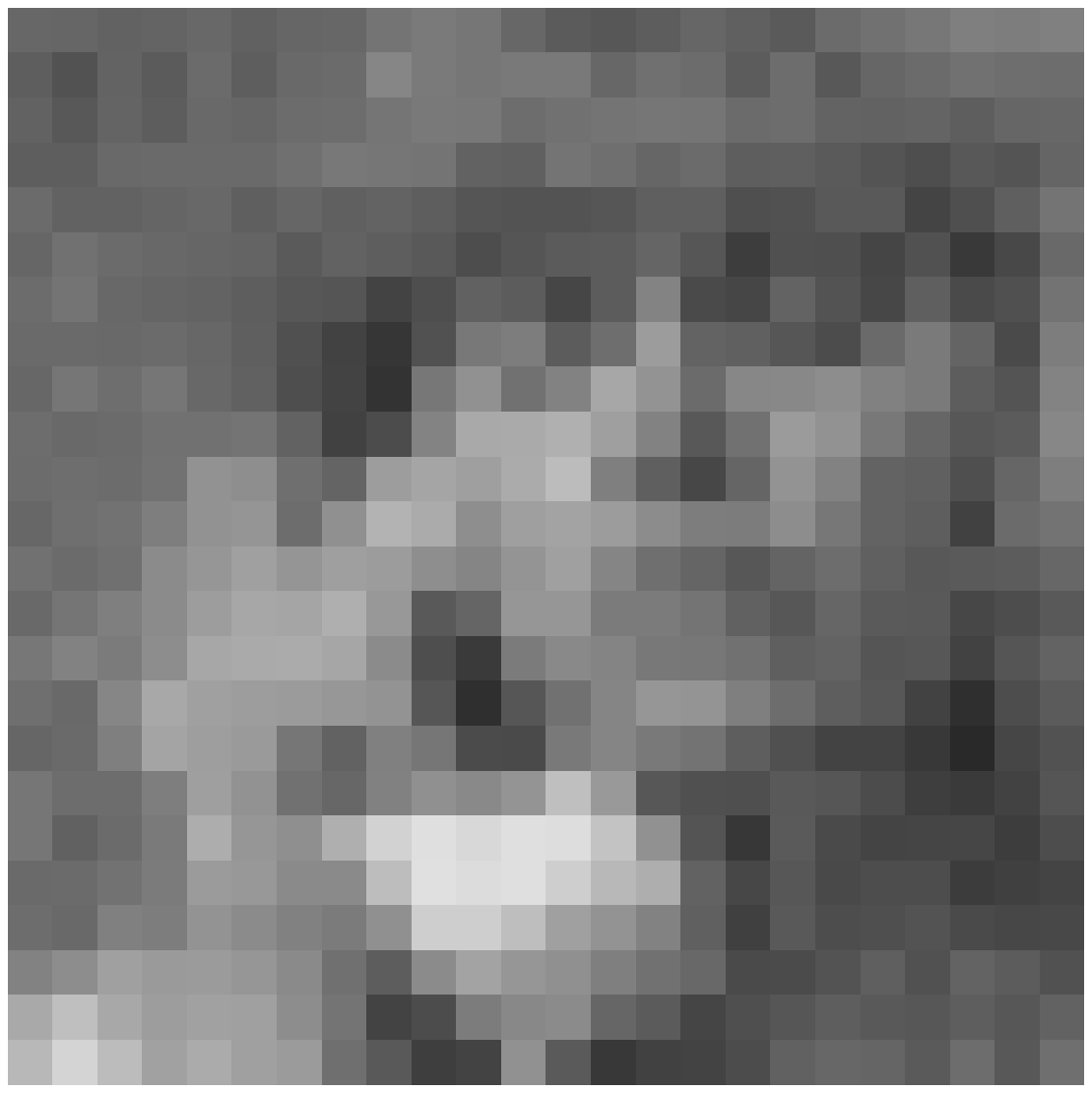}}\\\vspace{-.3cm}
\subfigure[MWCS reconstructed frame
1]{\includegraphics[width=0.325\linewidth]{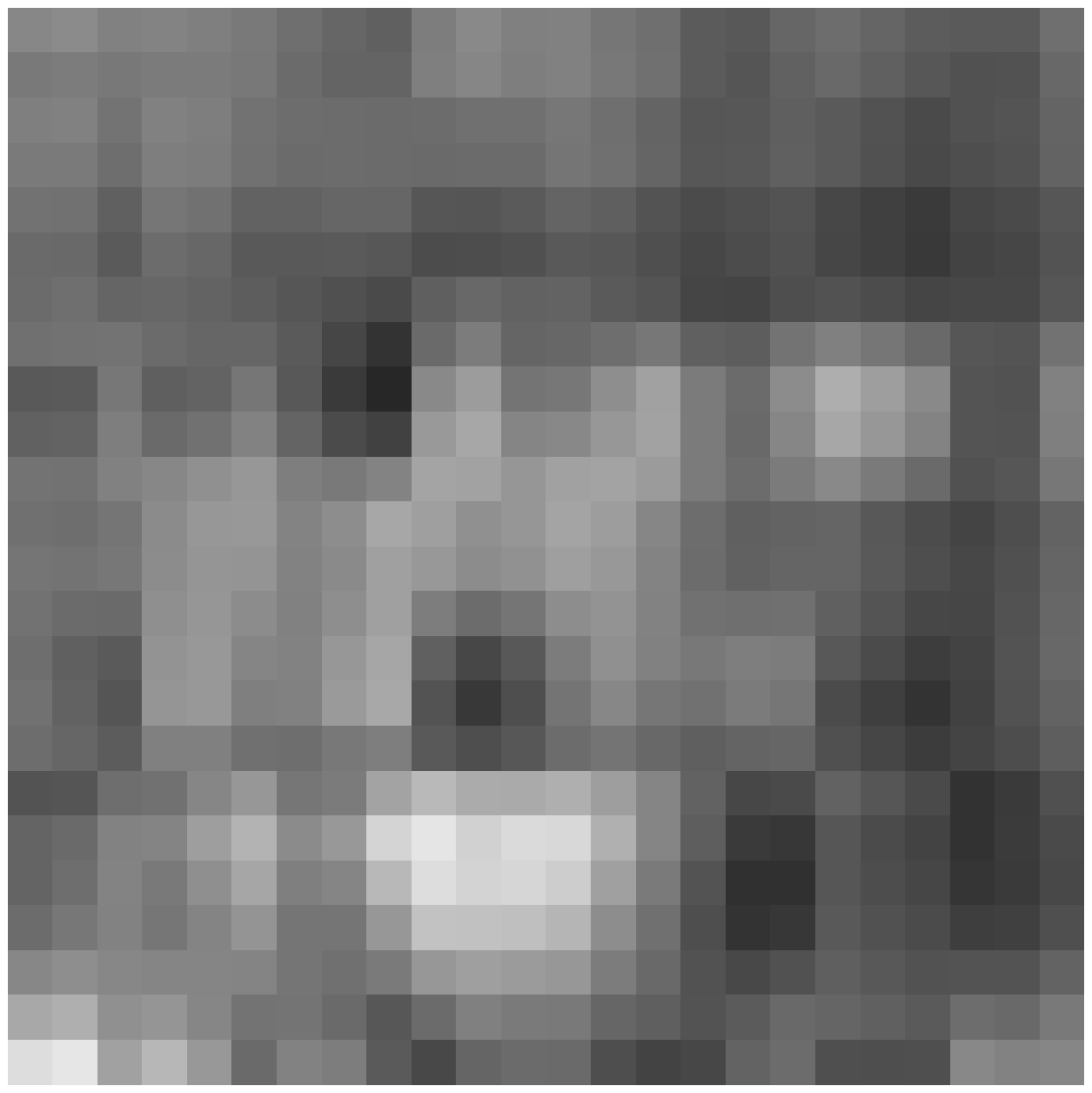}}
\subfigure[MWCS reconstructed frame
9]{\includegraphics[width=0.325\linewidth]{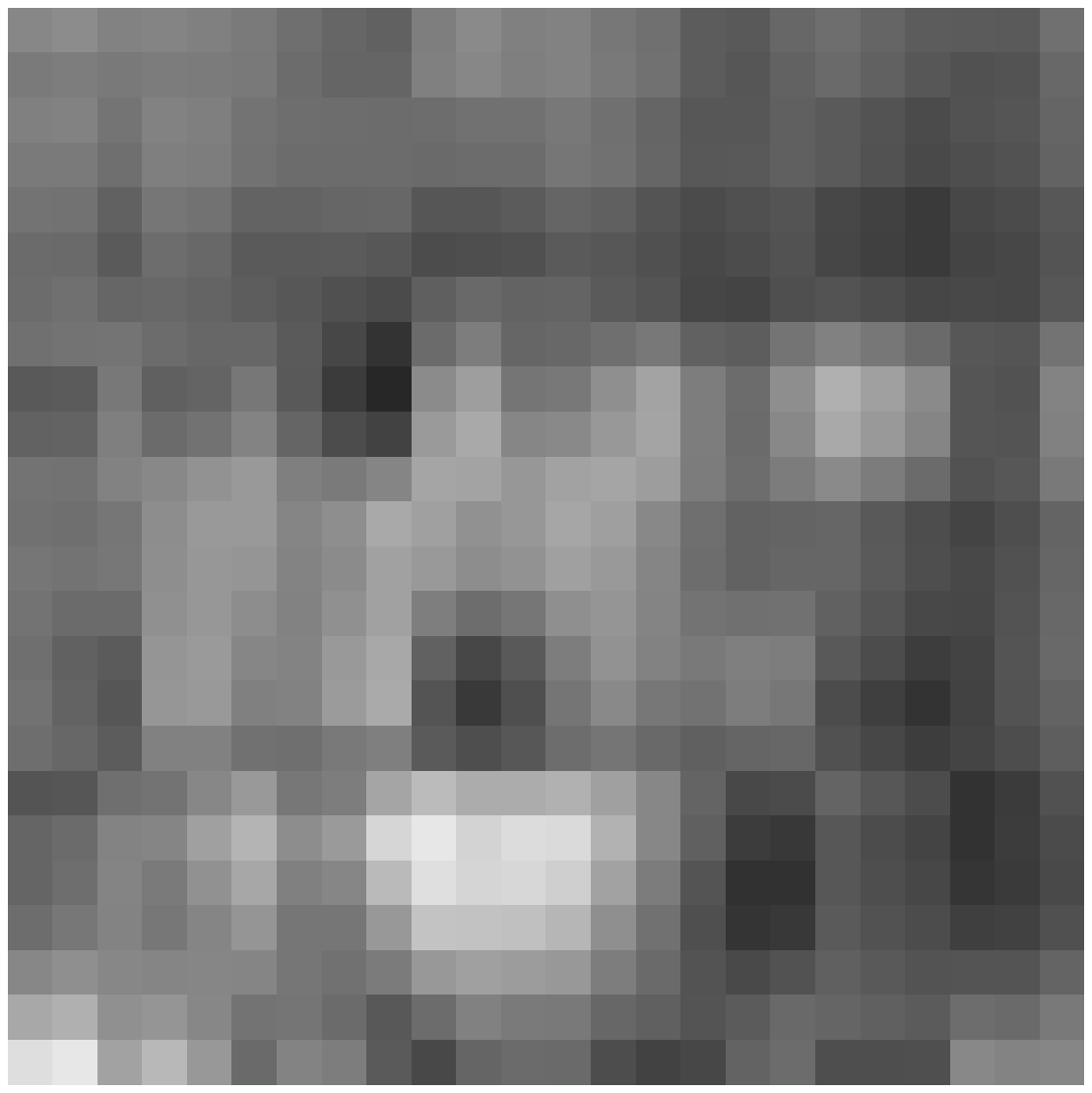}}
\subfigure[MWCS reconstructed frame
17]{\includegraphics[width=0.325\linewidth]{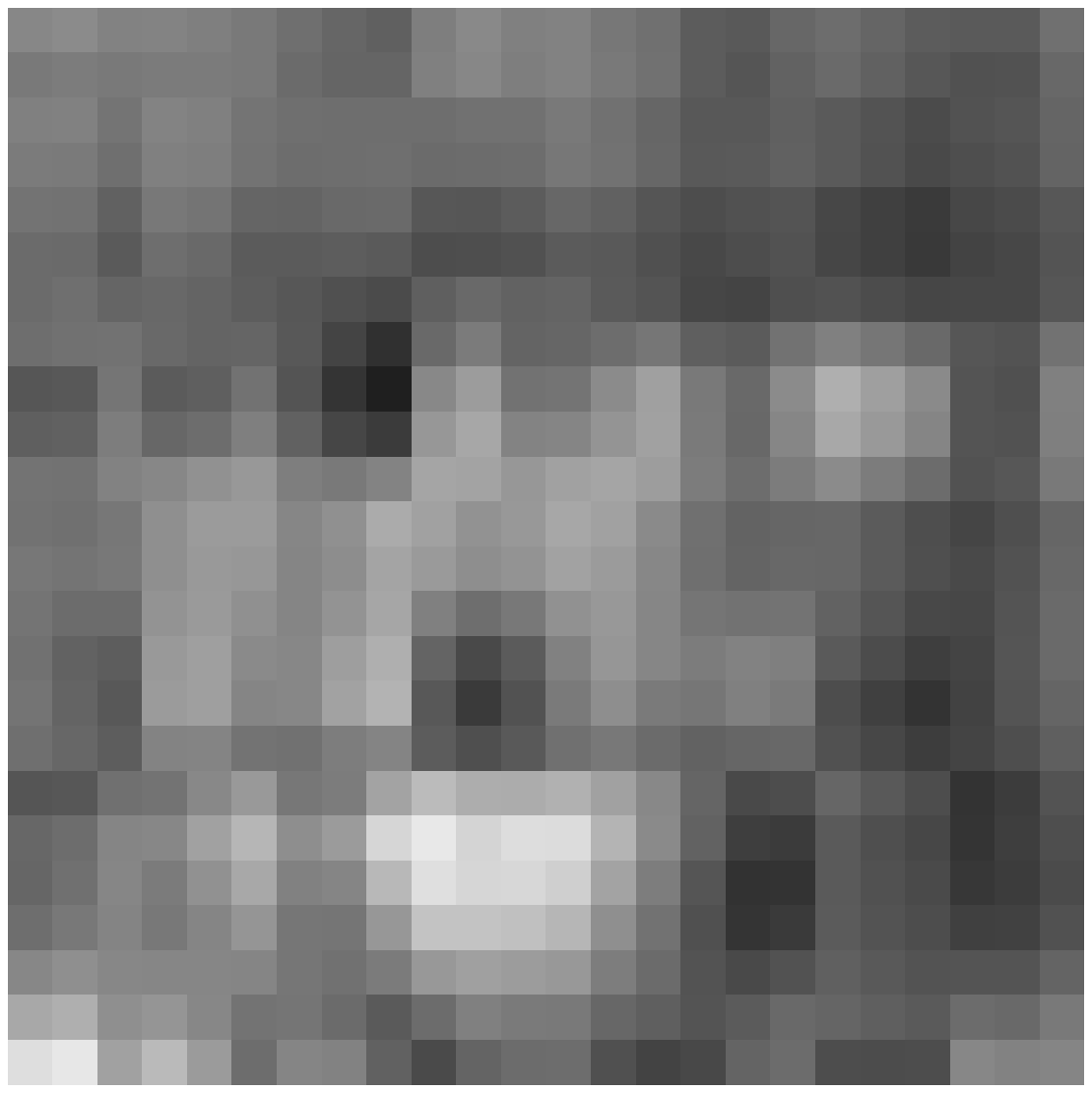}}\\\vspace{-.3cm}
\caption{Reconstructed video frames when m = 21 using 0.67
normalized number of samples by GTCS-P (HOSVD, PSNR = 29.33 dB),
GTCS-P (CT, PSNR = 28.79 dB), KCS (PSNR = 30.70 dB) and MWCS (Rank
= 4, PSNR = 22.98 dB).}\label{framesK21}
\end{center}
\end{figure}

\begin{figure}[htb]
\begin{center}
\subfigure[PSNR
comparison]{\label{MWCSsusiePSNR}\includegraphics[width=0.45\linewidth]{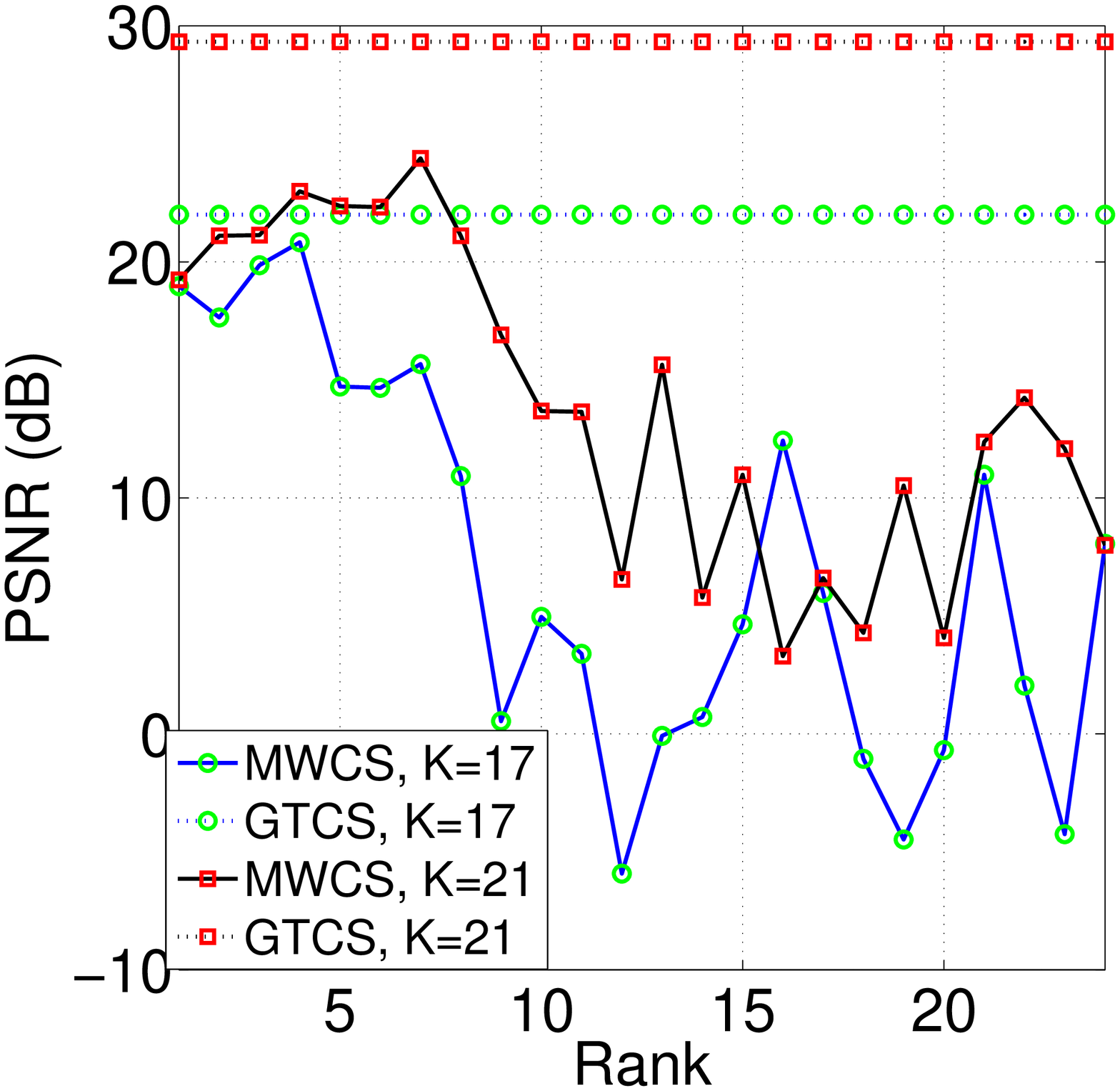}}
\subfigure[Recovery time
comparison]{\label{MWCSsusietime}\includegraphics[width=0.45\linewidth]{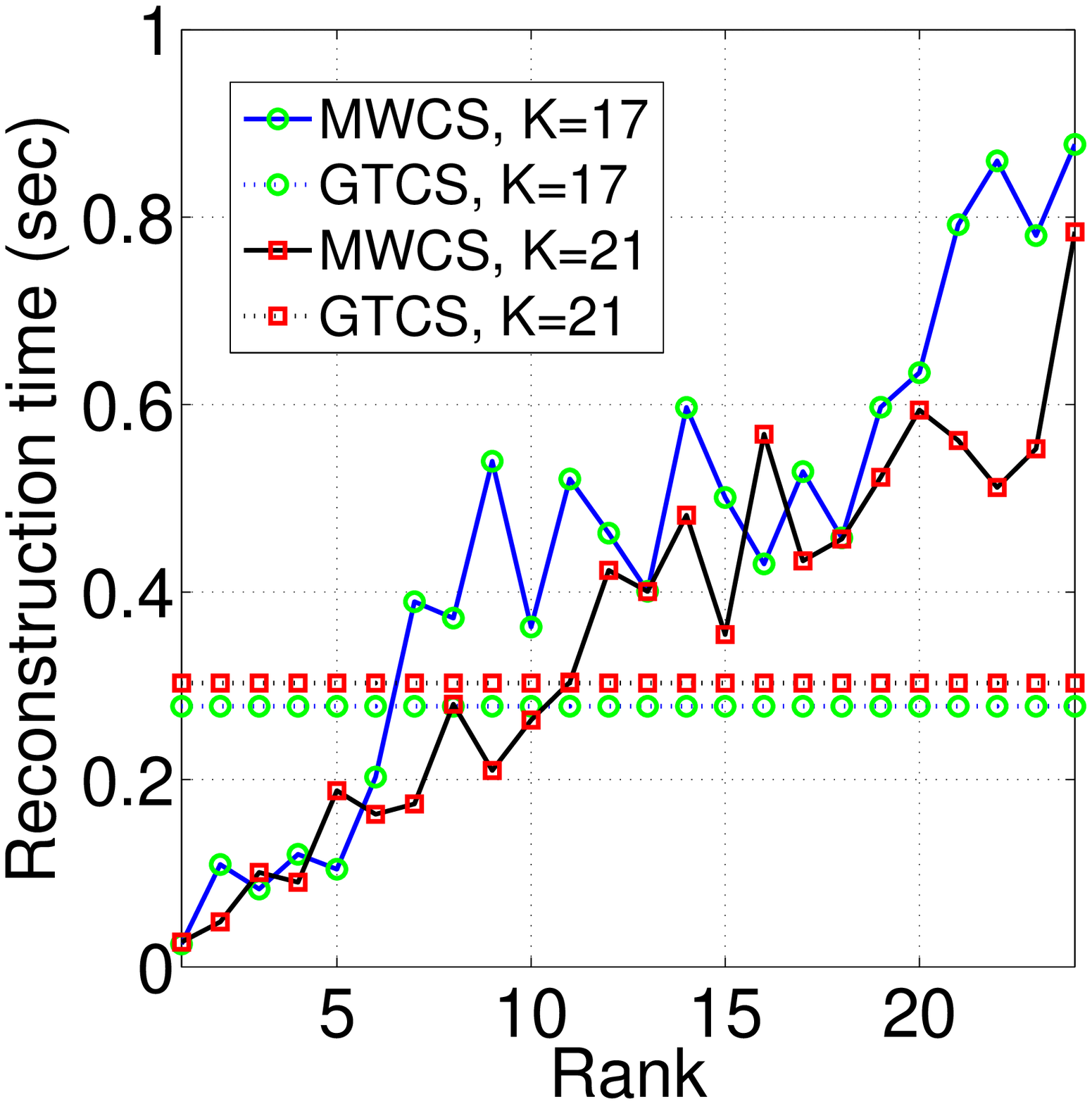}}
\caption{PSNR comparison of GTCS-P with MWCS on compressible video
when $m=17$, using 0.36 normalized number of samples and $m = 21$,
using 0.67 normalized number of samples. The highest PSNR of MWCS
with estimated tensor rank varying from 1 to 24 appears when Rank
= 4 and Rank = 7 respectively.}
\end{center}
\end{figure}

We specifically look into the recovered frames of all methods when
$m=17$ and $m=21$. Recovered frames 1, 9, 17 (originally as shown
in Figure \ref{originalframes}) are depicted as an example in
Figure \ref{framesK21}.
As shown in Figure \ref{MWCSsusiePSNR}, the
performance of MWCS relies highly on the estimation of the tensor
rank. We examine the performance of MWCS with various rank
estimations. Experimental results demonstrate that GTCS
outperforms MWCS not only in speed, but also in accuracy.
\section{Conclusion}\label{conclusion}
Extensions of CS theory to multidimensional signals have become an
emerging topic. Existing methods include Kronecker compressive
sensing (KCS) for sparse tensors and multi-way compressive sensing
(MWCS) for sparse and low-rank tensors. We introduced the
Generalized Tensor Compressive Sensing (GTCS)--a unified framework
for compressive sensing of higher-order tensors which preserves
the intrinsic structure of tensorial data with reduced
computational complexity at reconstruction. We demonstrated that
GTCS offers an efficient means for representation of
multidimensional data by providing simultaneous acquisition and
compression from all tensor modes. We introduced two
reconstruction procedures, a serial method (GTCS-S) and a
parallelizable method (GTCS-P), and compared the performance of
the proposed methods with KCS and MWCS. As shown, GTCS outperforms
KCS and MWCS in terms of both reconstruction accuracy (within a
range of compression ratios) and processing speed. The major
disadvantage of our methods (and of MWCS as well), is that the
achieved compression ratios may be worse than those offered by
KCS. GTCS is advantageous relative to vectorization-based
compressive sensing methods such as KCS because the corresponding
recovery problems are in terms of a multiple small measurement
matrices $U_i$'s, instead of a single, large measurement matrix
$A$, which results in greatly reduced complexity. In addition,
GTCS-P does not rely on tensor rank estimation, which considerably
reduces the computational complexity while improving the
reconstruction accuracy in comparison with other tensorial
decomposition-based method such as MWCS.
\section*{Acknowledgment}
The authors would like to thank Dr. Edgar A. Bernal with PARC for
his help with the simulation as well as shaping the paper. We also
want to extend our thanks to the anonymous reviewers for their
constructive suggestions.
\ifCLASSOPTIONcaptionsoff
  \newpage
\fi


%
\bibliographystyle{IEEEtran}
\bibliography{icme2013template}
%
\begin{IEEEbiography}[{\includegraphics[width=1in,height=1.25in,clip,keepaspectratio]{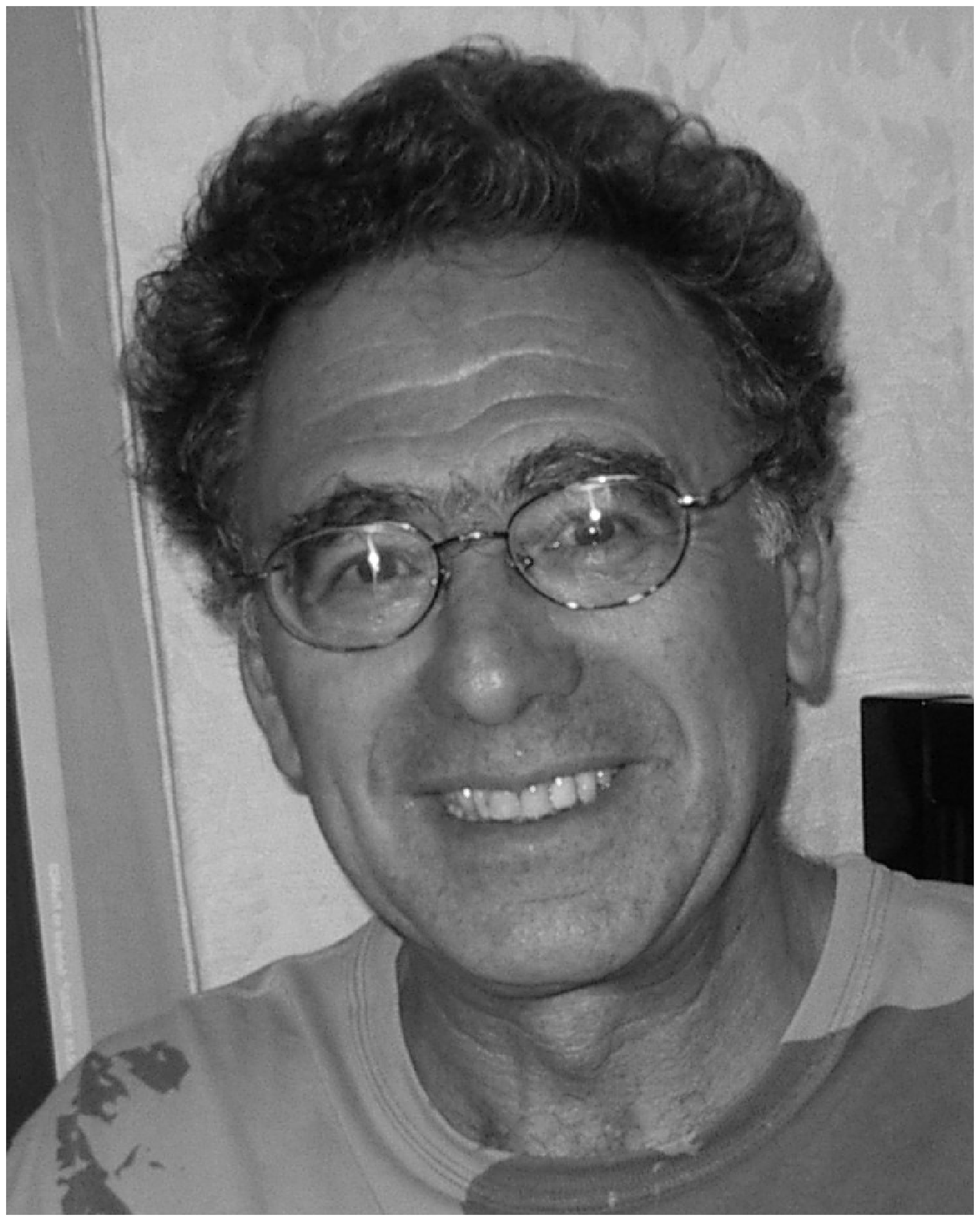}}]{Shmuel Friedland}
received all his degrees in Mathematics from Israel Institute of
Technology,(IIT), Haifa, Israel: B.Sc in 1967, M.Sc. in 1969, D.Sc. in 1971. He
held Postdoc positions in Weizmann Institute of Science, Israel; Stanford
University; IAS, Princeton. From 1975 to 1985, he was a member of Institute of
Mathematics, Hebrew U., Jerusalem, and was promoted to the rank of Professor in
1982. Since 1985 he is a Professor at University of Illinois at Chicago. He was
a visiting Professor in University of Wisconsin; Madison; IMA, Minneapolis;
IHES, Bures-sur-Yvette; IIT, Haifa; Berlin Mathematical School, Berlin.
Friedland contributed to the following fields of mathematics: one complex
variable, matrix and operator theory, numerical linear algebra, combinatorics,
ergodic theory and dynamical systems, mathematical physics, mathematical
biology, algebraic geometry. He authored about 170 papers, with many known
coauthors, including one Fields Medal winner. He received the first Hans
Schneider prize in Linear Algebra, jointly with M. Fiedler and I. Gohberg, in
1993. He was awarded recently a smoked salmon for solving the set-theoretic
version of the salmon problem: http://www.dms.uaf.edu/~eallman. For more
details on Friedland vita and research, see http://www.math.uic.edu/~friedlan.
\end{IEEEbiography}
\begin{IEEEbiography}[{\includegraphics[width=1in,height=1.25in,clip,keepaspectratio]{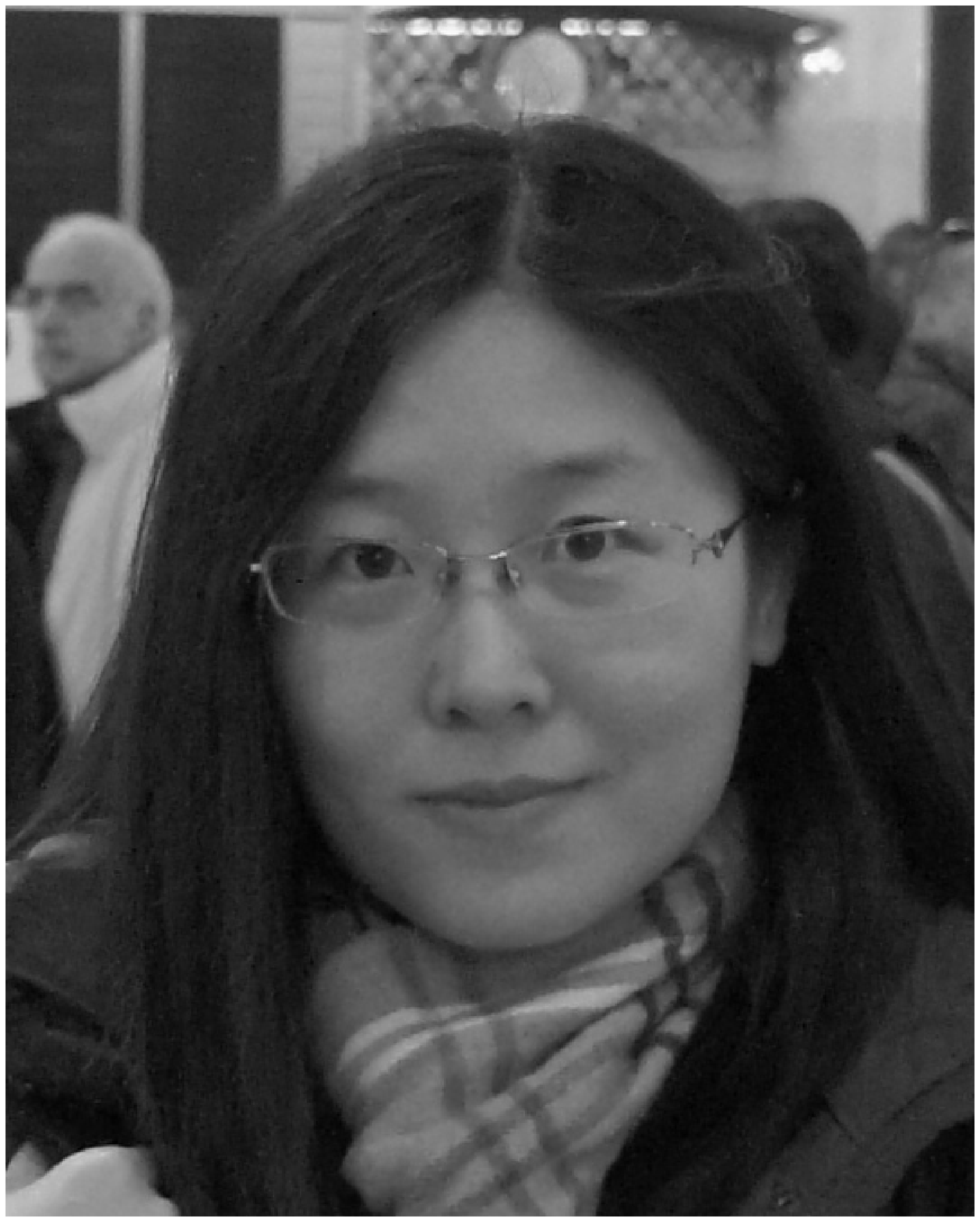}}]{Qun Li}
received the B.S. degree in Communications Engineering from
Nanjing University of Science and Technology, China, in 2009 and
the M.S. and Ph.D. degrees in Electrical Engineering from
University of Illinois at Chicago (UIC), U.S.A. in 2012 and 2013
respectively. She joined PARC, Xerox Corporation, NY, U.S.A. in
2013 as a research scientist after graduation. Her research
interests include machine learning, computer vision, image and
video analysis, higher-order data analysis, compressive sensing,
3D imaging, etc.
\end{IEEEbiography}
\begin{IEEEbiography}[{\includegraphics[width=1in,height=1.25in,clip,keepaspectratio]{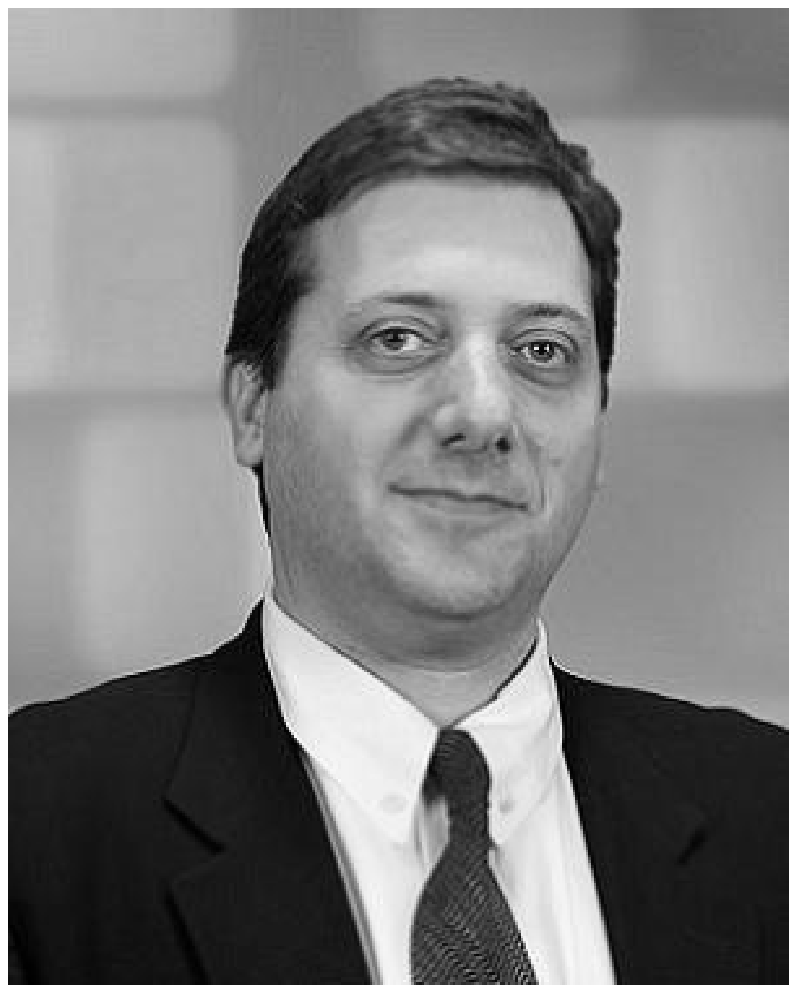}}]{Dan Schonfeld}
received the B.S. degree in electrical engineering and computer science from
the University of California, Berkeley, and the M.S. and Ph.D. degrees in
electrical and computer engineering from the Johns Hopkins University,
Baltimore, MD, in 1986, 1988, and 1990, respectively. He joined University of
Illinois at Chicago in 1990, where he is currently a Professor in the
Departments of Electrical and Computer Engineering, Computer Science, and
Bioengineering, and Co-Director of the Multimedia Communications Laboratory
(MCL). He has authored over 170 technical papers in various journals and
conferences.
\end{IEEEbiography}

\end{document}